\documentclass[final]{cvpr}
\usepackage{enumitem}
\usepackage{times}
\usepackage{epsfig}
\usepackage{graphicx}
\usepackage{amsmath}
\usepackage{amssymb}
\usepackage{booktabs}
\usepackage{multirow}
\usepackage{capt-of}
\usepackage{array}
\usepackage{wrapfig}

\usepackage[pagebackref=true,breaklinks=true,colorlinks,bookmarks=false]{hyperref}

\newcommand{\FWD}{FWD-U }
\newcommand{\FWDds}{FWD }
\newcommand{\FWDrgbd}{FWD-D }

\def\cvprPaperID{4419} 
\def\confYear{CVPR 2021}

\begin{document}

\title{FWD: Real-time Novel View Synthesis with Forward Warping and Depth}

\author{Ang Cao,  \, Chris Rockwell, \, Justin Johnson\\
University of Michigan, Ann Arbor\\
{\tt\small \{ancao,cnris,justincj\}@umich.edu}
}

\maketitle
\pagenumbering{gobble}
\begin{abstract}
    Novel view synthesis~(NVS) is a challenging task requiring systems to generate photorealistic images of scenes from new viewpoints, where both quality and speed are important for applications. 
    Previous image-based rendering~(IBR) methods are fast, but have poor quality when input views are sparse.
    Recent Neural Radiance Fields (NeRF) and generalizable variants give impressive results but are not real-time.
    In our paper, we propose a generalizable NVS method with sparse inputs, called \FWDds, which gives high-quality synthesis in real-time.
    With explicit depth and differentiable rendering, it achieves competitive results to the SOTA methods with 130-1000$\times$ speedup and better perceptual quality.
    If available, we can seamlessly integrate sensor depth during either training or inference to improve image quality while retaining real-time speed.
    With the growing prevalence of depths sensors,
    we hope that methods making use of depth will become increasingly useful.
\end{abstract}
\section{Introduction}

Given several posed images,  \emph{novel view synthesis}~(NVS) aims to generate photorealistic images depicting the scene from unseen viewpoints.
This long-standing task has applications in graphics, VR/AR, bringing life to still images.
It requires a deep visual understanding of geometry and semantics, making it appealing to test visual understanding.

\begin{figure}[ht!]
	\centering
			{\includegraphics[width=0.48\textwidth]{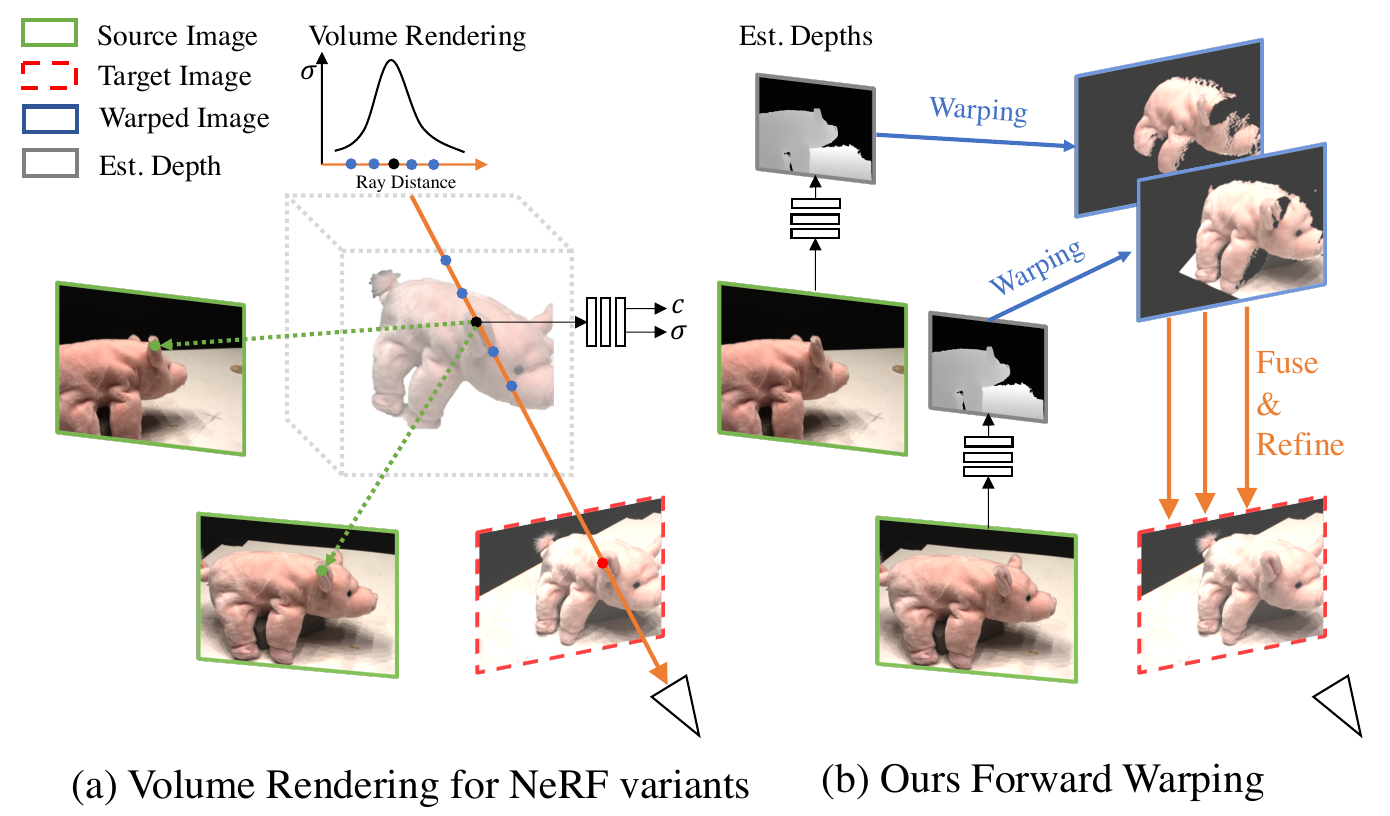}}
	\captionof{figure}{{\bf Real-time Novel View Synthesis.} 
We present a real-time and generalizable method to synthesize images from sparse inputs.
NeRF variants model the scene via an MLP, which is queried millions of times during rendering and leads to low speeds.
Our method utilizes explicit depths and point cloud renderers for fast rendering, inspired by SynSin~\cite{wiles2020synsin}. 
The model is trained end-to-end with a novel fusion transformer  to give high-quality results, where regressed depths and features are optimized for synthesis.
	}
	\label{fig:teaser}
	\vspace{-5mm}
\end{figure}

Early work on NVS focused on \emph{image-based rendering}~(IBR), where models generate target views from a set of input images.
Light field~\cite{levoy1996light} or proxy geometry~(like mesh surfaces)~\cite{choi2019extreme, hedman2018deep,riegler2020free,Riegler2021SVS} are typically constructed from posed inputs, and target views are synthesized by resampling or blending warped inputs.
Requiring dense input images, these methods are limited by 3D reconstruction quality,  and can perform poorly with \emph{sparse} input images. 

Recently, Neural Radiance Fields (NeRF) ~\cite{mildenhall2020nerf} have become the leading methods for NVS, using MLPs to represent the 5D radiance field of the scene  \emph{implicitly}.
The color and density of each sampling point are queried from the network and aggregated by volumetric rendering to get the pixel color. 
With dense sampling points and \emph{differentiable renderer}, 
explicit geometry isn't needed, and densities optimized for synthesis quality are learned.
Despite impressive results, they are not \emph{generalizable}, requiring MLP fitting for each scene with dense inputs.
Also, they are extremely \emph{slow} because of tremendous MLP query times for a single image.

Generalizable NeRF variants like  PixelNeRF~\cite{yu2020pixelnerf}, IBRNet~\cite{wang2021ibrnet} and MVSNeRF~\cite{chen2021mvsnerf} emerge very recently, synthesizing novel views of unseen scenes without per-scene optimization by modeling an MLP conditioned on sparse inputs. 
However, they still query the MLP millions of times, leading to slow speeds.
Albeit the progress of accelerating NeRF with per-scene optimization~\cite{yu2021plenoctrees,hedman2021baking,garbin2021fastnerf}, fast and generalizable NeRF variants are still under-explored.

In this paper, we target a \emph{generalizable} NVS method with \emph{sparse} inputs, refraining dense view collections. 
Both \emph{real-time speed} and \emph{high-quality} synthesis are expected, allowing interactive applications.
Classical IBR methods are fast but require dense input views for good results.
Generalizable NeRF variants show excellent quality without per-scene optimization
but require intense computations, leading to slow speeds.
Our method, termed FWD, achieves this target by  {\underline F}orward {\underline W}arping features based on {\underline D}epths.

Our key insight is that explicitly representing the \emph{depth} of each input pixel allows us to apply \emph{forward warping} to each input view using a differentiable point cloud renderer.
This avoids the expensive volumetric sampling used in NeRF-like methods, enabling real-time speed while maintaining high image quality.
This idea is deeply inspired by the success of SynSin~\cite{wiles2020synsin}, which employs a differentiable point cloud renderer for single image NVS. 
Our paper extends SynSin to multiple inputs settings and explores effective and efficient methods to fuse multi-view information.

Like prior NVS methods, our approach can be trained with RGB data only,
but it can be progressively enhanced
if noisy sensor depth data is available during training or inference.
Depth sensors are becoming more prevalent in consumer devices such as the iPhone 13 Pro and the LG G8 ThinQ, making RGB-D data more accessible than ever.
For this reason, we believe that methods making use of RGB-D will become increasingly useful over time.

Our method estimates depths for each input view to build a point cloud of latent features, then synthesizes novel views via a point cloud renderer.
To alleviate the inconsistencies between observations from various viewpoints, we introduce a view-dependent feature MLP into point clouds to model view-dependent effects.
We also propose a novel Transformer-based fusion module to effectively combine features from multiple inputs.
A refinement module is employed to inpaint missing regions and further improve synthesis quality. 
The whole model is trained end-to-end to minimize photometric and perceptual losses, learning depth and features optimized for synthesis quality.

Our design possesses several advantages compared with existing methods. 
First, it gives both high-quality and high-speed synthesis. 
Using explicit point clouds enables real-time rendering.
In the meanwhile,  differentiable renderer and end-to-end training empower high-quality synthesis results. 
Also, compared to NeRF-like methods, which cannot synthesize whole images during training because of intensive computations, 
our method could easily utilize perceptual loss and refinement module, which noticeably improves the visual quality of synthesis. 
Moreover, our model can seamlessly integrate sensor depths to further improve synthesis quality. 
Experimental results support these analyses.

We evaluate our method on the ShapeNet and DTU datasets, comparing it with representative NeRF-variants and IBR methods. 
It outperforms existing methods, considering speed and quality jointly:
compared to IBR methods we improve both speed and quality;
compared to recent NeRF-based methods we achieve competitive quality at realtime speeds~(130-1000$\times$ speedup).
A user study demonstrates that our method gives the most perceptually pleasing results among all methods.
The code is available at \url{https://github.com/Caoang327/fwd_code}.

\section{Related Work}

Novel view synthesis is a long-standing problem in computer vision, allowing for the generation of novel views given several scene images.
A variety of 3D representations (both implicit and explicit) have been used for NVS,
including  depth and multi-plane images \cite{tatarchenko2016multi, zhou2018stereo, srinivasan2019pushing, penner2017soft, chaurasia2013depth, shih20203d}, voxels \cite{sitzmann2019deepvoxels,guo2021fast}, meshes \cite{riegler2020free,hedman2018instant, hu2020worldsheet, Riegler2021SVS}, point clouds \cite{wiles2020synsin, liu2020infinite,Rockwell2021}
and neural scene representations \cite{rombach2021geometry, liu2020neural, genova2020local, jiang2020local, mescheder2019occupancy, park2019deepsdf,mildenhall2020nerf}.
In this work, we use point clouds as our 3D representations for computational and memory efficiency.

\par \noindent {\bf Image-based Rendering.}
IBR synthesizes novel views from a set of reference images by weighted blending~\cite{debevec1996modeling,levoy1996light,gortler1996lumigraph,hedman2018deep, penner2017soft, riegler2020free,choi2019extreme,Riegler2021SVS}.
They generally estimate proxy geometry from dense captured images for synthesis. 
For instance, Riegler et al.~\cite{riegler2020free} uses multi-view stereo~\cite{schoenberger2016mvs, yao2018mvsnet,wang2020patchmatchnet,wang2020patchmatchnet, luo2020consistent,Huang2018} to produce scene mesh surface and warps source view images to target views based on proxy geometry.
Despite promising results in some cases, they are essentially limited by the quality of 3D reconstructions, where dense inputs (tens to hundreds) with large overlap and reasonable baselines are necessary for decent results.
These methods estimate geometry as an intermediate task not directly optimized for image quality.
In contrast we input sparse views and learn depth jointly to optimize for synthesis quality.

\par \noindent {\bf Neural Scene Representations.}
Recent work uses implicit scene representations for view synthesis \cite{rombach2021geometry, liu2020neural, genova2020local, jiang2020local, mescheder2019occupancy, park2019deepsdf}.
Given many views, neural radiance fields~(NeRF) show impressive results~\cite{mildenhall2020nerf, zhang2020nerf++, martinbrualla2020nerfw, park2020nerfies, wang2021nerf},
but require expensive per-scene optimization.
Recent methods~\cite{wang2021ibrnet, yu2020pixelnerf, trevithick2020grf,chen2021mvsnerf, jain2021putting} generalize NeRF without per-scene optimization by learning a shared prior, with sparse inputs.
However these methods require expensive ray sampling and therefore are very slow.
In contrast, we achieve significant speedups using explicit representations.
Some concurrent work accelerates NeRF by reformulating the computation~\cite{garbin2021fastnerf}, using precomputation~\cite{yu2021plenoctrees,hedman2021baking},
or adding view dependence to explicit 3D representations~\cite{liu2020neural,wizadwongsa2021nex,yu2021plenoxels,TensoRF,mueller2022instant};
unlike ours, these all require dense input views and per-scene optimization.

\par  \noindent {\bf Utilizing RGB-D in NVS.}
The growing availability of annotated depth maps~\cite{dai2017scannet, Matterport3D, NIPS2016_0deb1c54, aanaes2016large,song2015sun, silberman2012indoor} facilitates depth utilization in NVS~\cite{novotny2019perspectivenet,liu2020infinite,hedman2016scalable}, which serves as extra supervision or input to networks.
Our method utilizes explicit depths as 3D representations, allowing using sensor depths as additional inputs for better quality.   
Given the increasing popularity of depth sensors, integrating sensor depths is a promising direction for real-world applications.
\begin{figure*}[t!]
	\centering
			{\includegraphics[width=\textwidth]{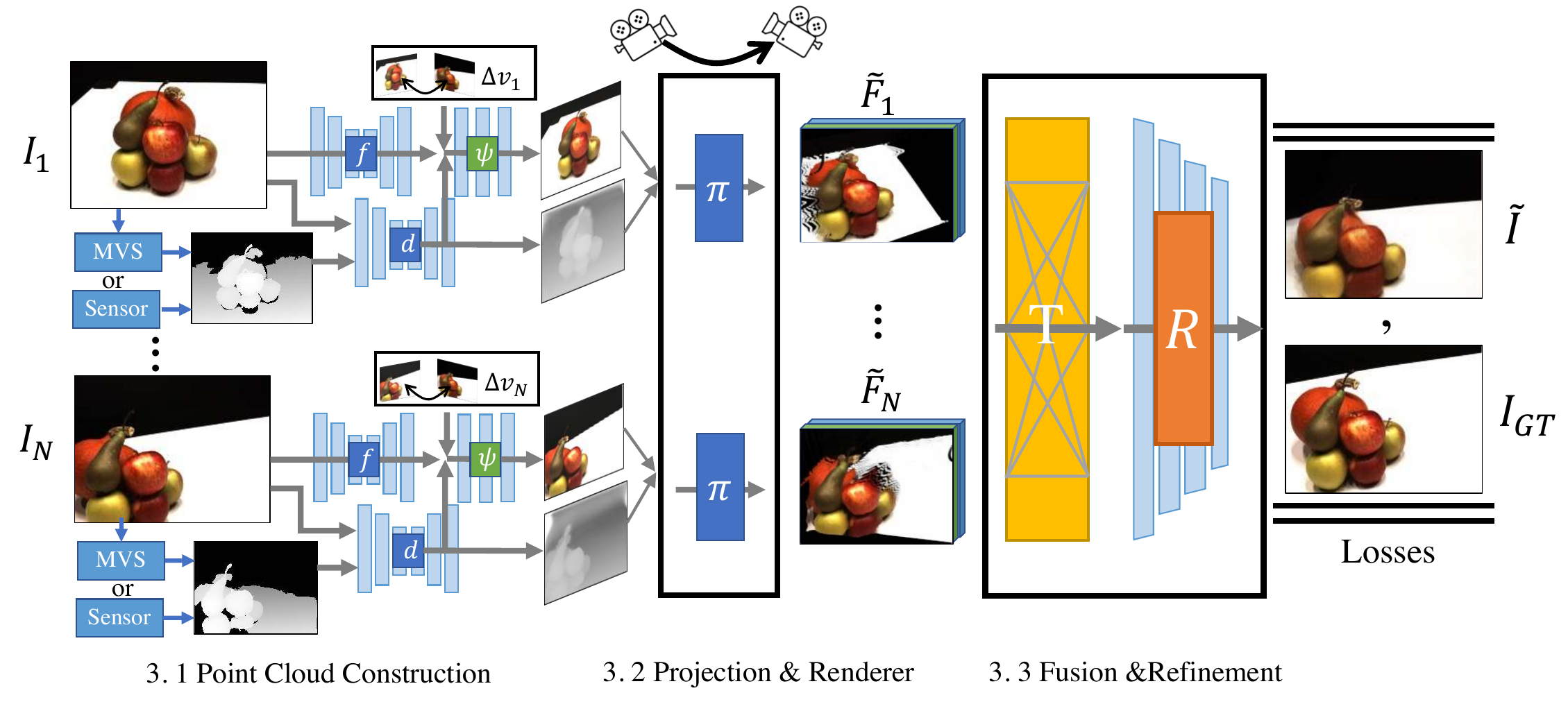}}
	\captionof{figure}{{\bf System Overview.} 
	Given a sparse set of images, we construct a point cloud $\mathcal{P}_{i}$ for each image $I_i$ using Feature Network $f$, View-Dependent Feature MLP $\psi$, and Depth Network $d$.
	Besides images, $d$ takes MVS estimated depths or sensor depths as inputs and regresses refined depths.
	Per-pixel features $F'_i$ are regressed by $f$ and $\psi$ based on images and relative view changes.
	A differentiable point cloud renderer $\pi$ is employed to project and render point clouds to target views.
	We use Transformer $T$ to fuse rendered results from arbitrary number inputs and apply refinement module $R$ for final results.
	The model is trained with photometric loss and content loss.  
	} 
	\label{fig:model}
	\vspace{-5mm}
\end{figure*}

Depth has been used in neural scene representations for speedups \cite{neff2021donerf, stelzner2021decomposing}, spaser inputs \cite{deng2021depth} and dynamic scenes\cite{xian2021space}. 
However, these works still require per-scene optimization.
Utilizing RGB-D inputs to accelerate generalizable NeRF like \cite{yu2020pixelnerf,wang2021ibrnet} is still an open problem.

\par \noindent {\bf Differentiable Rendering and Refinement.}
We use advances in differentiable rendering \cite{liu2019soft, jiang2020sdfdiff, chen2019learning,niemeyer2020differentiable,liu2020dist} to learn 3D end-to-end.
Learned geometries rely heavily on rendering and refinement \cite{yu2018generative, yang2017high, brock2018large, wang2018high} to quickly synthesize realistic results.
Refinement has improved dramatically owing to generative modeling \cite{ledig2017photo, karras2020analyzing, zhang2019self, CycleGAN2017} and rendering frameworks \cite{ravi2020accelerating, jatavallabhula2019kaolin, najibi2020dops, huang2020lstm}.
Instead of aggregating information across viewpoints before rendering~\cite{lombardi2019neural}, 
we render viewpoints separately and fuse using a Transformer \cite{46201, dosovitskiy2020image, carion2020end}, enabling attention across input views.

\section{Method}
Given a sparse set of input images $\{I_{i}\}_{i=1}^{N}$ and corresponding camera poses $\{R_i, T_i\}$,
our goal is to synthesize a novel view with camera pose $\{R_t, T_t\}$ fast and effectively.
The depths $\{D^{sen}_{i}\}$ of $I_i$ captured from sensors are optionally available, which are generally incomplete and noisy.

The insight of our method is that using explicit depths and forward warping enables real-time rendering speed and tremendous accelerations. 
Meanwhile, to alleviate quality degradations caused by inaccurate depth estimations,  a differentiable renderer and well-designed fusion \& refinement modules are employed, encouraging the model to learn geometry and features optimized for synthesis quality.

As illustrated in Figure \ref{fig:model}, with estimated depths, input view~$I_{i}$ is converted to a 3D point cloud~$\mathcal{P}_{i}$ containing geometries and view-dependent semantics of the view.  
A differentiable neural point cloud renderer $\pi$ is used to project point clouds to target viewpoints. 
Rather than directly aggregating point clouds across views before rendering, we propose a Transformer-based module $T$ fusing rendered results at target view. 
Finally, a refinement module $R$ is employed to generate final outputs.
The whole model is trained end-to-end with photometric and perceptual loss. 

\subsection{Point Cloud Construction}\label{sec:met_pce}
We use point clouds to represent scenes due to their efficiency, compact memory usage, and scalability to complex scenes.
For input view $I_i$, point cloud $\mathcal{P}_{i}$ is constructed by estimating depth $D_i$ and feature vectors $F'_i$ for each pixel in the input image, then projecting the feature vectors into 3D space using known camera intrinsics.
The depth $D_i$ is estimated by a \emph{depth network} $d$;
features $F'_i$ are computed by a \emph{spatial feature encoder} $f$ and \emph{view-dependent MLP} $\psi$.

\par \noindent {\bf Spatial Feature Encoder $f$.}
Scene semantics of input view $I_i$ are mapped to per-pixel feature vectors $F_i$ by spatial feature encoder $f$.
Each feature vector in $F_i$ is 61-dimensions and is concatenated with RGB channels, which is 64 dimensions in total.
$f$ is built on BigGAN architecture~\cite{brock2018large}.
 
\par \noindent {\bf Depth Network $d$.}
Estimating depth from a single image has scaling/shifting ambiguity, losing valuable multi-view cues and leading to inconsistent estimations across views.
Applying multi-view stereo algorithms~(MVS)~\cite{schoenberger2016mvs,yao2018mvsnet,wang2020patchmatchnet,9341659} solely on sparse inputs is challenging because of limited overlap and huge baselines between input views, leading to inaccurate and low-confidence estimations.
Therefore, we employ a hybrid design cascading a U-Net after the MVS module.
The U-Net takes image $I_i$ and estimated depths from the MVS module as inputs, refining depths with multi-view stereo cues and image cues. 
PatchmatchNet~\cite{wang2020patchmatchnet} is utilized as the MVS module, which is fast and lightweight.

\par \noindent {\bf Depth Estimation with sensor depths.}
As stated, U-Net receives an initial depth estimation from the MVS module and outputs a refined depth used to build the point cloud.   
If sensor depth $D^{sen}_{i}$ is available,  it is directly input to the U-Net as the initial depth estimations. 
In this setting, U-Net servers as completion and refinement module taking $D^{sen}_{i}$ and $I_i$ as inputs, since  $D^{sen}_{i}$ is usually noisy and incomplete.
During training,  loss $L_{s}$ is employed to encourage the U-Net output to match the sensor depth.
\begin{align}
    \mathcal{L}_{s} = \| M_i \odot D_i - M_i \odot D^{sen}_{i} \|
\end{align}
where $M_i$ is a binary mask indicating valid sensor depths.

\par \noindent {\bf View-Dependent Feature MLP $\psi$.}
The appearance of the point could vary across views because of lighting and view direction, causing inconsistency between multiple views.
Therefore, we propose to insert view direction changes into scene semantics to model this view-dependent effects. 
An MLP $\psi$ is designed to compute view-dependent features $F'_i$ by taking $F_i$ and relative view changes $\Delta v$ from input to target view as inputs.
For each point in the cloud, $\Delta v$ is calculated based on normalized view directions $v_i$ and $v_t$,  from the point to camera centers of input view $i$ and target view $t$.
The relative view direction change is calculated as: 
\begin{align}
	\Delta v = [(v_i - v_t) / \| v_i - v_t \|, v_i \cdot v_t], v_i, v_t \in \mathbb{R}^3.
\end{align}
and the view-dependent feature $F'_i$  is:
\begin{align}
	F'_{i} = \psi(F_{i}, \delta(\Delta v))
\end{align}
where $\delta$ is a two-layer MLP mapping $\Delta v$ to a 32-dimensions vector and $\psi$ is also a two-layer MLP.

\subsection{Point Cloud Renderer}\label{sec:met_p}
To observe the constructed point cloud $\mathcal{P}_i$  at target views, we employ a  neural point cloud renderer $\pi$.
$\mathcal{P}_i$ is first transformed to target view coordinates based on camera poses and then rendered by $\pi$.
The rendered feature maps $\tilde{F}_i$ share the same dimension as feature $F'_i$ at each pixel.
With explicit geometry transformation, our rendered results are geometrically consistent and correct across views.

We use the \emph{differentiable} renderer design of~\cite{wiles2020synsin}, which splats 3D points to the image plane and gets pixel values by blending point features.
The blending weights are computed based on \emph{z}-buffer depths and distances between pixel and point centers.
It is implemented using Pytorch3D~\cite{ravi2020accelerating}.

This fully differentiable renderer allows our model to be trained end-to-end, where photometric and perceptual loss gradients can be propagated to points' position and features.   
In this way, the model learns to estimate depths and features optimized for synthesis quality, leading to superior quality.
We show the effectiveness of it in experiments.
\subsection{Fusion and Refinement}\label{sec:met_fr}

Unlike SynSin~\cite{wiles2020synsin} using a single image for NVS, fusing multi-view inputs is required in our method.
A na\"ive fusion transforms each point cloud to target view and aggregates them into a large one for rendering. 
Despite high efficiency, it is vulnerable to inaccurate depths since points with wrong depths may occlude points from other views, leading to degraded results.
Methods like PointNet~\cite{qi2017pointnet} may be feasible to apply on the aggregated point cloud for refinement, but they are not efficient with significant point numbers.

Instead,  we render each point cloud individually at target viewpoints and fuse rendered results by a fusion Transformer $T$.
A refinement module $R$ is used to inpaint missing regions, decode feature maps and improve synthesis quality.

\begin{figure}[t!]
	\centering
			{\includegraphics[width=0.49\textwidth]{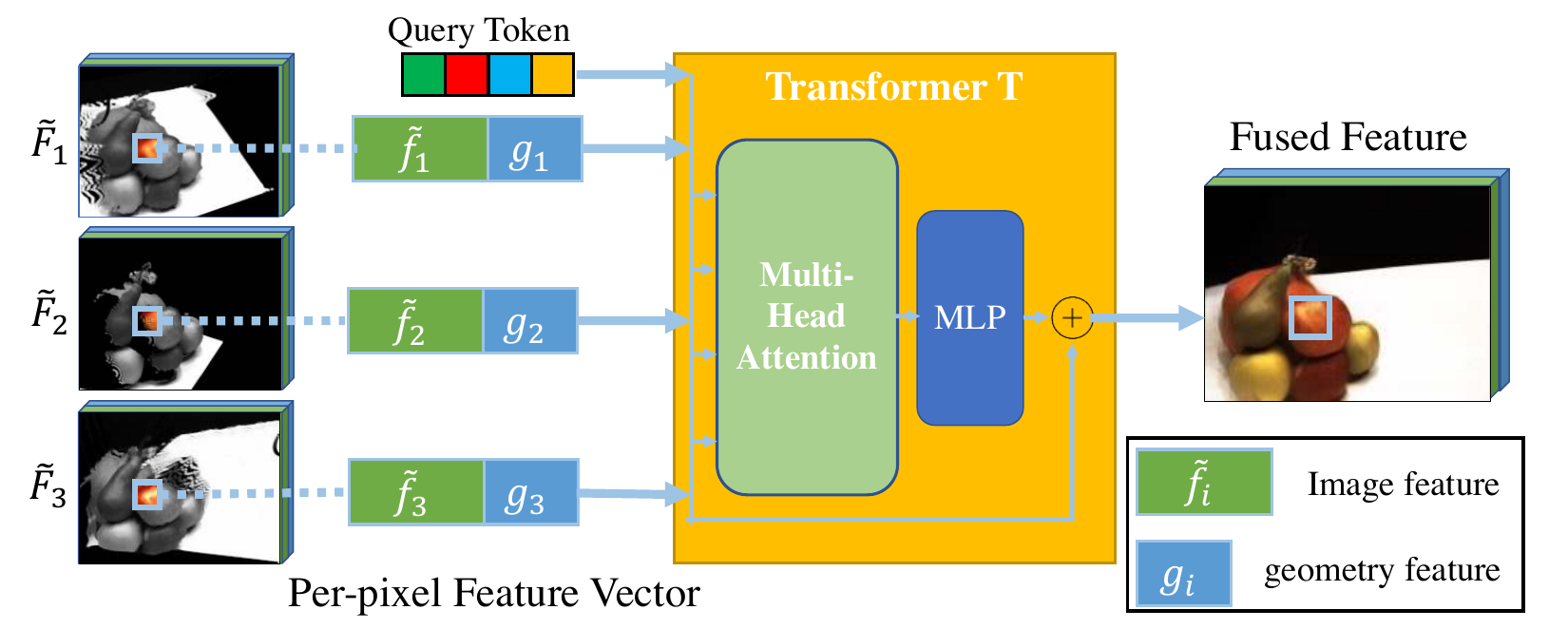}}
    \captionof{figure}{{\bf Fusion Transformer.} 
	We use a lightweight transformer $T$ to fuse the features from $N$ input views on each pixel.
	We use a learnable token to query the fusion results.
}
	\label{fig:model2}
	\vspace{-0.1in}
\end{figure}
\par \noindent {\bf Fusion Transformer $T$.}
Given an arbitrary number of rendered feature maps $\{\tilde{F}_i\}$, fusion should be effective, fast, and permutation invariant. 
Inspired by the success of Transformers, we propose a pixel-wise Transformer $T$ for fusion, detailed in Figure~\ref{fig:model2}.
At each pixel, $T$ inputs rendered feature vectors and queries fused results using a learnable ``token".
Applied on features, $T$ utilizes semantics for fusion.

Rendered results may lose geometry cues for fusion when rendered from 3D to 2D.
For instance, depths may reveal occlusion relationships across views,
and relative view changes from input to target views relate to each input's importance for fusion.
Therefore, we also explored to use geometry features as position encoding while not helpful.  

\par \noindent {\bf Refinement Module $R$.}
Built with 8 ResNet~\cite{he2016deep} blocks,  $R$ decodes fused feature maps $\tilde{F}$ to RGB images $\tilde{I}$ at target views. 
It inpaints regions  invisible for inputs in a semantically and geometrically meaningful manner.
Also, it corrects local errors caused by inaccurate depths and improves perceptual quality based on semantics contained by feature maps, leading to coherent and high-quality synthesis. 

\subsection{Training and Implementation Details}
\label{sec:method_details}
Our model is trained end-to-end with photometric $\mathcal{L}_{l_2}$ and perceptual $\mathcal{L}_{c}$ losses between generated and ground-truth target images.
The whole loss function is:
\begin{align}
	\mathcal{L} = \lambda_{l_2}\mathcal{L}_{l_2} + \lambda_{c} \mathcal{L}_c 
\end{align}

where $\lambda_{l_2}=5.0, \lambda_{c}=1.0$.
The model is trained end-to-end on 4 2080Ti GPUs for 3 days, using Adam~\cite{kingma2014adam} with learning rate $10^{-4}$ and $\beta_1{=}0.9, \beta_2{=}0.999$.
When sensors depths are available as inputs, $\mathcal{L}_{s}$  is used with $\lambda_{s}=5.0$.

\begin{table}[t]
    \centering
    \caption{
    \textbf{Model variants settings.}
    We predefine three model variants with different settings.
    \FWDds utilizes a pre-trained MVS module, in which way it gets access to depths during training.}
    \vspace{1mm}
    \resizebox{0.48 \textwidth}{!}{
        \begin{tabular}{p{1.2cm} p{0.75cm} p{0.75cm} p{2.4cm} p{2cm} p{2cm}}
    \toprule[2pt]
    Name & Test Depth & Train Depth &Depth Network &MVS  Module & Losses\\
    \midrule
    \FWD &  & &MVS + U-Net & Random ini. &$\mathcal{L}_{l_2} + \mathcal{L}_{c}$\\
    \FWDds &  & \checkmark  &MVS + U-Net & Pre-trained &$\mathcal{L}_{l_2} + \mathcal{L}_{c}$\\
    \FWDrgbd &\checkmark  & \checkmark & RGB-D + U-Net & - &$\mathcal{L}_{l_2} + \mathcal{L}_{c} + \mathcal{L}_{s}$\\
    \bottomrule [2 pt]
    \end{tabular}}
    \label{tab:name_table}
    \vspace{-5mm}
\end{table}


\section{Experiments}
The goal of our paper is \emph{real-time} and \emph{generalizable} novel view synthesis with \emph{sparse} inputs, which can optionally use sensor depths.
To this end, our experiments aim to identify the speed and quality at which our method can synthesize novel images and explore the advantage of explicit depths.
We evaluate our methods on ShapeNet~\cite{chang2015shapenet} and DTU~\cite{jensen2014large} datasets, comparing results with the SOTA methods and alternative approaches.
Experiments take place with held-out test scenes and no per-scene optimization.
We conduct ablations to validate the effectiveness of designs.

\par \noindent {\bf Metrics.}
We conduct A/B test to measure the visual quality, in which workers select the image most similar to the ground truth from competing methods.
Automatic image quality metrics including PSNR, SSIM~\cite{wang2004image} and LPIPS~\cite{zhang2018unreasonable} are also reported, and we find LPIPS best reflects the image quality as perceived by humans.
Frames per second~(FPS) during rendering is measured on the same platform~(single 2080Ti GPU with 4 CPU cores).
All evaluations are conducted using the same protocol~(same inputs and outputs).

\par \noindent {\bf Model Variants.}
Three models are evaluated with various accessibility to depths for training and test, as defined in Table~\ref{tab:name_table}.
\FWDds utilizes a pretrained PatchmatchNet~\cite{wang2020patchmatchnet} as the MVS module for depth estimations,
which is also updated during end-to-end training with photometric and perceptual loss. 
\FWD learns depth estimations in an {\underline U}nsupervised manner, sharing the same model and settings as \FWDds while PatchmatchNet is randomly initialized without any pretraining.
\FWDrgbd takes sensor depths as additional inputs during both training and inference.
It doesn't use any MVS module since sensor depths provide abundant geometry cues.
For pretraining PatchmatchNet, we train it following typical MVS settings and using the same  data splitting as NVS.


\begin{figure}[t!]
    \centering
    \begin{tabular}{c@{\vrule}c@{\hspace{0.5mm}}c@{\hspace{0.5mm}\vrule}c@{\hspace{0.5mm}}c@{\hspace{0.5 mm}\vrule}c@{\hspace{0.5mm}}c@{\hspace{0.5mm}}c@{}}
    \multicolumn{1}{c}{Input}& \multicolumn{2}{c}{PixelNeRF} & \multicolumn{2}{c}{\FWD}& \multicolumn{2}{c}{GT}\\
    \includegraphics[width=0.065\textwidth]{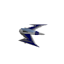}&
    \includegraphics[width=0.065\textwidth]{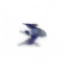}&
    \includegraphics[width=0.065\textwidth]{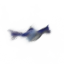}&
    \includegraphics[width=0.065\textwidth]{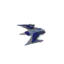}&
    \includegraphics[width=0.065\textwidth]{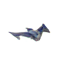}&
    \includegraphics[width=0.065\textwidth]{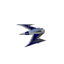}&
    \includegraphics[width=0.065\textwidth]{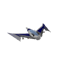}\\
    \includegraphics[width=0.065\textwidth]{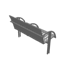}&
    \includegraphics[width=0.065\textwidth]{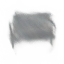}&
    \includegraphics[width=0.065\textwidth]{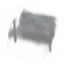}&
    \includegraphics[width=0.065\textwidth]{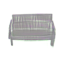}&
    \includegraphics[width=0.065\textwidth]{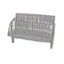}&
    \includegraphics[width=0.065\textwidth]{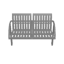}&
    \includegraphics[width=0.065\textwidth]{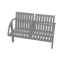}\\
    \includegraphics[width=0.065\textwidth]{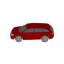}&
    \includegraphics[width=0.065\textwidth]{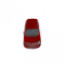}&
    \includegraphics[width=0.065\textwidth]{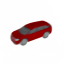}&
    \includegraphics[width=0.065\textwidth]{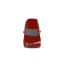}&
    \includegraphics[width=0.065\textwidth]{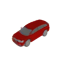}&
    \includegraphics[width=0.065\textwidth]{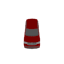}&
    \includegraphics[width=0.065\textwidth]{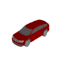}\\
    \includegraphics[width=0.065\textwidth]{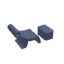}&
    \includegraphics[width=0.065\textwidth]{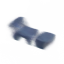}&
    \includegraphics[width=0.065\textwidth]{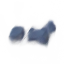}&
    \includegraphics[width=0.065\textwidth]{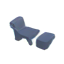}&
    \includegraphics[width=0.065\textwidth]{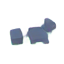}&
    \includegraphics[width=0.065\textwidth]{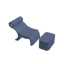}&
    \includegraphics[width=0.065\textwidth]{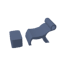}\\
    \end{tabular}
    \captionof{figure}{{\bf Qualitative results of category-agnostic NVS on ShapeNet.} 
    We test the capacity of our model by training it across 13 categories of ShapeNet with single view input,
    and compare with PixelNeRF~\cite{yu2020pixelnerf}.
    No gt depths are available during inference and training. 
    Our results have better visual quality and details.
}
    \label{fig:shapenet}
    \vspace{-2mm}
\end{figure}

\begin{table}[t!]
    \centering
    \caption{
    \textbf{Category-agnostic NVS on ShapeNet.}
    Quantitative results for category-agnostic view-synthesis are presented.
    }
    \vspace{1mm}
    \resizebox{0.48\textwidth}{!}{
    \begin{tabular}{p{1.3cm} p{0.6cm} p{0.6cm} p{0.6cm} p{0.5cm} | p{0.6cm} p{0.6cm} p{0.6cm} p{0.5cm}}
    \toprule[2pt]
          & \multicolumn{4}{c|}{1-view} & \multicolumn{4}{c}{2-view} \\
    model & PSNR & SSIM & LPIPS & FPS & PSNR & SSIM & LPIPS & FPS\\
    \midrule
    DVR~\cite{DVR}   & 22.70 & 0.860 & 0.130 & 1.5  & - & - & - & - \\
    SRN~\cite{NEURIPS2019_b5dc4e5d}   & 23.28 & 0.849 & 0.139 & 24 & - & - & - & - \\
    PixelNeRF & \bf 26.80 & 0.910 & 0.108 & 1.2 & \bf 28.88 &\bf 0.936& 0.076 & 1.1\\
    \FWD &  26.66 &  \bf0.911 & \bf0.055& \bf 364 &28.43 &0.931&\bf 0.043& \bf 336\\
    \bottomrule [2 pt]
    \end{tabular} %
    }
    \label{tab:shapenet}
    \vspace{-3mm}
\end{table}

\newlength{\figfivewidth}
\setlength{\figfivewidth}{0.11\textwidth}
\begin{figure*}[t]
   \centering
    \begin{tabular}{>{\footnotesize}c@{}@{}c@{\hspace{0.5mm}}c@{\hspace{0.5mm}}c@{\hspace{2mm} \unskip \vrule \hspace{2mm}}c@{\hspace{0.5mm}}c@{\hspace{0.5 mm}}c@{\hspace{0.5mm}}c@{\hspace{0.5mm}}c@{}}
   \multicolumn{4}{c}{Input: 3 views of held-out scene} & \multicolumn{5}{c}{Novel views}\\
   \multirow{1}{*}[7ex]{\rotatebox[origin=c]{90}{\FWDrgbd}}&
   \includegraphics[width=\figfivewidth]{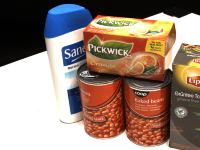}&
   \includegraphics[width=\figfivewidth]{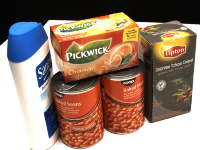}&
   \includegraphics[width=\figfivewidth]{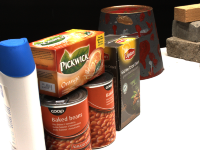}&
   \includegraphics[width=\figfivewidth]{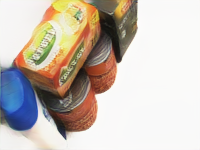}&
   \includegraphics[width=\figfivewidth]{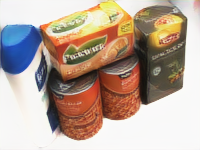}&
   \includegraphics[width=\figfivewidth]{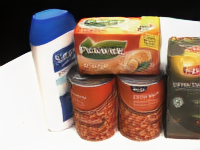}&
   \includegraphics[width=\figfivewidth]{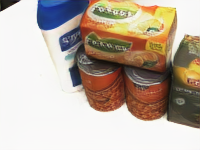}&
   \includegraphics[width=\figfivewidth]{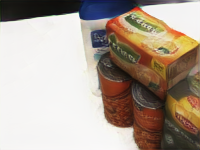}\\
   \multirow{1}{*}[6ex]{\rotatebox[origin=c]{90}{\FWDds}} &
   \includegraphics[width=\figfivewidth]{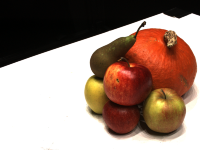}&
   \includegraphics[width=\figfivewidth]{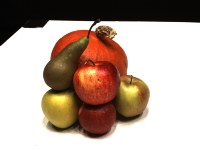}&
   \includegraphics[width=\figfivewidth]{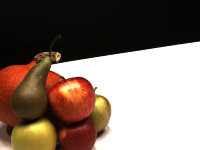}&
   \includegraphics[width=\figfivewidth]{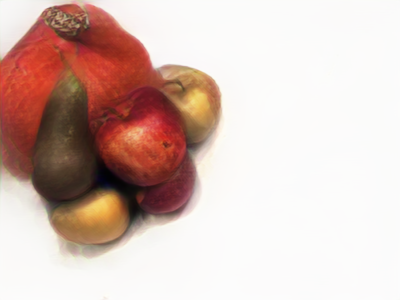}&
   \includegraphics[width=\figfivewidth]{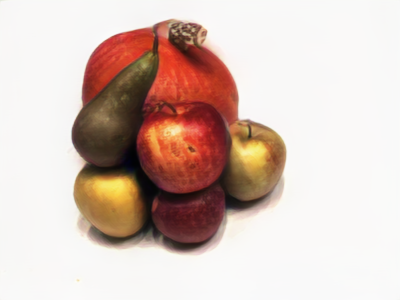}&
   \includegraphics[width=\figfivewidth]{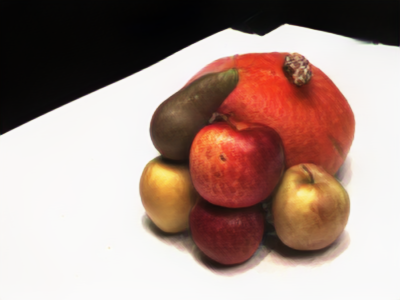}&
   \includegraphics[width=\figfivewidth]{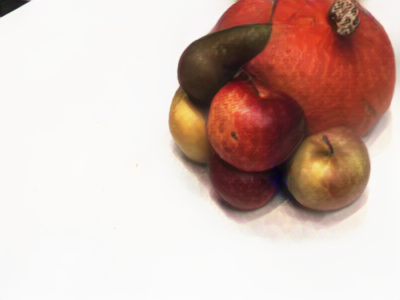}&
   \includegraphics[width=\figfivewidth]{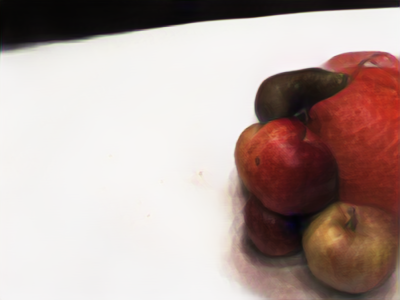}\\
   \multirow{1}{*}[7ex]{\rotatebox[origin=c]{90}{\FWD}} &
   \includegraphics[width=\figfivewidth]{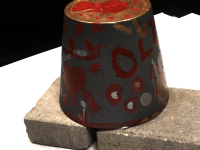}&
   \includegraphics[width=\figfivewidth]{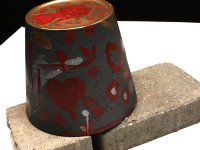}&
   \includegraphics[width=\figfivewidth]{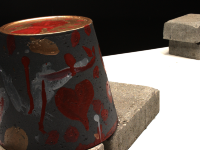}&
   \includegraphics[width=\figfivewidth]{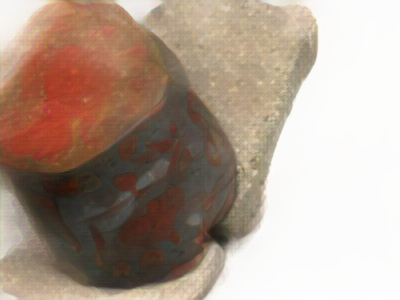}&
   \includegraphics[width=\figfivewidth]{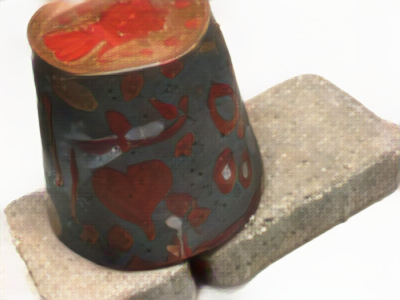}&
   \includegraphics[width=\figfivewidth]{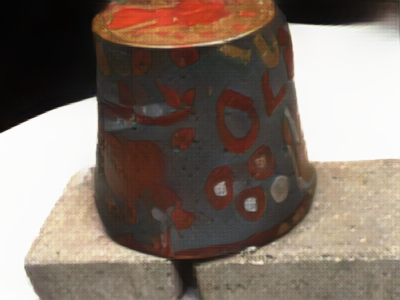}&
   \includegraphics[width=\figfivewidth]{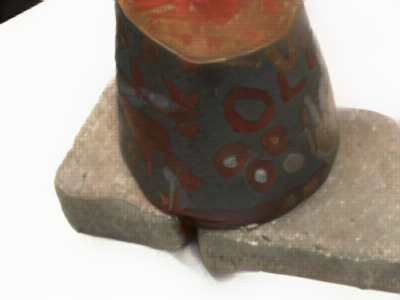}&
   \includegraphics[width=\figfivewidth]{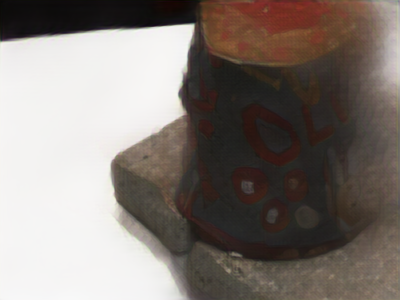}\\
   \end{tabular}
   \caption{\textbf{View synthesis results from FWD.}
   We show the view synthesis results with 3 input views on DTU dataset from \FWDrgbd (row. 1), \FWDds (row. 2) and \FWD (row. 3). 
   Our methods synthesize high-quality and geometrically correct novel views in real time.   
   }
   \label{fig:dtu_gt_depth}
   \vspace{-5mm}
   \end{figure*}

\subsection{ShapeNet Benchmarks}
We first evaluate our model for category-agnostic synthesis task on ShapeNet~\cite{chang2015shapenet}.
Following the setting of \cite{yu2020pixelnerf}, we train and evaluate a single model on 13 ShapeNet~\cite{chang2015shapenet} categories.
Each instance contains 24 fixed views of 64 $\times$ 64 resolution.
During training, one random view is selected as input and the rests are served as target views.
For testing, we synthesize all the other views from a fixed informative view.
The model is finetuned  with two random input views for 2-view experiments.
We find that U-Net is sufficient for good results on this dataset without the MVS module.

Qualitative comparisons to PixelNeRF are shown in Figure~\ref{fig:shapenet}, where \FWD gets noticeably superior results.
Our synthesized results are more realistic and closely matching to target views, while PixelNeRF's results tend to be blurry.
We observe the same trend in the DTU benchmark and evaluate the visual quality quantitatively there.

We show quantitative results in Table~\ref{tab:shapenet}, adding SRN~\cite{NEURIPS2019_b5dc4e5d} and DVR~\cite{DVR} as other baselines. 
Our method outperforms others significantly for LPIPS, indicating a much better perceptual quality,  as corroborated by qualitative results.
PixelNeRF has a slightly better PSNR while its results are blurry. 
Most importantly, \FWD runs at a speed of over 300 FPS, which is 300$\times$ faster than PixelNeRF.

\subsection{DTU MVS Benchmarks}
\label{exp:DTU}
We also evaluate our model on DTU MVS dataset~\cite{jensen2014large}, which is a real scene dataset consisting of 103 scenes.
Each scene contains one or multiple objects placed on a table, while images and incomplete depths are collected by the camera and structured light scanner mounted on an industrial robot arm.
Corresponding camera poses are provided.

As stated in ~\cite{yu2020pixelnerf}, this dataset is challenging since it consists of complex real scenes without apparent semantic similarities across scenes.
Also, images are taken under varying lighting conditions with distinct color inconsistencies between views.
Moreover, with only under 100 scenes available for training, it is prone to overfitting in training. 

We follow the same training and evaluation pipelines as PixelNeRF~\cite{yu2020pixelnerf} for all methods to give a fair comparison. 
The data consists of 88 training and 15 test scenes, between which there are no shared or highly similar scenes.
Images are down-sampled to a resolution of 300 $\times$ 400.
For training, three input views are randomly sampled, with the rest as target views.
For inference, we choose three fixed informative input views and synthesize other views of the scene.

\begin{figure*}[thp!]
    \centering
    \begin{tabular}{@{}c@{\hspace{0.125mm}\hspace{0.125mm}}c@{\hspace{0.5mm}}c@{\hspace{0.25mm}}c@{\hspace{0.25mm}}c@{\hspace{0.25 mm}}c@{\hspace{0.25mm}}c@{\hspace{0.25mm}}c@{\hspace{0.25mm} }c@{}}

    \small{Input Image} & \small {PixelNeRF} &\small{IBRNet}&\small \FWD & \small \FWDds & \small Blending+R  & \small \FWDrgbd   & \small Target View\\
    \includegraphics[width=0.12\textwidth]{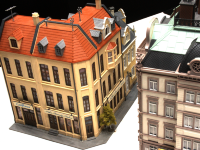}&
    \includegraphics[width=0.12\textwidth]{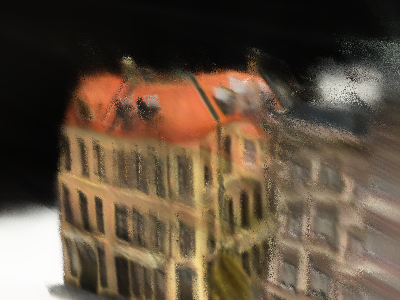}&
    \includegraphics[width=0.12\textwidth]{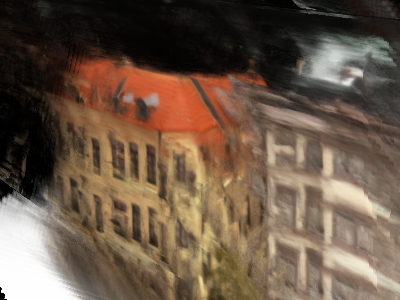}&
    \includegraphics[width=0.12\textwidth]{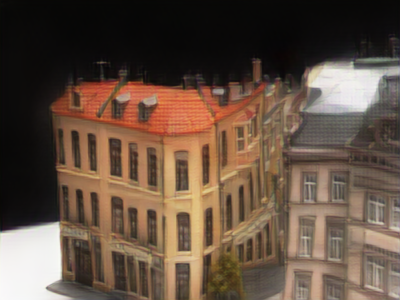}&
    \includegraphics[width=0.12\textwidth]{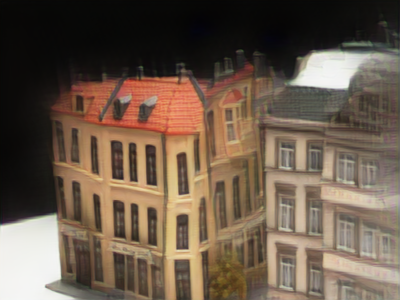}&
    \includegraphics[width=0.12\textwidth]{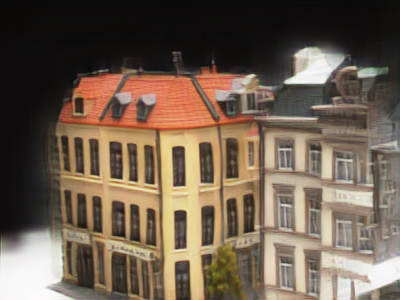}&
    \includegraphics[width=0.12\textwidth]{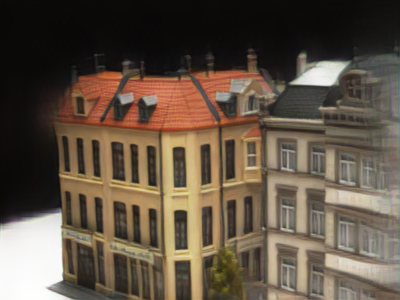}&
    \includegraphics[width=0.12\textwidth]{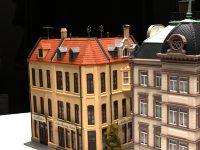}\\
    \includegraphics[width=0.12\textwidth]{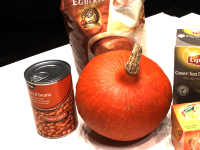}&
    \includegraphics[width=0.12\textwidth]{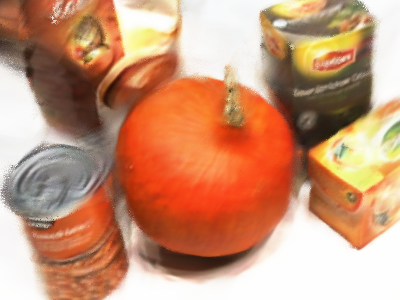}&
    \includegraphics[width=0.12\textwidth]{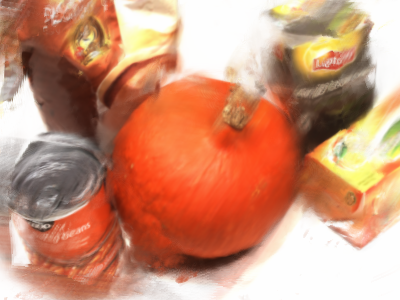}&
    \includegraphics[width=0.12\textwidth]{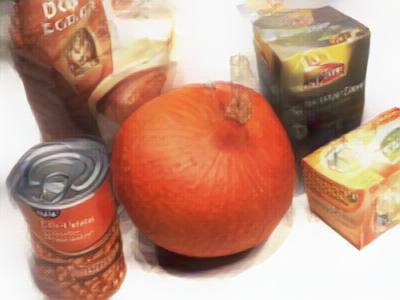}&
    \includegraphics[width=0.12\textwidth]{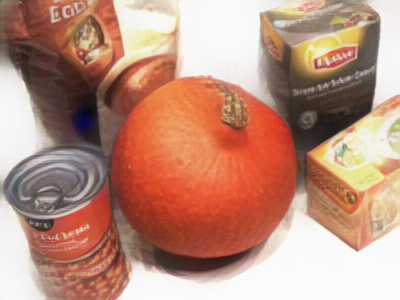}&
    \includegraphics[width=0.12\textwidth]{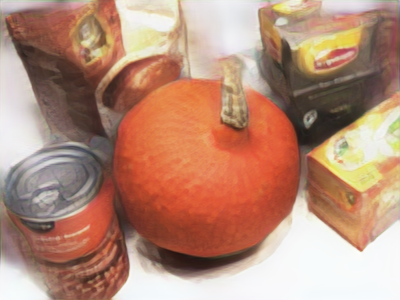}&
    \includegraphics[width=0.12\textwidth]{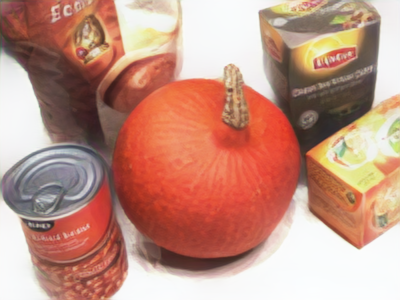}&
    \includegraphics[width=0.12\textwidth]{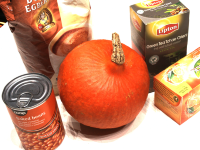}\\
    \includegraphics[width=0.12\textwidth]{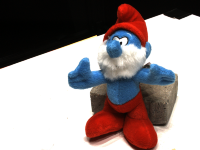}&
    \includegraphics[width=0.12\textwidth]{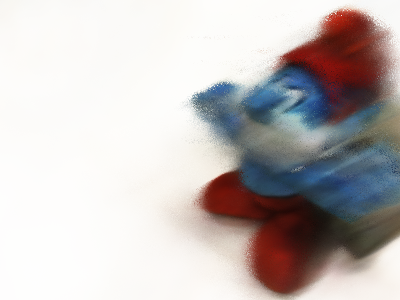}&
    \includegraphics[width=0.12\textwidth]{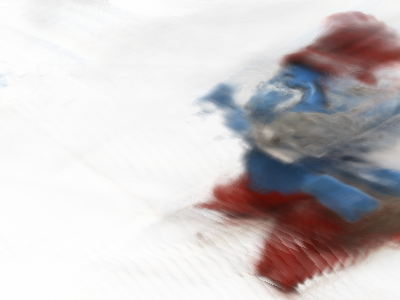}&
    \includegraphics[width=0.12\textwidth]{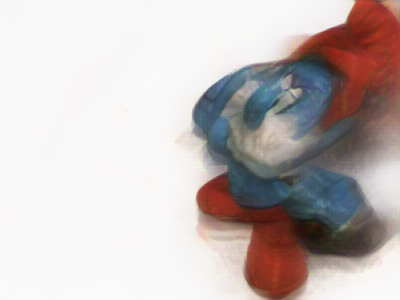}&
    \includegraphics[width=0.12\textwidth]{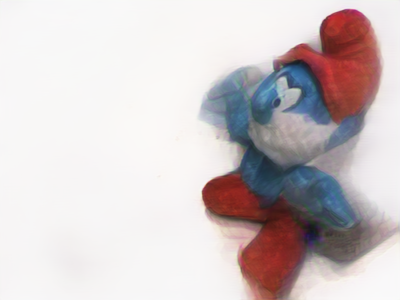}&
    \includegraphics[width=0.12\textwidth]{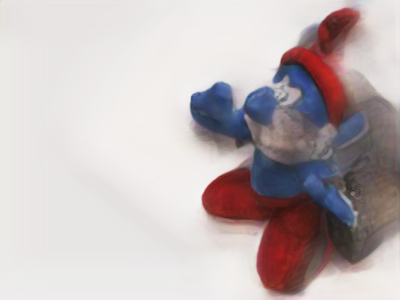}&
    \includegraphics[width=0.12\textwidth]{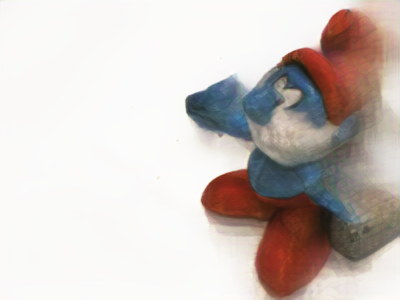}&
    \includegraphics[width=0.12\textwidth]{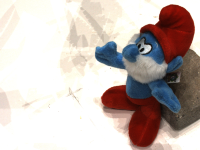}
    \end{tabular}
    \caption{
    \textbf{Qualitative Comparison.}
    We compare synthesis results from different methods with 3 input views~(one of them shown in figure).
    Our methods give geometrically consistent and visually appealing results, while other results suffering shaking artifacts at some views.
    Unlike other methods, \FWDrgbd and Blending+R get access sensor depths as inputs during inference. 
    }
    \label{fig:dtu_results_compar}
    \vspace{-4mm}
\end{figure*}

\begin{figure}
        \centering
              \includegraphics[width=0.45\textwidth]{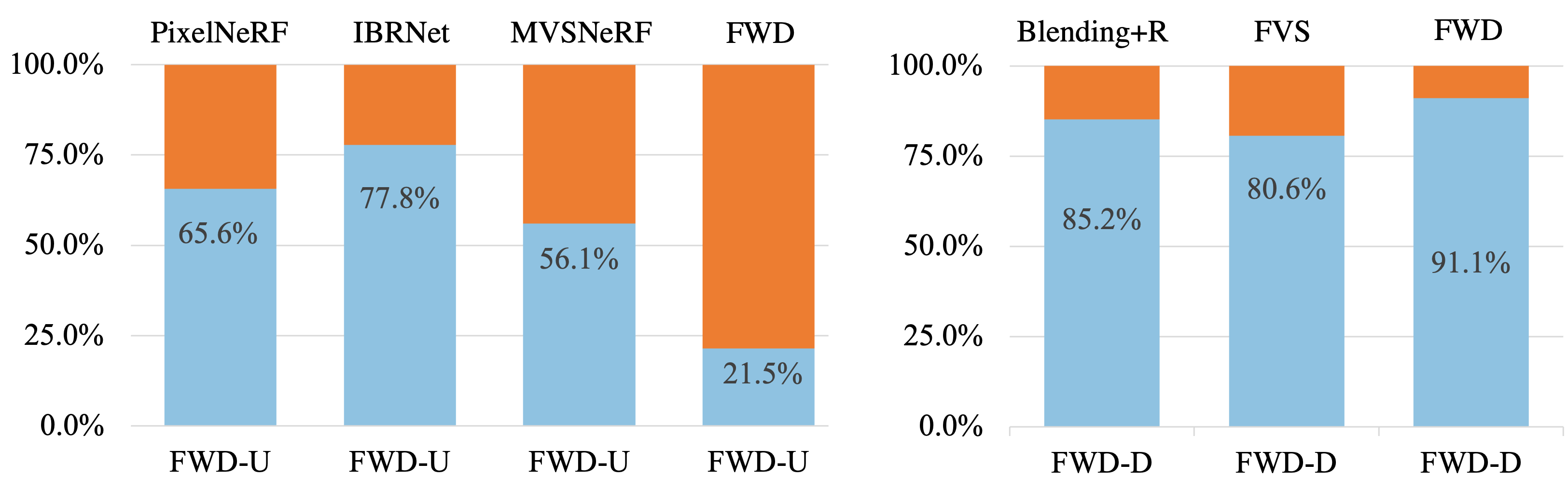}
    \captionof{figure}{{\bf User study on DTU.}
    We conduct a user study by asking subjects to select the results most similar to the ground truth.
    The numbers indicate the percentage of preference.
    Methods are grouped based whether using depths during test.
    We also report \FWDds vs. \FWDrgbd showing the advantages of sensor depths.}
    \label{fig:user_study}
    \vspace{-4mm}
    \end{figure}

\par \noindent {\bf Baselines.}
We evaluate a set of representatives of generalizable NeRF and IBR methods in two different scenarios: with RGB or RGB-D available as inputs during inference. 

PixelNeRF~\cite{yu2020pixelnerf}, IBRNet~\cite{wang2021ibrnet} and MVSNeRF~\cite{chen2021mvsnerf} are the SOTA generalizable NeRF variants, taking RGB as inputs.
We use the official PixelNeRF model trained on DTU MVS and carefully retrain IBRNet and MVSNeRF with the same 3-input-view settings.
PixelNeRF-DS is also included  as reported in~\cite{deng2021depth}, which is PixelNeRF supervised with depths.
Please note that our settings are very different from evaluations used in original papers of IBRNet and MVSNeRF.

A series of IBR methods are also evaluated. 
Since COLMAP~\cite{schoenberger2016mvs} fails to give reasonable outputs with sparse input images,
methods using COLMAP like FVS~\cite{riegler2020free}, DeepBlending~\cite{DeepBlending2018} cannot estimate scene geometry in this setting. 
For these methods, we use depths captured by sensors as estimated depths, which should give upper-bound performance of these methods.
To better cope with missing regions, we add our refinement model to DeepBlending~\cite{DeepBlending2018} and retrain it on DTU dataset, termed Blending-R.

For fairness, we evaluate all methods using the same protocol, distinct from some of their original settings. 
Although we try our best to adopt these methods, our reported results may still not perfectly reflect their true capacity.
\par \noindent {\bf Qualitative Results.}
Synthesis results are shown in Figure~\ref{fig:dtu_gt_depth}, where high-quality and geometrically correct novel views are synthesized in real-time (over 35 FPS) under significant viewpoint changes.
Our refinement module faithfully inpaints invisible regions; also, synthesized images have good shadows, light reflections, and varying appearances across views, showing the efficacy of view-dependent MLP.
With sensor depths, results can be further improved.

We show comparisons to baselines in Figure~\ref{fig:dtu_results_compar}. 
Our methods provide noticeably better results than baselines across different depth settings.
For models without depths in test, IBRNet and PixelNeRF give blurry results in areas of high detail such as the buildings in the top row,
while our \FWD and \FWDds give more realistic and sharper images. 
With sensor depths in test, baseline Blending-R produces more cogent outputs, but still struggles to distinguish objects from the background, such as in the middle row,
while \FWDrgbd gives faithfully synthesis and clear boundaries.

\par \noindent {\bf Quantitative Results.}
We evaluate synthesis quality quantitatively by user study following a standard A/B paradigm.
Workers choose the closest to a ground truth image between competing methods,
and are monitored using a qualifier and sentinel examples.
All views in the test set (690 in total) are evaluated, and each view is judged by three workers.

In Figure~\ref{fig:user_study}, user study results support qualitative observations.
Among all baselines with and without test depths , 
users choose our method as more closely matching ground truth images than others most of the time.
\FWD is selected over PixelNeRF in 65.6\% of examples, and 77.8\% compared to IBRNet. 
Also, over 90\% workers prefer \FWDrgbd to \FWDds, showing advantage of using sensor depths.

We show automated view synthesis metrics and speeds in Table~\ref{tab:dtu}.
Across all depth availability settings, our method is competitive with the SOTA baselines while significantly faster.
\FWDrgbd runs in real-time and gives substantially better image quality than others.
\FWDds has competitive metrics to PixelNeRF-DS while 1000$\times$ faster.
Notably, NeRF variants such as PixelNeRF, IBRNet, MVSNeRF, and PixelNeRF-DS are at least two orders of magnitude slower.

The exception to highly competitive performance is weaker PSNR and SSIM of our unsupervised \FWD against PixelNeRF and IBRNet.
However, \FWD has noticeably better perceptual quality with the best LPIPS,
and human raters prefer it to other methods in A/B tests.
The visual quality in figure~\ref{fig:dtu_results_compar} also illustrates the disparity between comparisons using PSNR and LPIPS.
Meanwhile, \FWD is above $1000 \times $ faster than PixelNeRF and above $100 \times$ faster than IBRNet.
Depth estimations, rendering and CNN would introduce tiny pixel shiftings, which harm the PSNR of our method.
NeRF-like methods are trained to optimize L2 loss for each pixel independently, leading to blur results.

Among all methods without test depths, \FWDds has the best results.
Although it uses a pretrained MVS module, we think this comparison is still reasonable since pretrained depth module is easy to get.
Also, training depths can be easily calculated from training images since they are dense.

Baseline comparisons also show that IBR methods are fast, but do not give images that are competitive with our method.
Our method outperforms them in both perceptual quality and standard metrics, showing the efficacy of proposed methods.
Note that Blending+R doesn't support variable number inputs and our refinement module improves its results significantly. 
We also compare \FWD with SynSin~\cite{wiles2020synsin} which only receives a single input image, showing the benefits of using multi-view inputs in NVS.

\subsection{Ablations and Analysis}

We evaluate the effectiveness of our designs and study depth in more detail through ablation experiments.

\begin{figure}[t]
    \centering
    \resizebox{0.48\textwidth}{!}{
    \begin{tabular}{@{}c@{\hspace{0.5mm}}c@{\hspace{0.5 mm}}c@{\hspace{0.5mm}}c@{}}
    \includegraphics[width=0.16\textwidth]{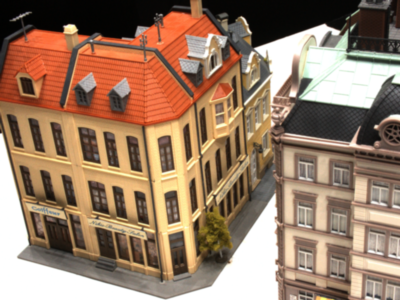}&
    \includegraphics[width=0.16\textwidth]{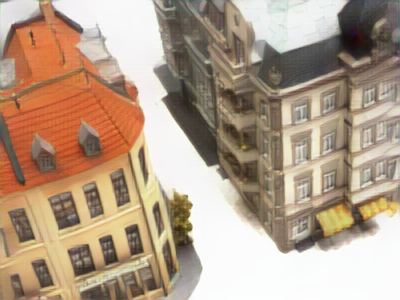}&
    \includegraphics[width=0.16\textwidth]{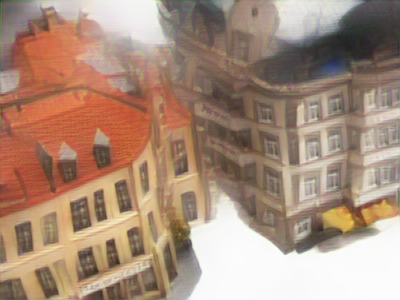}&
    \includegraphics[width=0.16\textwidth]{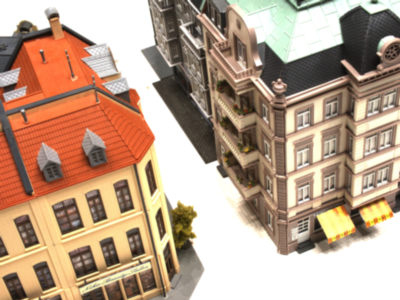}\\

    \includegraphics[width=0.16\textwidth]{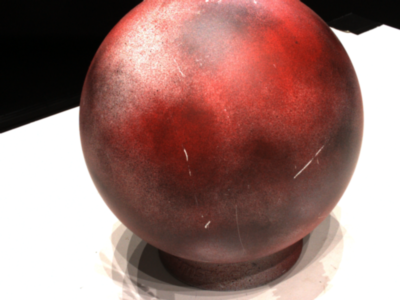}&
    \includegraphics[width=0.16\textwidth]{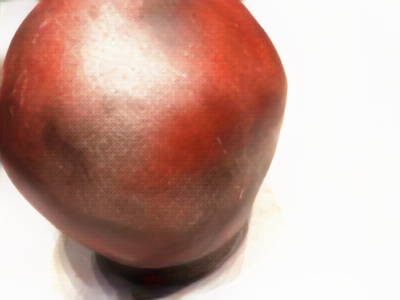}&
    \includegraphics[width=0.16\textwidth]{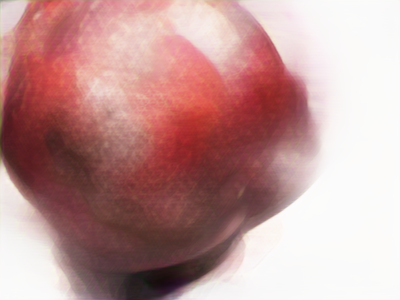}&
    \includegraphics[width=0.16\textwidth]{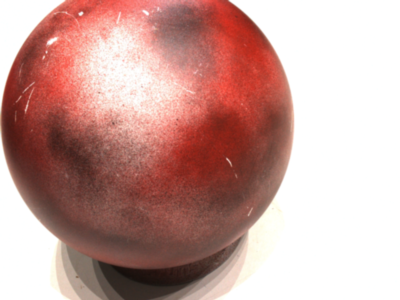}\\
    Input View & \FWD & w/o Transformer   & Target View \\
    \end{tabular}}

    \caption{
    \textbf{Ablation on Fusion Transformer.}
    We show results for \FWD with and without Transformed-based fusion. 
    }
    \label{fig:abl_transformer}
    \end{figure}

\noindent\textbf{Effects of Fusion Transformer.}
We design a model without Transformer, which concatenates point clouds across views into a bigger one for later rendering and refinement.
Its results  in \FWD settings are shown in Figure~\ref{fig:abl_transformer}.
The ablated version is vulnerable to inaccurate depths learned in unsupervised manner and synthesizes ``ghost objects'' since points with bad depths occlude other views' points.

We repeat the same ablation in \FWDrgbd settings, shown in Table~\ref{tab:design_ablation}, which settings give much better depth estimations with sensor depths.
The ablated model has notably worse results for all metrics,
indicating that the proposed method is not only powerful to tackle inaccurate depth estimations, but also fuse semantic features effectively.

\begin{table}[t]
    \centering
    \caption{
        {\bf Quantitative comparison on DTU real images}. 
    We compare our method with representatives of generalizable NeRF variants and IBR methods for image quality and rendering speed.
    Our method achieves significantly better speed-quality tradeoff, indicating the effectiveness and efficiency of our design.
    \hspace{\textwidth}
    $\dagger$~Unlike other methods, SynSin receives only one image as input.
    }
    \vspace{1mm}
    \resizebox{0.48\textwidth}{!}{
    \centering
    \begin{tabular}{p{0.9cm} p{0.9cm} p{2.78cm} p{0.9cm} p{0.9cm} p{0.9cm}p{0.6cm}}
    \toprule[2pt]
    Test & Train&Model & PSNR$\uparrow$& SSIM$\uparrow$ & LPIPS$\downarrow$ & FPS$\uparrow$ \\
    \midrule
    \multirow{5}{*}{RGB} & \multirow{5}{*}{RGB} &PixelNeRF~\cite{yu2020pixelnerf}   &\bf 19.24 & 0.687  &0.399 & 0.03\\
    &    &IBRNet~\cite{wang2021ibrnet}   &18.86  &\bf 0.695  &0.387  & 0.27\\
    &  &MVSNeRF~\cite{chen2021mvsnerf} &17.13  &0.611 & 0.444&  0.34 \\
    &  &SynSin~\cite{wiles2020synsin}~$\dagger$ &15.66 & 0.564 &0.388 &\bf 51.8\\
    &  &\bf \FWD & 17.42 & 0.598 &\bf 0.341 & 35.4\\
    \midrule
    \multirow{2}{*}{RGB} & \multirow{2}{*}{RGB-D} &PixelNeRF-DS~\cite{deng2021depth} & 19.87 & 0.710 &0.370 &0.03  \\
    &  &\bf \FWDds &\bf 20.15  &\bf 0.721 & \bf 0.259&\bf 35.4 \\
    \midrule
    \multirow{3}{*}{RGB-D}  
    & \multirow{3}{*}{RGB-D}  &Blending-R~\cite{DeepBlending2018}& 16.98 &0.661 & 0.351& 41.8\\
    & &FVS~\cite{riegler2020free} &15.92 &0.733 &0.267  &9.70 \\
    & &\bf \FWDrgbd &\bf 21.98  &\bf 0.791  &\bf 0.208  &\bf 43.2 \\
    \bottomrule [2 pt]
    \end{tabular}}
    \label{tab:dtu}
    \vspace{-6mm}
\end{table}

\noindent\textbf{Effects of View Dependent MLP.}
For ablation, we remove the view-dependent feature MLP and report its results in Table~\ref{tab:design_ablation}. 
Removing this module reduces model's ability to produce view-dependent appearances, leading to worse performance for all metrics.
We show more results in Supp.

\begin{figure}[t]
    \centering
    \resizebox{0.48\textwidth}{!}{
    \begin{tabular}{@{}c@{\hspace{0.5mm}}c@{\hspace{0.5mm}}c@{\hspace{0.5mm}}c@{\hspace{0.5 mm}}c@{\hspace{0.5mm}}c@{}}
    \includegraphics[width=0.16\textwidth]{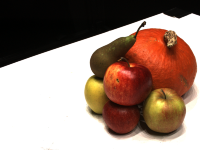}&
    \includegraphics[width=0.16\textwidth]{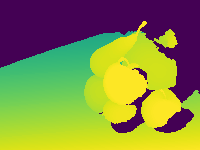}&
    \includegraphics[width=0.16\textwidth]{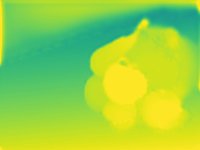}&
    \includegraphics[width=0.16\textwidth]{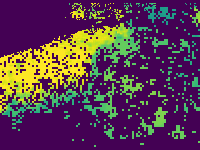}&
    \includegraphics[width=0.16\textwidth]{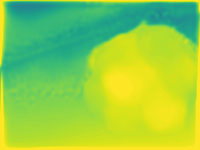}&
    \includegraphics[width=0.16\textwidth]{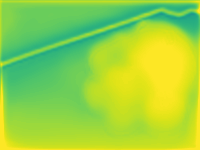}\\

    \includegraphics[width=0.16\textwidth]{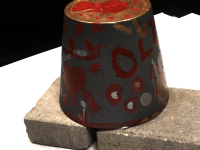}&
    \includegraphics[width=0.16\textwidth]{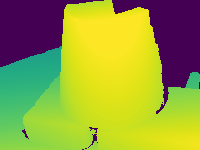}&
    \includegraphics[width=0.16\textwidth]{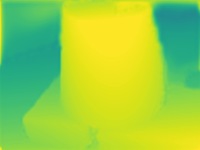}&
    \includegraphics[width=0.16\textwidth]{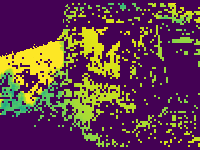}&
    \includegraphics[width=0.16\textwidth]{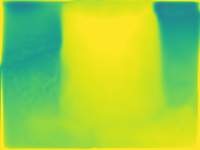}&
    \includegraphics[width=0.16\textwidth]{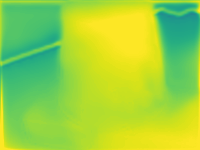}\\
     input image &  sensor depths &  \FWDrgbd & filtered MVS &  \FWDds  & \FWD \\

    \end{tabular}}

    \captionof{figure}{
    \textbf{Depth visualizations.}
    We visualize the normalized inverse depths involved in our method.
    Sensor depths are incomplete because of hardware limitations and MVS estimated depths are inaccurate, where many predictions have low confidence.
    This demonstrates the necessity of depth completion and refinement. 
    }
    \label{fig:depth}
    \vspace{-1mm}
    \end{figure}

\noindent\textbf{Depth Analysis and Ablations.}
We visualize depths in Figure~\ref{fig:depth}.
Estimating depths from sparse inputs is challenging and gives less accurate results because of the huge baselines between inputs. 
We show estimated depths from PatchmatchNet here, filtered based on the confidence scores.
Therefore, refinement is essential in our design to propagate multi-view geometry cues to the whole image. 
Our end-to-end model learns it by synthesis losses.

We ablate the depth network in Table~\ref{tab:depth_ablation} and report depth error $\delta_{3cm}$, which is the percentage of estimated depths within 3 cm of sensor depths.
MVS module is critical~(row 2) to give geometrically consistent depths. 
U-Net further refines depths and improves the synthesis quality~(row 3).
PatchmatchNet has its own shallow refinement layer,  already giving decent refinements.
Learning unsupervised MVS and NVS jointly from scratch is challenging ~(row 4),
and training depth network without supervision~\cite{8885975} first may give a good initialization for further jointly training.

\begin{table}[t!]
    \centering
    \caption{
    \textbf{Ablation Studies.}
    We show the effectiveness of Transformer Fusion and View-dependent MLP by ablation study on FWD-D.
    These designs improve synthesize quality noticeably while maintaining real-time rendering speed.}
    \vspace{1mm}
    \resizebox{0.48\textwidth}{!}{
    \begin{tabular}{p{3.5cm} p{0.8cm} p{0.8cm} p{0.8cm} p{0.8cm}}
    \toprule[2pt]
    Model & PSNR & SSIM & LPIPS & FPS\\
    \midrule
    Full model & \bf 21.98 & \bf 0.791& \bf 0.208 &43.2 \\
    w/o Transformer & 20.95 & 0.748 & 0.241 &\bf 48.4 \\
    w/o View dependence &21.16 & 0.769 &0.212 &44.0\\
    \bottomrule [2 pt]
    \end{tabular}}
    \label{tab:design_ablation}
    \vspace{-4mm}
\end{table}

\section{Conclusion}
\label{sec:conclusion}
We propose a real-time and generalizable method for NVS with sparse inputs by using explicit depths.
This method inherits the core idea of SynSin while extending it to multi-view input settings, which is more challenging. 
Our experiments show that estimating depths can give impressive results with a real-time speed, outperforming existing methods. 
Moreover, the proposed method could utilize sensor depths seamlessly and improve synthesis quality significantly. 
With the increasing availability of mobile depth sensors, we believe our method has exciting real-world 3D applications. 
We acknowledge there could be the potential for the technology to be used for negative purposes by nefarious actors, like synthesizing fake images for cheating.

There are also challenges and limitations yet to be explored.
1) Although using explicit depths gives tremendous speedups, it potentially inflicts depth reliance on our model.  
We designed a hybrid depth regressor to improve the quality of depth by combining MVS and single image depth estimations.
We also employed an effective fusion and refinement module to reduce the degrades caused by inaccurate depths. 
Despite these designs,   the depth estimator may still work poorly
in some challenging settings~(like very wide camera baselines), and it would influence the synthesis results.
Exploring other depth estimation methods like MiDaS~\cite{Ranftl2020} could be an interesting direction for future work.

2) The potential capacity of our method is not fully explored.
Like SynSin~\cite{wiles2020synsin}, our model~(depth/feature network and refinement module especially) is suitable and  beneficial from large-scale training data, 
while the DTU MVS dataset is not big enough and easy to overfit during training.
Evaluating our method on large-scale datasets like Hypersim~\cite{roberts2021hypersim} would potentially reveal more advantages of our model, which dataset is very challenging for NeRF-like methods.

3) Although our method gives more visually appealing results, our PSNR and SSIM are lower than NeRF-like methods. 
We hypothesize that our refinement module is not perfectly trained to decode RGB colors from feature vectors because of limited training data.
Also, tiny misalignments caused during the rendering process may also harm the PSNR, although it is not perceptually visible. 
\begin{table}[t]
    \centering
    \caption{
    \textbf{Depths network ablation and error.}
    We ablate depth network and compute $\delta_{3 cm}$ as error, which is the percentage of predicted depths within 3 cm of sensor depths.
    }
    \vspace{1mm}
    \resizebox{0.48\textwidth}{!}{
    \begin{tabular}{p{0.7cm} p{1.1cm} p{1.6cm} p{0.8cm} p{0.8cm}p{0.8cm} p{1cm} p{0.8cm}}
    \toprule[2pt]
     Test & Train &Model &PSNR & SSIM & LPIPS & $\delta_{3 cm}$\\
    \midrule
    RGB  & RGB-D & \FWDds&20.15 & 0.721 & 0.259 & 79.07 \\
    RGB & RGB-D & -w/o MVS &16.69 &0.594 &0.357 & 61.62 \\
    RGB & RGB-D & -w/o U-Net &19.10 &0.702 &0.285 &73.62\\
    RGB  & RGB & \FWD &17.42 &0.598 &0.341 & 54.27  \\
    \bottomrule [2 pt]
    \end{tabular}}
    \label{tab:depth_ablation}
    \vspace{-4mm}
\end{table}

\noindent\textbf{Acknowledgement.}
Toyota Research Institute provided funds to support this work.
We thank Dandan Shan, Hao Ouyang, Jiaxin Xie, Linyi Jin, Shengyi Qian for helpful discussions.

\clearpage

{\small
\bibliographystyle{ieee_fullname}
\bibliography{egbib}

\begin{thebibliography}{10}\itemsep=-1pt

\bibitem{aanaes2016large}
Henrik Aan{\ae}s, Rasmus~Ramsb{\o}l Jensen, George Vogiatzis, Engin Tola, and
  Anders~Bjorholm Dahl.
\newblock Large-scale data for multiple-view stereopsis.
\newblock {\em IJCV}, 120(2):153--168, 2016.

\bibitem{yu2021plenoxels}
{Alex Yu and Sara Fridovich-Keil}, Matthew Tancik, Qinhong Chen, Benjamin
  Recht, and Angjoo Kanazawa.
\newblock Plenoxels: Radiance fields without neural networks, 2021.

\bibitem{brock2018large}
Andrew Brock, Jeff Donahue, and Karen Simonyan.
\newblock Large scale {GAN} training for high fidelity natural image synthesis.
\newblock In {\em ICLR}, 2019.

\bibitem{carion2020end}
Nicolas Carion, Francisco Massa, Gabriel Synnaeve, Nicolas Usunier, Alexander
  Kirillov, and Sergey Zagoruyko.
\newblock End-to-end object detection with transformers.
\newblock In {\em ECCV}, pages 213--229. Springer, 2020.

\bibitem{Matterport3D}
Angel Chang, Angela Dai, Thomas Funkhouser, Maciej Halber, Matthias Niessner,
  Manolis Savva, Shuran Song, Andy Zeng, and Yinda Zhang.
\newblock Matterport3d: Learning from rgb-d data in indoor environments.
\newblock {\em International Conference on 3D Vision (3DV)}, 2017.

\bibitem{chang2015shapenet}
Angel~X Chang, Thomas Funkhouser, Leonidas Guibas, Pat Hanrahan, Qixing Huang,
  Zimo Li, Silvio Savarese, Manolis Savva, Shuran Song, Hao Su, et~al.
\newblock Shapenet: An information-rich 3d model repository.
\newblock {\em Technical Report arXiv:1512.03012}, 2015.

\bibitem{chaurasia2013depth}
Gaurav Chaurasia, Sylvain Duchene, Olga Sorkine-Hornung, and George Drettakis.
\newblock Depth synthesis and local warps for plausible image-based navigation.
\newblock {\em ACM Transactions on Graphics (TOG)}, 32(3):1--12, 2013.

\bibitem{TensoRF}
Anpei Chen, Zexiang Xu, Andreas Geiger, , Jingyi Yu, and Hao Su.
\newblock Tensorf: Tensorial radiance fields, 2022.

\bibitem{chen2021mvsnerf}
Anpei Chen, Zexiang Xu, Fuqiang Zhao, Xiaoshuai Zhang, Fanbo Xiang, Jingyi Yu,
  and Hao Su.
\newblock M{VSN}e{RF}: Fast generalizable radiance field reconstruction from
  multi-view stereo.
\newblock In {\em Proceedings of the IEEE/CVF International Conference on
  Computer Vision}, pages 14124--14133, 2021.

\bibitem{NIPS2016_0deb1c54}
Weifeng Chen, Zhao Fu, Dawei Yang, and Jia Deng.
\newblock Single-image depth perception in the wild.
\newblock In {\em NeurIPS}, volume~29, 2016.

\bibitem{chen2019learning}
Wenzheng Chen, Jun Gao, Huan Ling, Edward~J Smith, Jaakko Lehtinen, Alec
  Jacobson, and Sanja Fidler.
\newblock Learning to predict 3d objects with an interpolation-based
  differentiable renderer.
\newblock 2019.

\bibitem{choi2019extreme}
Inchang Choi, Orazio Gallo, Alejandro Troccoli, Min~H Kim, and Jan Kautz.
\newblock Extreme view synthesis.
\newblock In {\em ICCV}, pages 7781--7790, 2019.

\bibitem{dai2017scannet}
Angela Dai, Angel~X. Chang, Manolis Savva, Maciej Halber, Thomas Funkhouser,
  and Matthias Nie{\ss}ner.
\newblock Scannet: Richly-annotated 3d reconstructions of indoor scenes.
\newblock In {\em Proc. Computer Vision and Pattern Recognition (CVPR), IEEE},
  2017.

\bibitem{8885975}
Yuchao Dai, Zhidong Zhu, Zhibo Rao, and Bo Li.
\newblock Mvs2: Deep unsupervised multi-view stereo with multi-view symmetry.
\newblock In {\em 2019 International Conference on 3D Vision (3DV)}, pages
  1--8, 2019.

\bibitem{debevec1996modeling}
Paul~E Debevec, Camillo~J Taylor, and Jitendra Malik.
\newblock Modeling and rendering architecture from photographs: A hybrid
  geometry-and image-based approach.
\newblock In {\em Proceedings of the 23rd annual conference on Computer
  graphics and interactive techniques}, pages 11--20, 1996.

\bibitem{deng2021depth}
Kangle Deng, Andrew Liu, Jun-Yan Zhu, and Deva Ramanan.
\newblock Depth-supervised nerf: Fewer views and faster training for free.
\newblock {\em arXiv preprint arXiv:2107.02791}, 2021.

\bibitem{dosovitskiy2020image}
Alexey Dosovitskiy, Lucas Beyer, Alexander Kolesnikov, Dirk Weissenborn,
  Xiaohua Zhai, Thomas Unterthiner, Mostafa Dehghani, Matthias Minderer, Georg
  Heigold, Sylvain Gelly, et~al.
\newblock An image is worth 16x16 words: Transformers for image recognition at
  scale.
\newblock {\em arXiv preprint arXiv:2010.11929}, 2020.

\bibitem{garbin2021fastnerf}
Stephan~J Garbin, Marek Kowalski, Matthew Johnson, Jamie Shotton, and Julien
  Valentin.
\newblock Fastnerf: High-fidelity neural rendering at 200fps.
\newblock In {\em Proceedings of the IEEE/CVF International Conference on
  Computer Vision}, pages 14346--14355, 2021.

\bibitem{genova2020local}
Kyle Genova, Forrester Cole, Avneesh Sud, Aaron Sarna, and Thomas Funkhouser.
\newblock Local deep implicit functions for 3d shape.
\newblock In {\em Proceedings of the IEEE/CVF Conference on Computer Vision and
  Pattern Recognition}, pages 4857--4866, 2020.

\bibitem{gortler1996lumigraph}
Steven~J Gortler, Radek Grzeszczuk, Richard Szeliski, and Michael~F Cohen.
\newblock The lumigraph.
\newblock In {\em Proceedings of the 23rd annual conference on Computer
  graphics and interactive techniques}, pages 43--54, 1996.

\bibitem{guo2021fast}
Pengsheng Guo, Miguel~Angel Bautista, Alex Colburn, Liang Yang, Daniel
  Ulbricht, Joshua~M Susskind, and Qi Shan.
\newblock Fast and {E}xplicit {N}eural {V}iew {S}ynthesis.
\newblock In {\em Proceedings of the IEEE/CVF Winter Conference on Applications
  of Computer Vision}, pages 3791--3800, 2022.

\bibitem{he2016deep}
Kaiming He, Xiangyu Zhang, Shaoqing Ren, and Jian Sun.
\newblock Deep residual learning for image recognition.
\newblock In {\em CVPR}, pages 770--778, 2016.

\bibitem{hedman2018instant}
Peter Hedman and Johannes Kopf.
\newblock Instant 3d photography.
\newblock {\em TOG}, 37(4):1--12, 2018.

\bibitem{hedman2018deep}
Peter Hedman, Julien Philip, True Price, Jan-Michael Frahm, George Drettakis,
  and Gabriel Brostow.
\newblock Deep blending for free-viewpoint image-based rendering.
\newblock {\em ACM Transactions on Graphics (TOG)}, 37(6):1--15, 2018.

\bibitem{DeepBlending2018}
Peter Hedman, Julien Philip, True Price, Jan-Michael Frahm, George Drettakis,
  and Gabriel Brostow.
\newblock Deep blending for free-viewpoint image-based rendering.
\newblock 37(6):257:1--257:15, 2018.

\bibitem{hedman2016scalable}
Peter Hedman, Tobias Ritschel, George Drettakis, and Gabriel Brostow.
\newblock Scalable inside-out image-based rendering.
\newblock {\em ToG}, 35(6):1--11, 2016.

\bibitem{hedman2021baking}
Peter Hedman, Pratul~P Srinivasan, Ben Mildenhall, Jonathan~T Barron, and Paul
  Debevec.
\newblock Baking neural radiance fields for real-time view synthesis.
\newblock In {\em Proceedings of the IEEE/CVF International Conference on
  Computer Vision}, pages 5875--5884, 2021.

\bibitem{hu2020worldsheet}
Ronghang Hu, Nikhila Ravi, Alexander~C Berg, and Deepak Pathak.
\newblock Worldsheet: Wrapping the world in a 3d sheet for view synthesis from
  a single image.
\newblock In {\em Proceedings of the IEEE/CVF International Conference on
  Computer Vision}, pages 12528--12537, 2021.

\bibitem{Huang2018}
Po-Han Huang, Kevin Matzen, Johannes Kopf, Narendra Ahuja, and Jia-Bin Huang.
\newblock Deepmvs: Learning multi-view stereopsis.
\newblock {\em Conference on Computer Vision and Pattern Recognition (CVPR)},
  2018.

\bibitem{huang2020lstm}
Rui Huang, Wanyue Zhang, Abhijit Kundu, Caroline Pantofaru, David~A Ross,
  Thomas Funkhouser, and Alireza Fathi.
\newblock An lstm approach to temporal 3d object detection in lidar point
  clouds.
\newblock {\em arXiv preprint arXiv:2007.12392}, 2020.

\bibitem{jain2021putting}
Ajay Jain, Matthew Tancik, and Pieter Abbeel.
\newblock Putting nerf on a diet: Semantically consistent few-shot view
  synthesis.
\newblock In {\em Proceedings of the IEEE/CVF International Conference on
  Computer Vision}, pages 5885--5894, 2021.

\bibitem{jatavallabhula2019kaolin}
Krishna~Murthy Jatavallabhula, Edward Smith, Jean-Francois Lafleche,
  Clement~Fuji Tsang, Artem Rozantsev, Wenzheng Chen, Tommy Xiang, Rev
  Lebaredian, and Sanja Fidler.
\newblock Kaolin: A pytorch library for accelerating 3d deep learning research.
\newblock {\em arXiv preprint arXiv:1911.05063}, 2019.

\bibitem{jensen2014large}
Rasmus Jensen, Anders Dahl, George Vogiatzis, Engil Tola, and Henrik Aan{\ae}s.
\newblock Large scale multi-view stereopsis evaluation.
\newblock In {\em 2014 IEEE Conference on Computer Vision and Pattern
  Recognition}, pages 406--413. IEEE, 2014.

\bibitem{jiang2020local}
Chiyu Jiang, Avneesh Sud, Ameesh Makadia, Jingwei Huang, Matthias Nie{\ss}ner,
  Thomas Funkhouser, et~al.
\newblock Local implicit grid representations for 3d scenes.
\newblock In {\em Proceedings of the IEEE/CVF Conference on Computer Vision and
  Pattern Recognition}, pages 6001--6010, 2020.

\bibitem{jiang2020sdfdiff}
Yue Jiang, Dantong Ji, Zhizhong Han, and Matthias Zwicker.
\newblock Sdfdiff: Differentiable rendering of signed distance fields for 3d
  shape optimization.
\newblock In {\em Proceedings of the IEEE/CVF Conference on Computer Vision and
  Pattern Recognition}, pages 1251--1261, 2020.

\bibitem{karras2020analyzing}
Tero Karras, Samuli Laine, Miika Aittala, Janne Hellsten, Jaakko Lehtinen, and
  Timo Aila.
\newblock Analyzing and improving the image quality of stylegan.
\newblock In {\em CVPR}, pages 8110--8119, 2020.

\bibitem{kingma2014adam}
Diederik~P Kingma and Jimmy Ba.
\newblock Adam: A method for stochastic optimization.
\newblock 2015.

\bibitem{ledig2017photo}
C. {Ledig}, L. {Theis}, F. {Huszár}, J. {Caballero}, A. {Cunningham}, A.
  {Acosta}, A. {Aitken}, A. {Tejani}, J. {Totz}, Z. {Wang}, and W. {Shi}.
\newblock Photo-realistic single image super-resolution using a generative
  adversarial network.
\newblock In {\em CVPR}, pages 105--114, 2017.

\bibitem{levoy1996light}
Marc Levoy and Pat Hanrahan.
\newblock Light field rendering.
\newblock In {\em Proceedings of the 23rd annual conference on Computer
  graphics and interactive techniques}, pages 31--42, 1996.

\bibitem{liu2020infinite}
Andrew Liu, Richard Tucker, Varun Jampani, Ameesh Makadia, Noah Snavely, and
  Angjoo Kanazawa.
\newblock Infinite nature: Perpetual view generation of natural scenes from a
  single image.
\newblock In {\em Proceedings of the IEEE/CVF International Conference on
  Computer Vision}, pages 14458--14467, 2021.

\bibitem{liu2020neural}
Lingjie Liu, Jiatao Gu, Kyaw Zaw~Lin, Tat-Seng Chua, and Christian Theobalt.
\newblock Neural sparse voxel fields.
\newblock {\em Advances in Neural Information Processing Systems},
  33:15651--15663, 2020.

\bibitem{liu2019soft}
Shichen Liu, Weikai Chen, Tianye Li, and Hao Li.
\newblock Soft rasterizer: Differentiable rendering for unsupervised
  single-view mesh reconstruction.
\newblock {\em ICCV}, 2019.

\bibitem{liu2020dist}
Shaohui Liu, Yinda Zhang, Songyou Peng, Boxin Shi, Marc Pollefeys, and Zhaopeng
  Cui.
\newblock Dist: Rendering deep implicit signed distance function with
  differentiable sphere tracing.
\newblock In {\em Proceedings of the IEEE/CVF Conference on Computer Vision and
  Pattern Recognition}, pages 2019--2028, 2020.

\bibitem{lombardi2019neural}
Stephen Lombardi, Tomas Simon, Jason Saragih, Gabriel Schwartz, Andreas
  Lehrmann, and Yaser Sheikh.
\newblock Neural volumes: Learning dynamic renderable volumes from images.
\newblock {\em arXiv preprint arXiv:1906.07751}, 2019.

\bibitem{luo2020consistent}
Xuan Luo, Jia-Bin Huang, Richard Szeliski, Kevin Matzen, and Johannes Kopf.
\newblock Consistent video depth estimation.
\newblock {\em ACM Transactions on Graphics (TOG)}, 39(4):71--1, 2020.

\bibitem{martinbrualla2020nerfw}
Ricardo Martin-Brualla, Noha Radwan, Mehdi S.~M. Sajjadi, Jonathan~T. Barron,
  Alexey Dosovitskiy, and Daniel Duckworth.
\newblock {NeRF in the Wild: Neural Radiance Fields for Unconstrained Photo
  Collections}.
\newblock In {\em CVPR}, 2021.

\bibitem{mescheder2019occupancy}
Lars Mescheder, Michael Oechsle, Michael Niemeyer, Sebastian Nowozin, and
  Andreas Geiger.
\newblock Occupancy networks: Learning 3d reconstruction in function space.
\newblock In {\em Proceedings of the IEEE/CVF Conference on Computer Vision and
  Pattern Recognition}, pages 4460--4470, 2019.

\bibitem{mildenhall2020nerf}
Ben Mildenhall, Pratul~P. Srinivasan, Matthew Tancik, Jonathan~T. Barron, Ravi
  Ramamoorthi, and Ren Ng.
\newblock Nerf: Representing scenes as neural radiance fields for view
  synthesis.
\newblock In {\em ECCV}, 2020.

\bibitem{mueller2022instant}
Thomas M\"uller, Alex Evans, Christoph Schied, and Alexander Keller.
\newblock Instant neural graphics primitives with a multiresolution hash
  encoding.
\newblock {\em arXiv:2201.05989}, Jan. 2022.

\bibitem{najibi2020dops}
Mahyar Najibi, Guangda Lai, Abhijit Kundu, Zhichao Lu, Vivek Rathod, Thomas
  Funkhouser, Caroline Pantofaru, David Ross, Larry~S Davis, and Alireza Fathi.
\newblock Dops: learning to detect 3d objects and predict their 3d shapes.
\newblock In {\em CVPR}, pages 11913--11922, 2020.

\bibitem{neff2021donerf}
Thomas Neff, Pascal Stadlbauer, Mathias Parger, Andreas Kurz, Joerg~H Mueller,
  Chakravarty R~Alla Chaitanya, Anton Kaplanyan, and Markus Steinberger.
\newblock {DON}e{RF}: Towards real-time rendering of compact neural radiance
  fields using depth oracle networks.
\newblock In {\em Computer Graphics Forum}, volume~40, pages 45--59. Wiley
  Online Library, 2021.

\bibitem{niemeyer2020differentiable}
Michael Niemeyer, Lars Mescheder, Michael Oechsle, and Andreas Geiger.
\newblock Differentiable volumetric rendering: Learning implicit 3d
  representations without 3d supervision.
\newblock In {\em Proceedings of the IEEE/CVF Conference on Computer Vision and
  Pattern Recognition}, pages 3504--3515, 2020.

\bibitem{DVR}
Michael Niemeyer, Lars Mescheder, Michael Oechsle, and Andreas Geiger.
\newblock Differentiable volumetric rendering: Learning implicit 3d
  representations without 3d supervision.
\newblock In {\em Proc. IEEE Conf. on Computer Vision and Pattern Recognition
  (CVPR)}, 2020.

\bibitem{novotny2019perspectivenet}
David Novotny, Ben Graham, and Jeremy Reizenstein.
\newblock Perspectivenet: A scene-consistent image generator for new view
  synthesis in real indoor environments.
\newblock {\em Advances in Neural Information Processing Systems},
  32:7601--7612, 2019.

\bibitem{park2019deepsdf}
Jeong~Joon Park, Peter Florence, Julian Straub, Richard Newcombe, and Steven
  Lovegrove.
\newblock Deepsdf: Learning continuous signed distance functions for shape
  representation.
\newblock In {\em Proceedings of the IEEE/CVF Conference on Computer Vision and
  Pattern Recognition}, pages 165--174, 2019.

\bibitem{park2020nerfies}
Keunhong Park, Utkarsh Sinha, Jonathan~T Barron, Sofien Bouaziz, Dan~B Goldman,
  Steven~M Seitz, and Ricardo Martin-Brualla.
\newblock Nerfies: Deformable neural radiance fields.
\newblock In {\em Proceedings of the IEEE/CVF International Conference on
  Computer Vision}, pages 5865--5874, 2021.

\bibitem{penner2017soft}
Eric Penner and Li Zhang.
\newblock Soft 3d reconstruction for view synthesis.
\newblock {\em ToG}, 36(6):1--11, 2017.

\bibitem{qi2017pointnet}
Charles~R Qi, Hao Su, Kaichun Mo, and Leonidas~J Guibas.
\newblock Pointnet: Deep learning on point sets for 3d classification and
  segmentation.
\newblock In {\em Proceedings of the IEEE conference on computer vision and
  pattern recognition}, pages 652--660, 2017.

\bibitem{Ranftl2020}
Ren\'{e} Ranftl, Katrin Lasinger, David Hafner, Konrad Schindler, and Vladlen
  Koltun.
\newblock Towards robust monocular depth estimation: Mixing datasets for
  zero-shot cross-dataset transfer.
\newblock {\em IEEE Transactions on Pattern Analysis and Machine Intelligence
  (TPAMI)}, 2020.

\bibitem{ravi2020accelerating}
Nikhila Ravi, Jeremy Reizenstein, David Novotny, Taylor Gordon, Wan-Yen Lo,
  Justin Johnson, and Georgia Gkioxari.
\newblock Accelerating 3d deep learning with pytorch3d.
\newblock {\em arXiv preprint arXiv:2007.08501}, 2020.

\bibitem{riegler2020free}
Gernot Riegler and Vladlen Koltun.
\newblock Free view synthesis.
\newblock In {\em ECCV}, pages 623--640. Springer, 2020.

\bibitem{Riegler2021SVS}
Gernot Riegler and Vladlen Koltun.
\newblock Stable view synthesis.
\newblock In {\em Proceedings of the IEEE Conference on Computer Vision and
  Pattern Recognition}, 2021.

\bibitem{roberts2021hypersim}
Mike Roberts, Jason Ramapuram, Anurag Ranjan, Atulit Kumar, Miguel~Angel
  Bautista, Nathan Paczan, Russ Webb, and Joshua~M Susskind.
\newblock Hypersim: A photorealistic synthetic dataset for holistic indoor
  scene understanding.
\newblock In {\em Proceedings of the IEEE/CVF International Conference on
  Computer Vision}, pages 10912--10922, 2021.

\bibitem{Rockwell2021}
Chris Rockwell, David~F. Fouhey, and Justin Johnson.
\newblock Pixelsynth: Generating a 3d-consistent experience from a single
  image.
\newblock In {\em ICCV}, 2021.

\bibitem{rombach2021geometry}
Robin Rombach, Patrick Esser, and Bj{\"o}rn Ommer.
\newblock Geometry-free view synthesis: Transformers and no 3d priors.
\newblock In {\em Proceedings of the IEEE/CVF International Conference on
  Computer Vision}, pages 14356--14366, 2021.

\bibitem{schoenberger2016mvs}
Johannes~Lutz Sch\"{o}nberger, Enliang Zheng, Marc Pollefeys, and Jan-Michael
  Frahm.
\newblock Pixelwise view selection for unstructured multi-view stereo.
\newblock In {\em European Conference on Computer Vision (ECCV)}, 2016.

\bibitem{shih20203d}
Meng-Li Shih, Shih-Yang Su, Johannes Kopf, and Jia-Bin Huang.
\newblock 3d photography using context-aware layered depth inpainting.
\newblock In {\em Proceedings of the IEEE/CVF Conference on Computer Vision and
  Pattern Recognition}, pages 8028--8038, 2020.

\bibitem{silberman2012indoor}
Nathan Silberman, Derek Hoiem, Pushmeet Kohli, and Rob Fergus.
\newblock Indoor segmentation and support inference from rgbd images.
\newblock In {\em ECCV}, pages 746--760, 2012.

\bibitem{sitzmann2019deepvoxels}
Vincent Sitzmann, Justus Thies, Felix Heide, Matthias Nie{\ss}ner, Gordon
  Wetzstein, and Michael Zollhofer.
\newblock Deepvoxels: Learning persistent 3d feature embeddings.
\newblock In {\em CVPR}, pages 2437--2446, 2019.

\bibitem{NEURIPS2019_b5dc4e5d}
Vincent Sitzmann, Michael Zollhoefer, and Gordon Wetzstein.
\newblock Scene representation networks: Continuous 3d-structure-aware neural
  scene representations.
\newblock In H. Wallach, H. Larochelle, A. Beygelzimer, F. d\textquotesingle
  Alch\'{e}-Buc, E. Fox, and R. Garnett, editors, {\em Advances in Neural
  Information Processing Systems}, volume~32. Curran Associates, Inc., 2019.

\bibitem{song2015sun}
Shuran Song, Samuel~P Lichtenberg, and Jianxiong Xiao.
\newblock Sun rgb-d: A rgb-d scene understanding benchmark suite.
\newblock In {\em CVPR}, pages 567--576, 2015.

\bibitem{srinivasan2019pushing}
Pratul~P Srinivasan, Richard Tucker, Jonathan~T Barron, Ravi Ramamoorthi, Ren
  Ng, and Noah Snavely.
\newblock Pushing the boundaries of view extrapolation with multiplane images.
\newblock In {\em CVPR}, pages 175--184, 2019.

\bibitem{stelzner2021decomposing}
Karl Stelzner, Kristian Kersting, and Adam~R Kosiorek.
\newblock Decomposing 3d scenes into objects via unsupervised volume
  segmentation.
\newblock {\em arXiv preprint arXiv:2104.01148}, 2021.

\bibitem{tatarchenko2016multi}
Maxim Tatarchenko, Alexey Dosovitskiy, and Thomas Brox.
\newblock Multi-view 3d models from single images with a convolutional network.
\newblock In {\em European Conference on Computer Vision}, pages 322--337.
  Springer, 2016.

\bibitem{trevithick2020grf}
Alex Trevithick and Bo Yang.
\newblock Grf: Learning a general radiance field for 3d representation and
  rendering.
\newblock In {\em Proceedings of the IEEE/CVF International Conference on
  Computer Vision}, pages 15182--15192, 2021.

\bibitem{46201}
Ashish Vaswani, Noam Shazeer, Niki Parmar, Jakob Uszkoreit, Llion Jones,
  Aidan~N. Gomez, Lukasz Kaiser, and Illia Polosukhin.
\newblock Attention is all you need.
\newblock 2017.

\bibitem{wang2020patchmatchnet}
Fangjinhua Wang, Silvano Galliani, Christoph Vogel, Pablo Speciale, and Marc
  Pollefeys.
\newblock Patchmatchnet: Learned multi-view patchmatch stereo.
\newblock In {\em Proceedings of the IEEE/CVF Conference on Computer Vision and
  Pattern Recognition}, pages 14194--14203, 2021.

\bibitem{wang2021ibrnet}
Qianqian Wang, Zhicheng Wang, Kyle Genova, Pratul~P Srinivasan, Howard Zhou,
  Jonathan~T Barron, Ricardo Martin-Brualla, Noah Snavely, and Thomas
  Funkhouser.
\newblock Ibrnet: Learning multi-view image-based rendering.
\newblock In {\em Proceedings of the IEEE/CVF Conference on Computer Vision and
  Pattern Recognition}, pages 4690--4699, 2021.

\bibitem{wang2018high}
Ting-Chun Wang, Ming-Yu Liu, Jun-Yan Zhu, Andrew Tao, Jan Kautz, and Bryan
  Catanzaro.
\newblock High-resolution image synthesis and semantic manipulation with
  conditional gans.
\newblock In {\em CVPR}, pages 8798--8807, 2018.

\bibitem{wang2004image}
Zhou Wang, Alan~C Bovik, Hamid~R Sheikh, and Eero~P Simoncelli.
\newblock Image quality assessment: from error visibility to structural
  similarity.
\newblock {\em IEEE transactions on image processing}, 13(4):600--612, 2004.

\bibitem{wang2021nerf}
Zirui Wang, Shangzhe Wu, Weidi Xie, Min Chen, and Victor~Adrian Prisacariu.
\newblock Nerf--: Neural radiance fields without known camera parameters.
\newblock {\em arXiv preprint arXiv:2102.07064}, 2021.

\bibitem{wiles2020synsin}
Olivia Wiles, Georgia Gkioxari, Richard Szeliski, and Justin Johnson.
\newblock Synsin: End-to-end view synthesis from a single image.
\newblock In {\em Proceedings of the IEEE/CVF Conference on Computer Vision and
  Pattern Recognition}, pages 7467--7477, 2020.

\bibitem{wizadwongsa2021nex}
Suttisak Wizadwongsa, Pakkapon Phongthawee, Jiraphon Yenphraphai, and Supasorn
  Suwajanakorn.
\newblock Nex: Real-time view synthesis with neural basis expansion.
\newblock In {\em Proceedings of the IEEE/CVF Conference on Computer Vision and
  Pattern Recognition}, pages 8534--8543, 2021.

\bibitem{xian2021space}
Wenqi Xian, Jia-Bin Huang, Johannes Kopf, and Changil Kim.
\newblock Space-time neural irradiance fields for free-viewpoint video.
\newblock In {\em Proceedings of the IEEE/CVF Conference on Computer Vision and
  Pattern Recognition}, pages 9421--9431, 2021.

\bibitem{9341659}
Jiaxin Xie, Chenyang Lei, Zhuwen Li, Li~Erran Li, and Qifeng Chen.
\newblock Video depth estimation by fusing flow-to-depth proposals.
\newblock In {\em 2020 IEEE/RSJ International Conference on Intelligent Robots
  and Systems (IROS)}, pages 10100--10107, 2020.

\bibitem{yang2017high}
Chao Yang, Xin Lu, Zhe Lin, Eli Shechtman, Oliver Wang, and Hao Li.
\newblock High-resolution image inpainting using multi-scale neural patch
  synthesis.
\newblock In {\em CVPR}, pages 6721--6729, 2017.

\bibitem{yao2018mvsnet}
Yao Yao, Zixin Luo, Shiwei Li, Tian Fang, and Long Quan.
\newblock Mvsnet: Depth inference for unstructured multi-view stereo.
\newblock In {\em Proceedings of the European Conference on Computer Vision
  (ECCV)}, pages 767--783, 2018.

\bibitem{yu2021plenoctrees}
Alex Yu, Ruilong Li, Matthew Tancik, Hao Li, Ren Ng, and Angjoo Kanazawa.
\newblock {PlenOctrees} for real-time rendering of neural radiance fields.
\newblock In {\em arXiv}, 2021.

\bibitem{yu2020pixelnerf}
Alex Yu, Vickie Ye, Matthew Tancik, and Angjoo Kanazawa.
\newblock pixelnerf: Neural radiance fields from one or few images.
\newblock In {\em Proceedings of the IEEE/CVF Conference on Computer Vision and
  Pattern Recognition}, pages 4578--4587, 2021.

\bibitem{yu2018generative}
Jiahui Yu, Zhe Lin, Jimei Yang, Xiaohui Shen, Xin Lu, and Thomas~S Huang.
\newblock Generative image inpainting with contextual attention.
\newblock In {\em CVPR}, pages 5505--5514, 2018.

\bibitem{zhang2019self}
Han Zhang, Ian Goodfellow, Dimitris Metaxas, and Augustus Odena.
\newblock Self-attention generative adversarial networks.
\newblock In {\em ICML}, pages 7354--7363. PMLR, 2019.

\bibitem{zhang2020nerf++}
Kai Zhang, Gernot Riegler, Noah Snavely, and Vladlen Koltun.
\newblock Nerf++: Analyzing and improving neural radiance fields.
\newblock {\em arXiv preprint arXiv:2010.07492}, 2020.

\bibitem{zhang2018unreasonable}
Richard Zhang, Phillip Isola, Alexei~A Efros, Eli Shechtman, and Oliver Wang.
\newblock The unreasonable effectiveness of deep features as a perceptual
  metric.
\newblock In {\em CVPR}, pages 586--595, 2018.

\bibitem{zhou2018stereo}
Tinghui Zhou, Richard Tucker, John Flynn, Graham Fyffe, and Noah Snavely.
\newblock Stereo magnification: Learning view synthesis using multiplane
  images.
\newblock In {\em SIGGRAPH}, 2018.

\bibitem{CycleGAN2017}
Jun-Yan Zhu, Taesung Park, Phillip Isola, and Alexei~A Efros.
\newblock Unpaired image-to-image translation using cycle-consistent
  adversarial networks.
\newblock In {\em ICCV}, 2017.

\end{thebibliography}
}

\clearpage

We show more visualizations and comparisons in the supplementary and attached videos.
Please watch the results videos for more results.

We also provide implementation details, license of other methods and datasets, as well as other experiments results.
See the following contents label for more information.
{\hypersetup{linkcolor=black}
\tableofcontents }

\section{Model Architectures and Training details}

As stated in the paper, our model consists of spatial feature network $f$, depth network $d$, view-dependent feature MLP $\psi$,
neural point cloud renderer $\pi$, fusion transformer $T$ and refinement module $R$.
We show architecture details.

\par \noindent  \textbf{Spatial Feature Network $f$.}
The spatial feature network $f$ contains 8 ResNet blocks, with output channels of 32, 32, 32, 64, 64, 64,64, 61 and no downsampling.
Each ResNet block utilizes 3 $\times$ 3 /stride 1/padding 1 convolution followed by instance norm and ReLU.
Similar to \cite{brock2018large,wiles2020synsin}, spectral normalization is utilized for each convolution for stable training.
The input images are padded by 0 to fit network.

\par \noindent \textbf{Depth Network $d$.}
We utilize a classical U-Net for depth refinement, consisting 8 downsampling blocks and 8 upsampling blocks.
We pad constant zero to make the input have feasible shape.
For PatchMatchNet~\cite{wang2020patchmatchnet} to estimate the initial depths, 
we follow the original pipeline, in which we downsample the input images into 4 scales and predict the depths in a coarse-to-fine manner.

\par \noindent \textbf{View-dependent feature MLP $\psi$.}
The relative view direction change vector is first passed through a two-layer MLP without 16 and 32 output features perspectively to get a 32-dims feature embedding.
Then this 32 dims feature embedding is concatenated with the original 64 dims feature vector and passed through another two-layer MLP with output 64 and 64 output features.
We use ReLU as activation function following MLP and no normalization layers.

\par \noindent \textbf{Neural point cloud renderer $\pi$.}
The point cloud renderer is implemented in Pytorch3D~\cite{ravi2020accelerating},
which takes a point cloud and pose $P$ and project it to a 150 $\times$ 200 feature map, where each pixel has 64-dims feature.
The renderer fills zero with 64-dims for pixels which are invisible.

The blending weight $\alpha$ of the 3D point $x$ for pixel $l$ is
\begin{equation}
    \alpha = 1 - \text{torch.clamp}(\sqrt{\frac{s}{r^2}}, \text{min}=1e-3, \text{max}=1.0),
\end{equation}
where $s$ is the Euclidean distance between point $x$'s splatted center and pixel $l$; $r$ is the radius of point $x$ in NDC coordinate.
We set $r=1.5$ pixels in our experiments.
To render the value of each pixel, we employ alpha-compositing to blend all feasible points.
The rendered feature $F_{l}$ of pixel $l$ is:
\begin{equation}
    F_{l} = \sum_{i=1}^{K}\alpha_i F_i \prod_{j=1}^{i-1}(1-\alpha_j)
\end{equation}
where $F_i$, $\alpha_i$ is the feature and alpha of 3D point $i$.
The rendering is based on points' depths and we blend the Top K points with the nearest depths.
We use $K=16$ in our paper, meaning we blend at most 16 points to get the results of one pixel.

We compute $D_l$, the depth of pixel $l$ using the blending weights by:
    \begin{equation}
        D_l = \sum_{i=1}^K \alpha_i d_i
    \end{equation}
where $d_i$ is the depth of point $i$.
\par \noindent \textbf{Fusion transformer $T$.}
The Key and Value are feature vectors from multiple views, with the shape as NView$ \times N \times C$,
where NView is the number of input views~(3 in our experiments),
$N$ is the batch size, which is $H \times W$, 
and $C$ is the feature dimension, which is 64.
The query is a learnable token with the shape as $1 \times C$, and expanded to N batch.
The transformer is a multi-head attention with 4 heads, projecting key into 16 dims embedding and output 64 dims vectors.

\par \noindent  \textbf{Refinement module $R$.}
We use a ResNet decoder as our refinement module, which consists 8 ResNet blocks, in which we downsample at the 3rd layers and upsample at 6th and 7th layers.
The output feature dims are 64, 128, 256, 256, 128, 128, 128, 3.

\par \noindent  \textbf{Two-stage training for FWD-U.}
We find that a two-stage training scheme would slightly improve the performance of FWD-U model~(~0.4dB). 
In stage one, we first train the depth networks by constructing RGB point clouds for each input view and projecting them to the target view via differentiable point cloud rendering. 
We directly aggregate all the point clouds from various input views and get the rendered RGB images at target views. 
The depth network is trained by photometric loss between rendered views and target images. 
This step works as a simple unsupervised MVS scheme and gives a proper initialization of the depth network.
We then train the whole model with the initialized depth network in stage two. 
This two-stage training scheme's intuition is that jointly training depth network and other components from scratch are unstable, and our first training stage could give a initialization for depth network. 

\section{License Discussions}

We discuss the licenses of assets. 

\subsection{Dataset}
\par \noindent \textbf{ShapeNet Dataset~\cite{chang2015shapenet} .}
We conduct our experiments on the ShapeNet dataset and we cite the paper~\cite{chang2015shapenet} as required by the author.
More specifically, we download the data from \href{https://s3.eu-central-1.amazonaws.com/avg-projects/differentiable_volumetric_rendering/data/NMR_Dataset.zip}{NMR}, which is hosted by DVR~\cite{niemeyer2020differentiable} authors.
\par \noindent \textbf{DTU MVS Dataset~\cite{aanaes2016large}.}
We conduct our experiments on the DTU MVS dataset. 
This dataset doesn't include any license. 
On the other hand, we cite the paper~\cite{aanaes2016large} which is required by the paper.
\subsection{Methods}
\par \noindent \textbf{PixelNeRF~\cite{yu2020pixelnerf} .}
We evaluate the official code of PixelNeRF~\cite{yu2020pixelnerf} for comparison.
The code is hosted in the github page \href{https://github.com/sxyu/pixel-nerf}{https://github.com/sxyu/pixel-nerf},
which uses the BSD 2-Clause "Simplified" License.

\par \noindent \textbf{IBRNet~\cite{wang2021ibrnet} .}
We use the official code of IBRNet~\cite{wang2021ibrnet} for comparison, which is hosted at \href{https://github.com/googleinterns/IBRNet}{https://github.com/googleinterns/IBRNet}.
This project has the Apache License 2.0.

\par \noindent \textbf{MVSNeRF~\cite{chen2021mvsnerf}.}
We use the official code of MVSNeRF~\cite{chen2021mvsnerf} for comparison, which is hosted at \href{https://github.com/apchenstu/mvsnerf}{https://github.com/apchenstu/mvsnerf} with MIT License.
\par \noindent \textbf{SynSin~\cite{wiles2020synsin}.}
The code of SynSin~\cite{wiles2020synsin} is hosted at \href{https://github.com/facebookresearch/synsin}{https://github.com/facebookresearch/synsin} with Copyright (c) 2020, Facebook All rights reserved.
\par \noindent \textbf{Stable View Synthesis~(SVS)~\cite{Riegler2021SVS}.}
We use the code hosted at \href{https://github.com/isl-org/StableViewSynthesis}{https://github.com/isl-org/StableViewSynthesis} for evaluation, with MIT License and Copyright (c) 2021 Intel ISL (Intel Intelligent Systems Lab).
\par \noindent \textbf{DeepBlending~\cite{DeepBlending2018}.}
We implement it according to the code at \href{https://github.com/Phog/DeepBlending}{https://github.com/Phog/DeepBlending}.
This work is under Apache License.
\par \noindent \textbf{PixelNeRF-DS\cite{deng2021depth}.}
We use the number reported in~\cite{deng2021depth}.
\par \noindent \textbf{Pytorch3D.}
We use the code from Pytorch3D~\cite{ravi2020pytorch3d}: \href{https://github.com/facebookresearch/pytorch3d}{https://github.com/facebookresearch/pytorch3d} for our differentiable renderer.
The code is licensed with BSD 3-Clause License.

\section{User Study Details}

As detailed in the paper, we conduct user study to evaluate perceptual quality of synthesized images.
We provide more information about user study here.

We employ the standard A/B test paradigm as our user study format, which asks  workers to select the closest result to a ground truth image between two competing results.
Results of method A and B and ground truth target view are available during test.
All views in the test set~(690 views in total) are evaluated and each view is judged by 3 workers.

All tests are conducted using thehive.ai, a website similar to Amazon Mechanical Turk. 
Workers were given instructions and examples of tests, and then they were given a test to identify whether they understand the task.
Only workers passing the test were allowed to conduct A/B test.
Three images of the same view are shown in the test, where the first image is the ground truth target views and the rest two images are synthesized images.
Workers are asked to select ``left" or ``right'' image to indicate their preference. 
Results of method A and B are randomly placed for every test for fairness.
We show the instructions and examples below.

\begin{figure}
    \centering
    \includegraphics[width=0.5\textwidth]{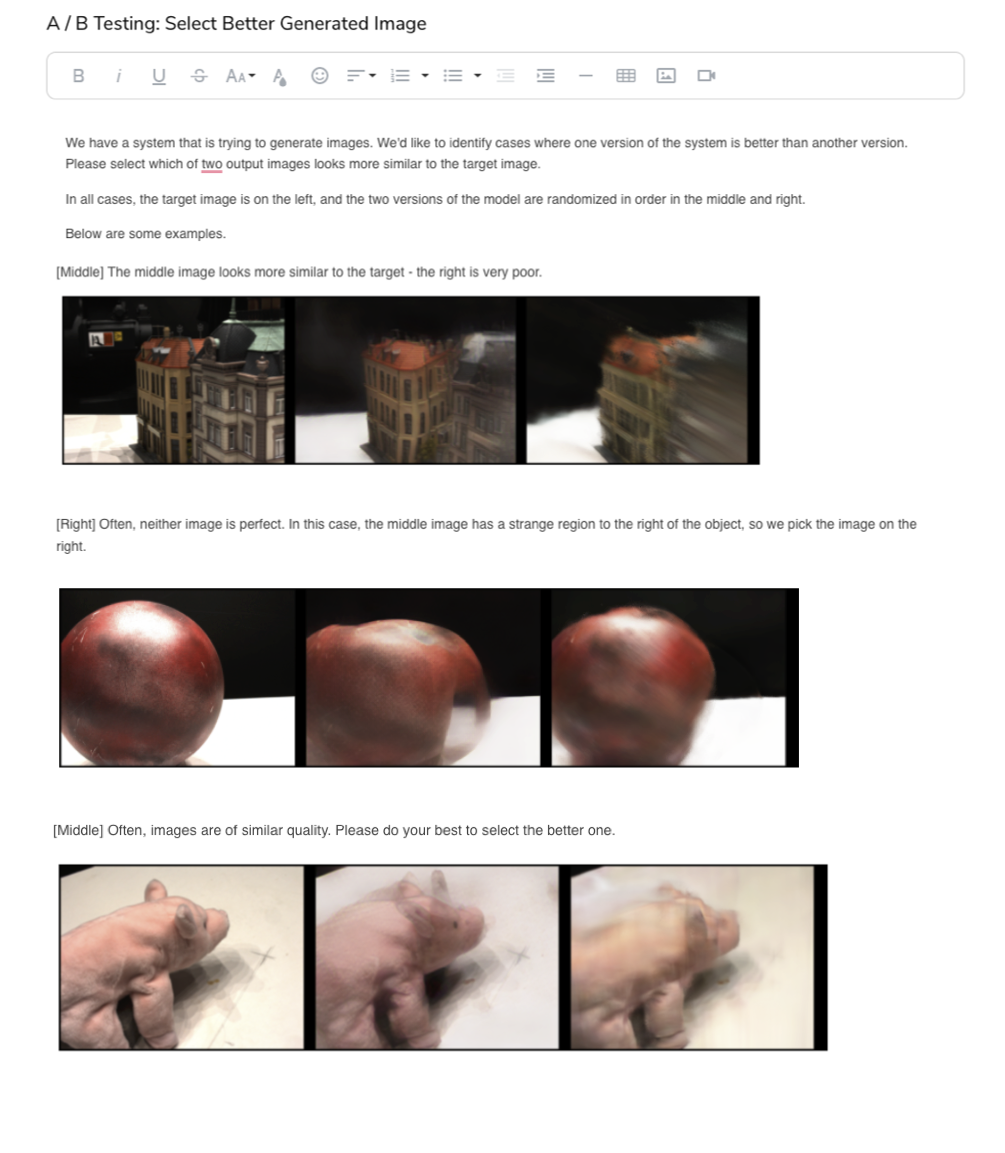}
    \captionof{figure}{
    \textbf{Instructions of A/B test.}
    We show the instructions of A/B test used in our paper.
    }
\end{figure}
\section{Additional Results}
Again, please see attached videos for comparison between ours and other methods.

\subsection{Time statistics.}
We show times spent on each component of \FWDrgbd model during a single forward pass in Table~\ref{tab:time}.
As shown in the Table, only 30 percent and 12 percent of times are spent on the Rendering process and Fusion process, indicating that our renderer and Transformer are highly-efficient.

Moreover, we compare the time distributions of our method with PixelNeRF and IBRNet in Table~\ref{tab:supp_time_details}.

\begin{table}[t]
    \centering
    \caption{
    \textbf{ Inference time of FWD-D.}
    We count times used for each stage in single forward pass with a batch size of 24. }
    \resizebox{0.48\textwidth}{!}{
        \centering
        \begin{tabular}{p{1.3cm} p{1cm} p{1cm} p{1cm} p{1cm} p{1.2cm} p{1cm}}
        \toprule[2pt]
         & Constr. & Render & Fusion &Refine. & Total\\
         \midrule
        Time~(ms) &40.39 &172.57 & 70.11 & 302.59 &585.66 \\
        Percent  & 6.90 & 29.47 & 11.97 & 61.67 & 100.00\\
        \bottomrule [2 pt]
        \end{tabular}}
    \label{tab:time}
\end{table}

\begin{table}[t]
    \centering
    \captionof{table}{
    \textbf{Inference time between different methods.}
    We study the time distributions for each method to render a total scene~(46 views) for 
    PixelNeRF, IBRNet and FWD.
    Feature Encoder refers feature extractions for input views, including feature encoding for PixelNeRF and IBRNet,
    feature encoding and depth regression for FWD.
    Renderer refers to the time spent for synthesizing each view.}
    \begin{tabular}{p{2cm} p{1.5cm} p{1.5cm} p{1.5cm}}
    \toprule[2pt]
    model & Feature Encoder & Renderer & Total\\
    \midrule
    PixelNeRF &0.696s & 30min& 30min \\
    IBRNet &0.87s & 163.2s & 173s \\
    FWD &0.085s & 1.10s & 1.19s \\
    \bottomrule [2 pt]
    \end{tabular}
    \label{tab:supp_time_details}
\end{table}
\subsection{Other ablations.}

\subsection{View number ablations.}
\begin{table*}[ht!]
    \centering
    \caption{We investigate the novel view synthesis performance and speed when provided with more input views. }
    \resizebox{\ifdim\width>\linewidth
        \linewidth
      \else
        \width
      \fi}{!}
    {
    \begin{tabular}{ l |c c c c | c c c c| c c c c}
    \toprule[2pt]
      &\multicolumn{4}{c}{3 views} &\multicolumn{4}{c}{6 views} &\multicolumn{4}{c}{9 views} \\
      &PSNR & SSIM &LPIPS & Time &PSNR & SSIM &LPIPS & Time &PSNR & SSIM &LPIPS & Time\\
    \midrule
    PixelNeRF &19.24 &0.687 &0.399 &0.025 &20.29 &0.725 &0.372 & 0.013 &20.91 &0.75 &0.348 & 0.009 min\\
    IBRNet &18.86 & 0.695 & 0.387 & 0.265 & 20.93 & 0.780 & 0.312 & 0.176 & 21.30 & 0.805 & 0.285 & 0.125 \\
    FWD-U &17.42 &0.598 &0.341 &35.4 &18.11 &0.623 &0.323 & 15.0 &18.93 & 0.648 &0.304 &8.5  \\
    FWD &20.15 & 0.721 &0.259 &35.4 & 21.40 &0.758 &0.235 & 15.0 &21.98 &0.779 &0.218 & 8.5\\
    FWD-D &21.98 & 0.791 &0.208 &43.2 & 22.54 &0.802 &0.199 & 29.5 &22.95 &0.816 &0.188 & 22.8\\

    \bottomrule [2 pt]
    
    \end{tabular}
    }
    \label{tab:multiview}
\end{table*}

\subsection{Synthesis Results.}
\begin{figure*}
   \centering
   \begin{tabular}{>{\footnotesize}c@{}@{}c@{\hspace{0.5mm}}c@{\hspace{0.5mm}}c@{\hspace{0.5 mm}}c@{\hspace{0.5mm}}c@{\hspace{0.5mm}}c@{}}
   \parbox[t]{4mm}{\multirow{3}{*}{\rotatebox[origin=c]{90}{\text{Input: 3 views of held-out scene}}}} & 
   \includegraphics[width=0.16\textwidth]{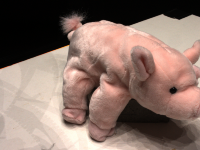}&
   \includegraphics[width=0.16\textwidth]{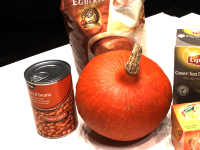}&
   \includegraphics[width=0.16\textwidth]{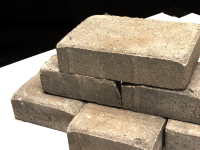}&
   \includegraphics[width=0.16\textwidth]{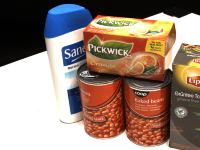}&
   \includegraphics[width=0.16\textwidth]{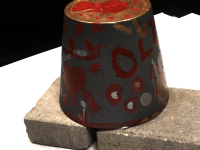}&
   \includegraphics[width=0.16\textwidth]{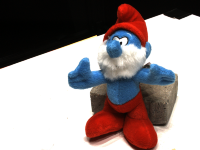}\\
&
  \includegraphics[width=0.16\textwidth]{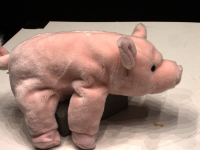}&
  \includegraphics[width=0.16\textwidth]{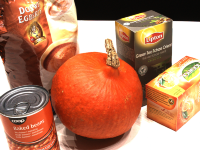}&
  \includegraphics[width=0.16\textwidth]{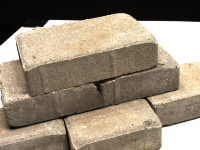}&
  \includegraphics[width=0.16\textwidth]{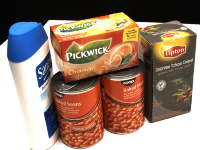}&
  \includegraphics[width=0.16\textwidth]{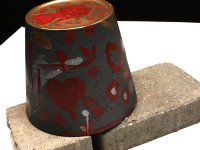}&
  \includegraphics[width=0.16\textwidth]{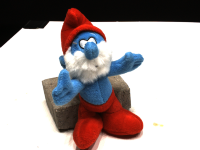}\\
   &
   \includegraphics[width=0.16\textwidth]{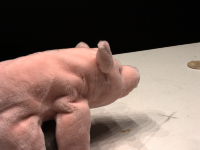}&
   \includegraphics[width=0.16\textwidth]{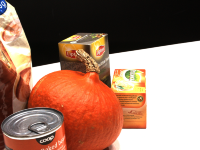}&
   \includegraphics[width=0.16\textwidth]{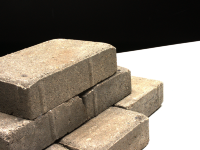}&
   \includegraphics[width=0.16\textwidth]{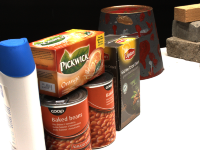}&
   \includegraphics[width=0.16\textwidth]{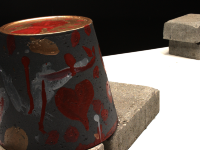}&
   \includegraphics[width=0.16\textwidth]{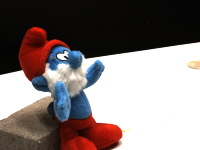}\\
   \bottomrule[2pt]
   \parbox[t]{4mm}{\multirow{5}{*}{\rotatebox[origin=c]{90}{\text{Novel views}}}} &
   \includegraphics[width=0.16\textwidth]{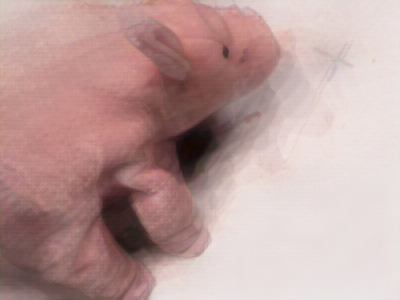}&
   \includegraphics[width=0.16\textwidth]{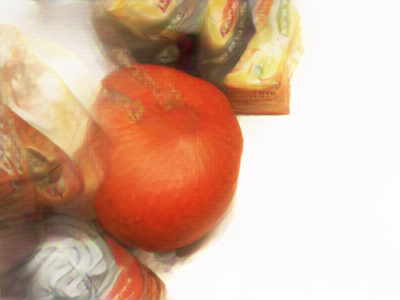}&
   \includegraphics[width=0.16\textwidth]{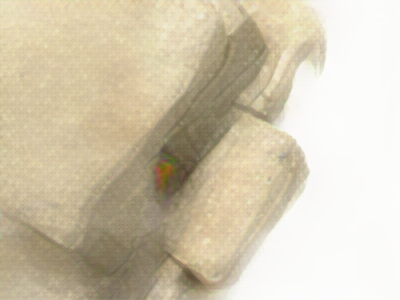}&
   \includegraphics[width=0.16\textwidth]{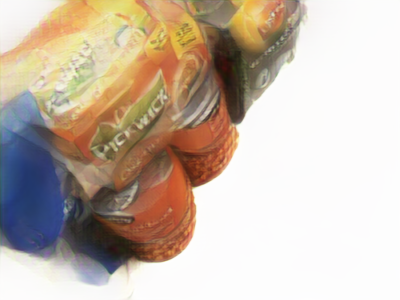}&
   \includegraphics[width=0.16\textwidth]{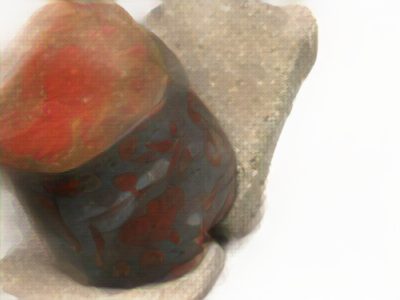}&
   \includegraphics[width=0.16\textwidth]{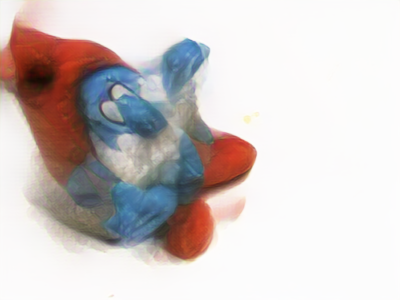}\\

   &
   \includegraphics[width=0.16\textwidth]{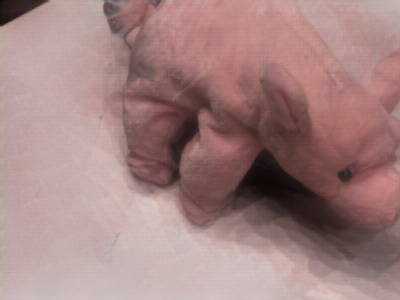}&
   \includegraphics[width=0.16\textwidth]{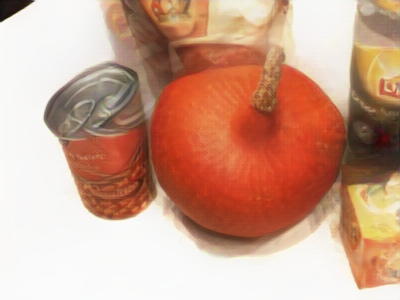}&
   \includegraphics[width=0.16\textwidth]{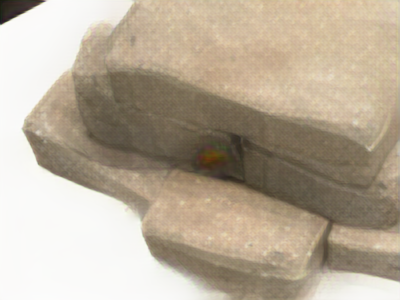}&
   \includegraphics[width=0.16\textwidth]{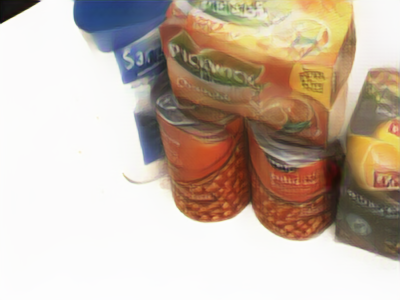}&
   \includegraphics[width=0.16\textwidth]{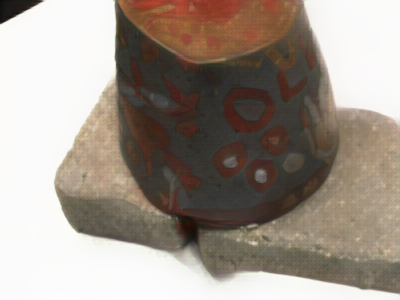}&
   \includegraphics[width=0.16\textwidth]{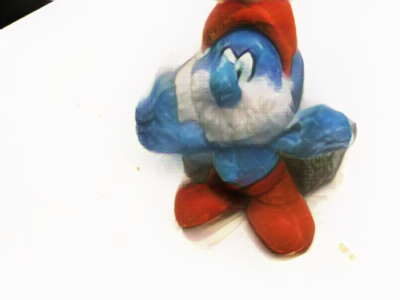}\\
   &
   \includegraphics[width=0.16\textwidth]{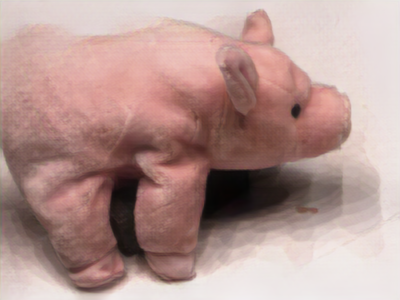}&
   \includegraphics[width=0.16\textwidth]{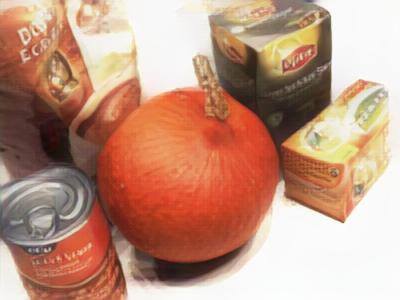}&
   \includegraphics[width=0.16\textwidth]{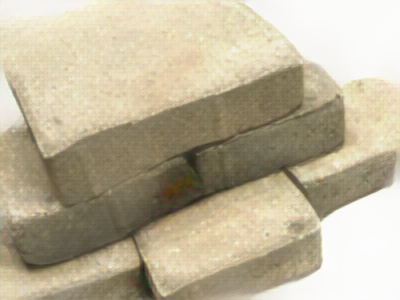}&
   \includegraphics[width=0.16\textwidth]{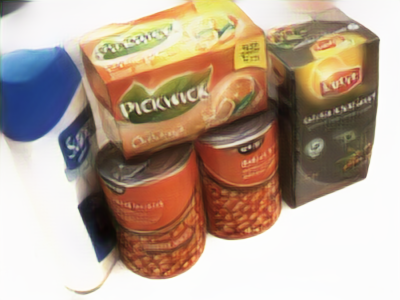}&
   \includegraphics[width=0.16\textwidth]{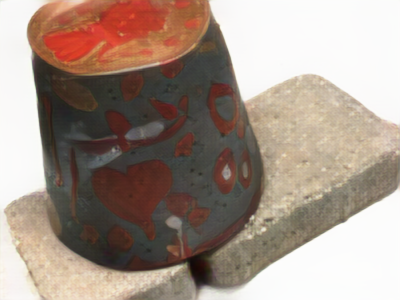}&
   \includegraphics[width=0.16\textwidth]{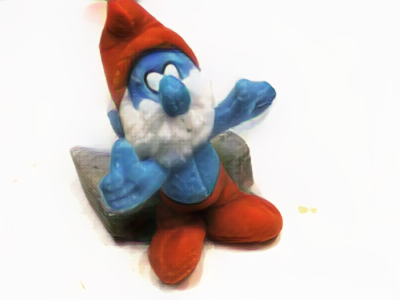}\\
   &
   \includegraphics[width=0.16\textwidth]{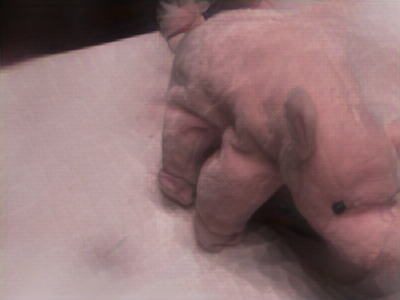}&
   \includegraphics[width=0.16\textwidth]{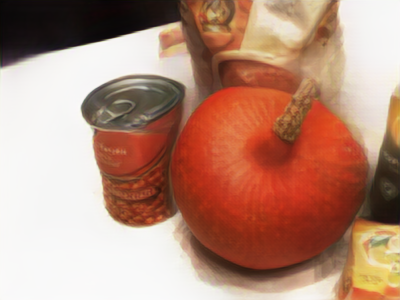}&
   \includegraphics[width=0.16\textwidth]{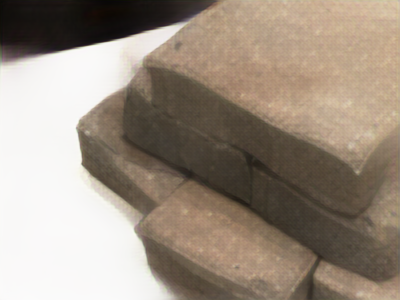}&
   \includegraphics[width=0.16\textwidth]{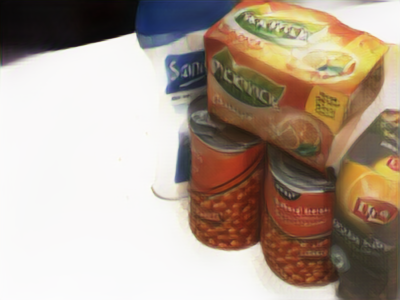}&
   \includegraphics[width=0.16\textwidth]{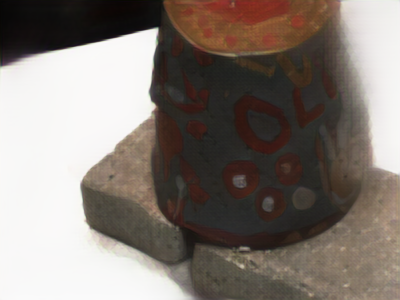}&
   \includegraphics[width=0.16\textwidth]{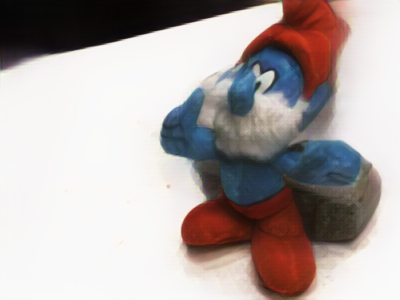}\\
   &
   \includegraphics[width=0.16\textwidth]{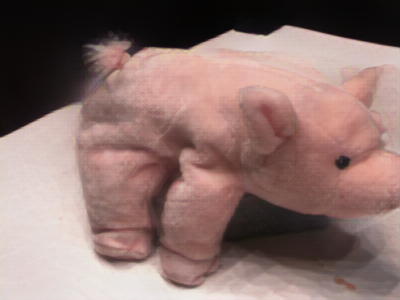}&
   \includegraphics[width=0.16\textwidth]{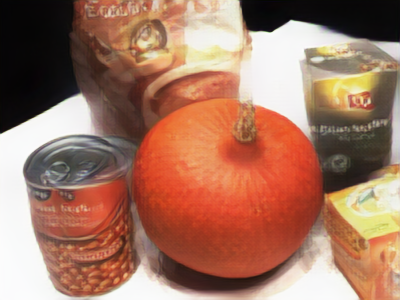}&
   \includegraphics[width=0.16\textwidth]{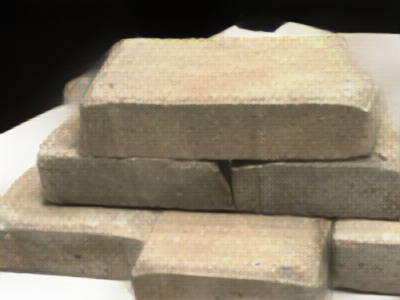}&
   \includegraphics[width=0.16\textwidth]{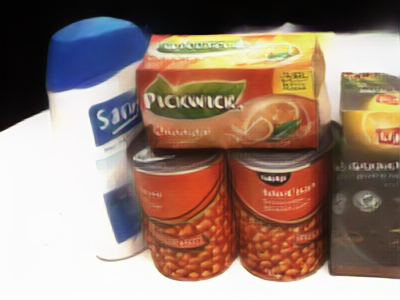}&
   \includegraphics[width=0.16\textwidth]{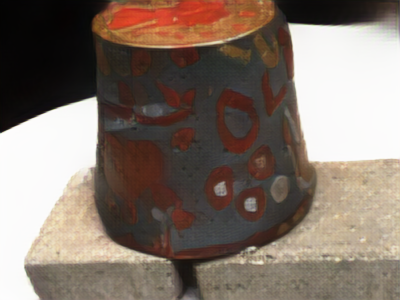}&
   \includegraphics[width=0.16\textwidth]{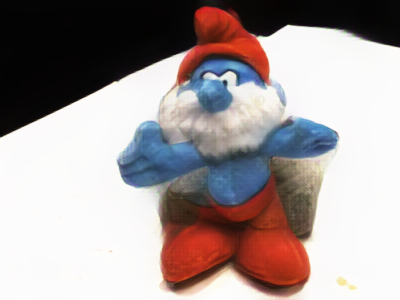}\\
   \end{tabular}
   \caption{\textbf{Efficient view synthesis from very sparse views for FWD-U.}
   We show the view synthesis results with 3 input views on DTU MVS test dataset for FWD-U trained with unsupervised depths.
   }
   \label{fig:fwd_u}
   \end{figure*}

\begin{figure*}
   \centering
   \begin{tabular}{>{\footnotesize}c@{}@{}c@{\hspace{0.5mm}}c@{\hspace{0.5mm}}c@{\hspace{0.5 mm}}c@{\hspace{0.5mm}}c@{\hspace{0.5mm}}c@{}}
   \parbox[t]{4mm}{\multirow{3}{*}{\rotatebox[origin=c]{90}{\text{Input: 3 views of held-out scene}}}} & 
   \includegraphics[width=0.16\textwidth]{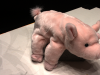}&
   \includegraphics[width=0.16\textwidth]{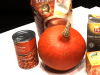}&
   \includegraphics[width=0.16\textwidth]{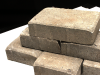}&
   \includegraphics[width=0.16\textwidth]{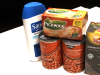}&
   \includegraphics[width=0.16\textwidth]{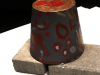}&
   \includegraphics[width=0.16\textwidth]{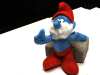}\\
&
  \includegraphics[width=0.16\textwidth]{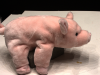}&
  \includegraphics[width=0.16\textwidth]{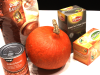}&
  \includegraphics[width=0.16\textwidth]{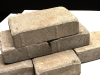}&
  \includegraphics[width=0.16\textwidth]{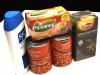}&
  \includegraphics[width=0.16\textwidth]{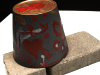}&
  \includegraphics[width=0.16\textwidth]{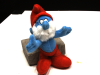}\\
   &
   \includegraphics[width=0.16\textwidth]{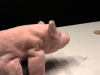}&
   \includegraphics[width=0.16\textwidth]{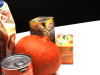}&
   \includegraphics[width=0.16\textwidth]{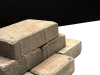}&
   \includegraphics[width=0.16\textwidth]{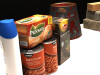}&
   \includegraphics[width=0.16\textwidth]{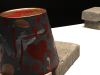}&
   \includegraphics[width=0.16\textwidth]{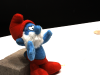}\\
   \bottomrule[2pt]
   \parbox[t]{4mm}{\multirow{5}{*}{\rotatebox[origin=c]{90}{\text{Novel views}}}} &
   \includegraphics[width=0.16\textwidth]{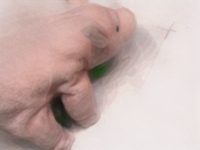}&
   \includegraphics[width=0.16\textwidth]{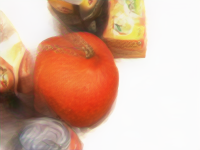}&
   \includegraphics[width=0.16\textwidth]{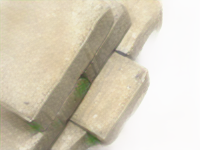}&
   \includegraphics[width=0.16\textwidth]{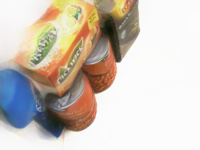}&
   \includegraphics[width=0.16\textwidth]{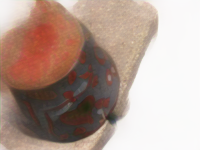}&
   \includegraphics[width=0.16\textwidth]{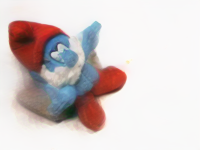}\\

   &
   \includegraphics[width=0.16\textwidth]{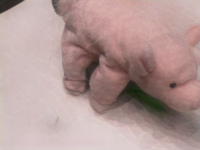}&
   \includegraphics[width=0.16\textwidth]{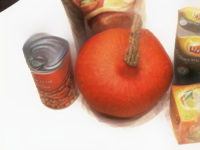}&
   \includegraphics[width=0.16\textwidth]{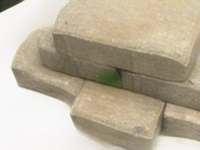}&
   \includegraphics[width=0.16\textwidth]{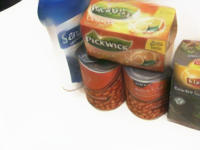}&
   \includegraphics[width=0.16\textwidth]{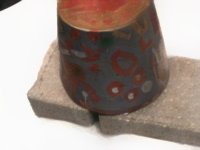}&
   \includegraphics[width=0.16\textwidth]{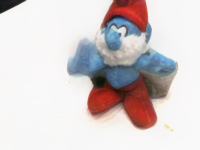}\\
   &
   \includegraphics[width=0.16\textwidth]{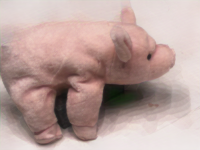}&
   \includegraphics[width=0.16\textwidth]{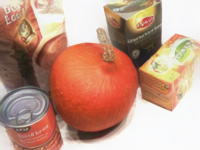}&
   \includegraphics[width=0.16\textwidth]{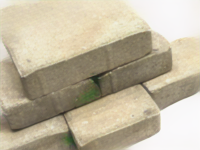}&
   \includegraphics[width=0.16\textwidth]{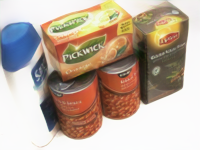}&
   \includegraphics[width=0.16\textwidth]{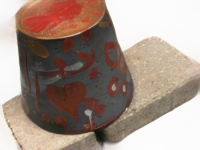}&
   \includegraphics[width=0.16\textwidth]{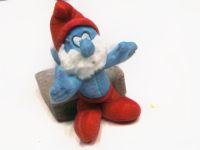}\\
   &
   \includegraphics[width=0.16\textwidth]{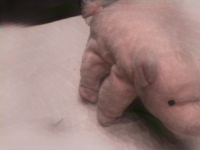}&
   \includegraphics[width=0.16\textwidth]{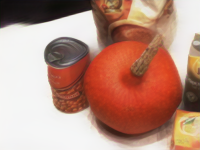}&
   \includegraphics[width=0.16\textwidth]{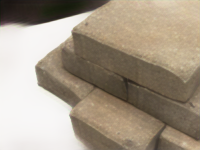}&
   \includegraphics[width=0.16\textwidth]{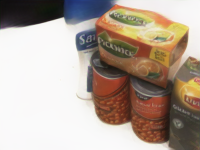}&
   \includegraphics[width=0.16\textwidth]{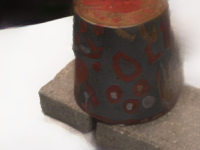}&
   \includegraphics[width=0.16\textwidth]{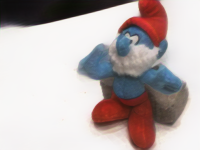}\\
   &
   \includegraphics[width=0.16\textwidth]{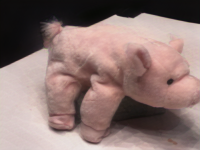}&
   \includegraphics[width=0.16\textwidth]{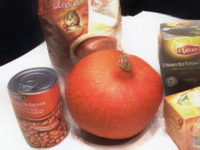}&
   \includegraphics[width=0.16\textwidth]{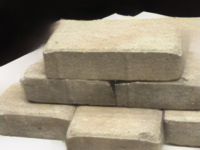}&
   \includegraphics[width=0.16\textwidth]{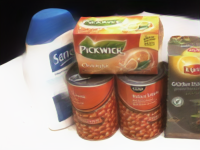}&
   \includegraphics[width=0.16\textwidth]{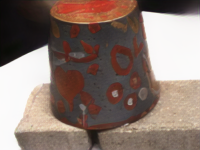}&
   \includegraphics[width=0.16\textwidth]{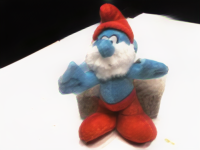}\\
   \end{tabular}
   \caption{\textbf{Efficient view synthesis from very sparse views for FWD. }
   We show the view synthesis results with 3 input views on DTU MVS test dataset for FWD.
   }
   \label{fig:fwd}
   \end{figure*}

\begin{figure*}
    \centering
    \begin{tabular}{>{\footnotesize}c@{}@{}c@{\hspace{0.5mm}}c@{\hspace{0.5mm}}c@{\hspace{0.5 mm}}c@{\hspace{0.5mm}}c@{\hspace{0.5mm}}c@{}}
    \parbox[t]{4mm}{\multirow{3}{*}{\rotatebox[origin=c]{90}{\text{Input: 3 views of held-out scene}}}} & 
    \includegraphics[width=0.16\textwidth]{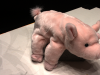}&
    \includegraphics[width=0.16\textwidth]{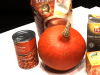}&
    \includegraphics[width=0.16\textwidth]{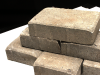}&
    \includegraphics[width=0.16\textwidth]{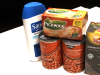}&
    \includegraphics[width=0.16\textwidth]{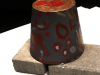}&
    \includegraphics[width=0.16\textwidth]{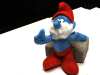}\\
 &
   \includegraphics[width=0.16\textwidth]{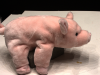}&
   \includegraphics[width=0.16\textwidth]{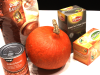}&
   \includegraphics[width=0.16\textwidth]{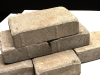}&
   \includegraphics[width=0.16\textwidth]{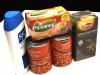}&
   \includegraphics[width=0.16\textwidth]{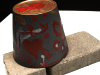}&
   \includegraphics[width=0.16\textwidth]{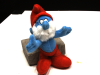}\\
    &
    \includegraphics[width=0.16\textwidth]{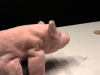}&
    \includegraphics[width=0.16\textwidth]{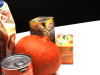}&
    \includegraphics[width=0.16\textwidth]{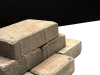}&
    \includegraphics[width=0.16\textwidth]{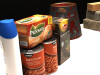}&
    \includegraphics[width=0.16\textwidth]{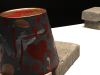}&
    \includegraphics[width=0.16\textwidth]{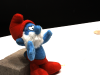}\\
    \bottomrule[2pt]
    \parbox[t]{4mm}{\multirow{5}{*}{\rotatebox[origin=c]{90}{\text{Novel views}}}} &
    \includegraphics[width=0.16\textwidth]{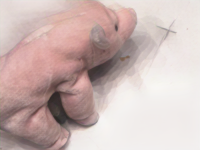}&
    \includegraphics[width=0.16\textwidth]{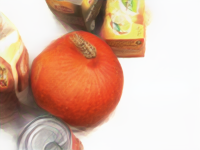}&
    \includegraphics[width=0.16\textwidth]{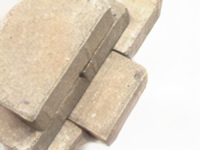}&
    \includegraphics[width=0.16\textwidth]{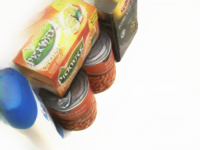}&
    \includegraphics[width=0.16\textwidth]{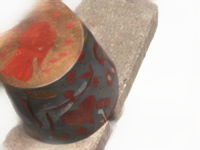}&
    \includegraphics[width=0.16\textwidth]{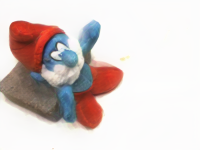}\\
 
    &
    \includegraphics[width=0.16\textwidth]{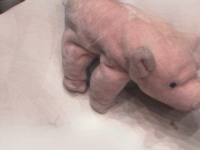}&
    \includegraphics[width=0.16\textwidth]{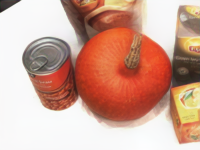}&
    \includegraphics[width=0.16\textwidth]{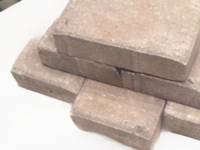}&
    \includegraphics[width=0.16\textwidth]{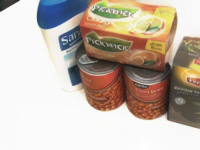}&
    \includegraphics[width=0.16\textwidth]{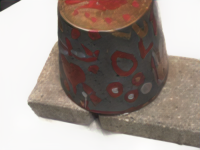}&
    \includegraphics[width=0.16\textwidth]{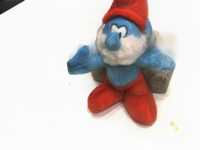}\\
    &
    \includegraphics[width=0.16\textwidth]{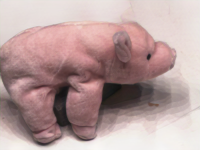}&
    \includegraphics[width=0.16\textwidth]{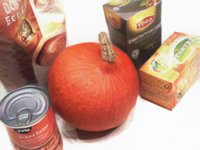}&
    \includegraphics[width=0.16\textwidth]{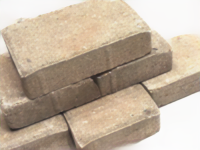}&
    \includegraphics[width=0.16\textwidth]{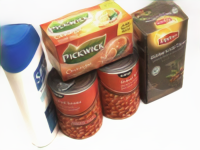}&
    \includegraphics[width=0.16\textwidth]{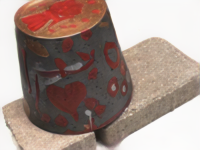}&
    \includegraphics[width=0.16\textwidth]{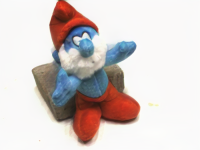}\\
    &
    \includegraphics[width=0.16\textwidth]{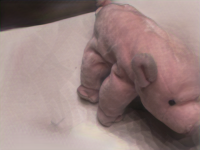}&
    \includegraphics[width=0.16\textwidth]{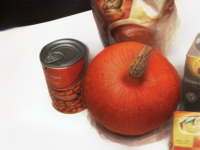}&
    \includegraphics[width=0.16\textwidth]{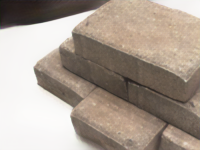}&
    \includegraphics[width=0.16\textwidth]{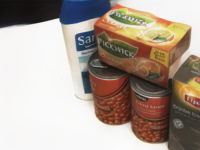}&
    \includegraphics[width=0.16\textwidth]{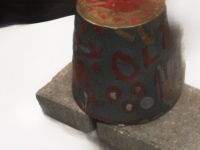}&
    \includegraphics[width=0.16\textwidth]{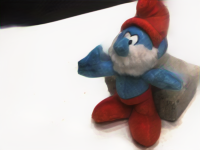}\\
    &
    \includegraphics[width=0.16\textwidth]{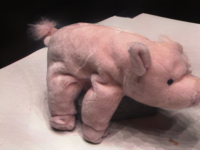}&
    \includegraphics[width=0.16\textwidth]{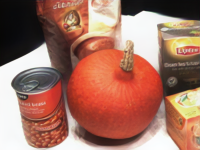}&
    \includegraphics[width=0.16\textwidth]{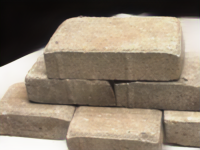}&
    \includegraphics[width=0.16\textwidth]{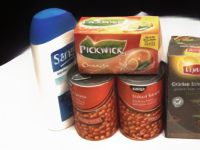}&
    \includegraphics[width=0.16\textwidth]{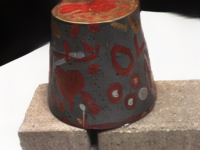}&
    \includegraphics[width=0.16\textwidth]{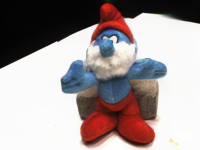}\\
    \end{tabular}
    \captionof{figure}{\textbf{Efficient view synthesis from very sparse views for FWD-D. }
    We show the view synthesis results with 3 input views on DTU MVS test dataset for FWD-D.
    }
    \label{fig:fwd_d}
    \end{figure*}

\begin{figure*}
   \centering
   \begin{tabular}{>{\footnotesize}c@{}@{}c@{\hspace{0.5mm}}c@{\hspace{0.5mm}}c@{\hspace{0.5 mm}}c@{\hspace{0.5mm}}c@{\hspace{0.5mm}}c@{}}
   \parbox[t]{4mm}{\multirow{3}{*}{\rotatebox[origin=c]{90}{\text{Input: 3 views of held-out scene}}}} & 
   \includegraphics[width=0.16\textwidth]{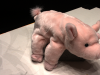}&
   \includegraphics[width=0.16\textwidth]{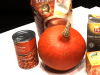}&
   \includegraphics[width=0.16\textwidth]{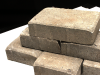}&
   \includegraphics[width=0.16\textwidth]{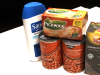}&
   \includegraphics[width=0.16\textwidth]{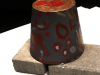}&
   \includegraphics[width=0.16\textwidth]{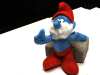}\\
&
  \includegraphics[width=0.16\textwidth]{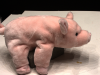}&
  \includegraphics[width=0.16\textwidth]{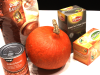}&
  \includegraphics[width=0.16\textwidth]{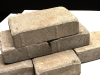}&
  \includegraphics[width=0.16\textwidth]{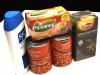}&
  \includegraphics[width=0.16\textwidth]{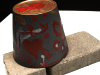}&
  \includegraphics[width=0.16\textwidth]{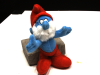}\\
   &
   \includegraphics[width=0.16\textwidth]{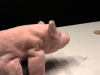}&
   \includegraphics[width=0.16\textwidth]{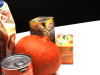}&
   \includegraphics[width=0.16\textwidth]{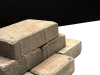}&
   \includegraphics[width=0.16\textwidth]{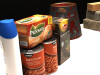}&
   \includegraphics[width=0.16\textwidth]{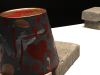}&
   \includegraphics[width=0.16\textwidth]{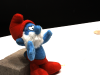}\\
   \bottomrule[2pt]
   \parbox[t]{4mm}{\multirow{5}{*}{\rotatebox[origin=c]{90}{\text{Novel views}}}} &
   \includegraphics[width=0.16\textwidth]{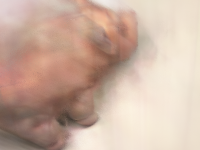}&
   \includegraphics[width=0.16\textwidth]{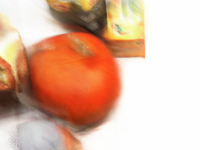}&
   \includegraphics[width=0.16\textwidth]{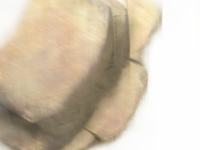}&
   \includegraphics[width=0.16\textwidth]{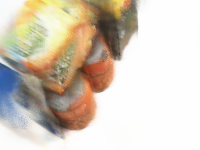}&
   \includegraphics[width=0.16\textwidth]{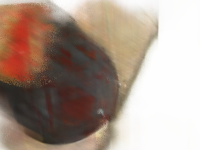}&
   \includegraphics[width=0.16\textwidth]{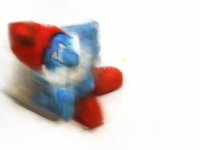}\\

   &
   \includegraphics[width=0.16\textwidth]{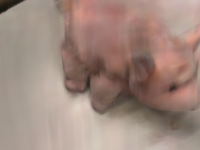}&
   \includegraphics[width=0.16\textwidth]{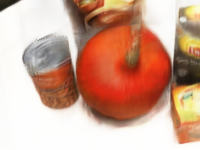}&
   \includegraphics[width=0.16\textwidth]{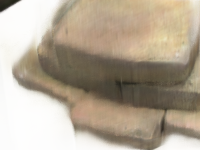}&
   \includegraphics[width=0.16\textwidth]{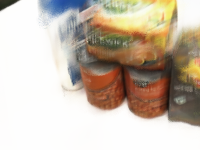}&
   \includegraphics[width=0.16\textwidth]{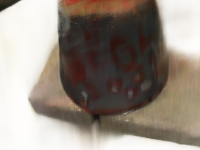}&
   \includegraphics[width=0.16\textwidth]{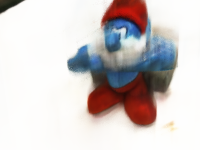}\\
   &
   \includegraphics[width=0.16\textwidth]{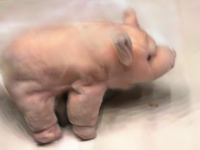}&
   \includegraphics[width=0.16\textwidth]{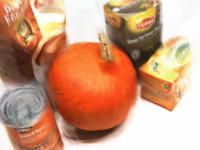}&
   \includegraphics[width=0.16\textwidth]{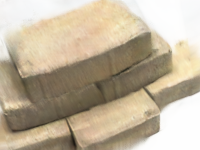}&
   \includegraphics[width=0.16\textwidth]{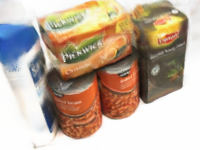}&
   \includegraphics[width=0.16\textwidth]{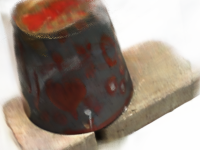}&
   \includegraphics[width=0.16\textwidth]{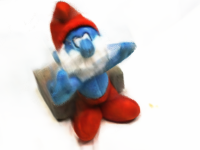}\\
   &
   \includegraphics[width=0.16\textwidth]{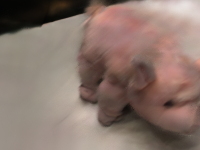}&
   \includegraphics[width=0.16\textwidth]{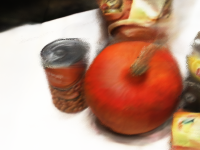}&
   \includegraphics[width=0.16\textwidth]{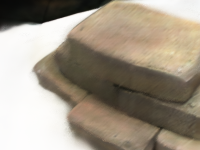}&
   \includegraphics[width=0.16\textwidth]{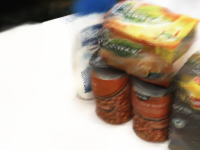}&
   \includegraphics[width=0.16\textwidth]{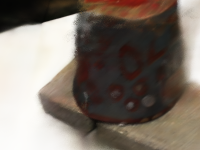}&
   \includegraphics[width=0.16\textwidth]{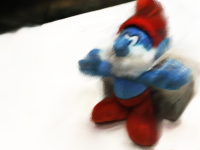}\\
   &
   \includegraphics[width=0.16\textwidth]{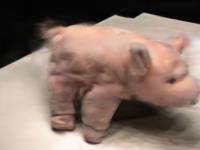}&
   \includegraphics[width=0.16\textwidth]{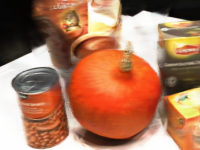}&
   \includegraphics[width=0.16\textwidth]{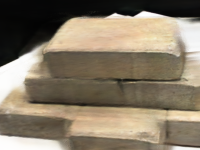}&
   \includegraphics[width=0.16\textwidth]{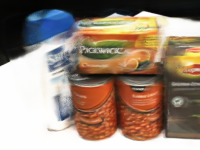}&
   \includegraphics[width=0.16\textwidth]{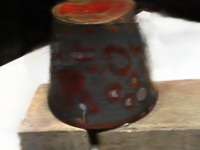}&
   \includegraphics[width=0.16\textwidth]{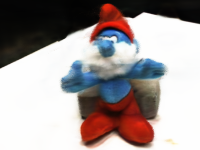}\\
   \end{tabular}
   \captionof{figure}{\textbf{Efficient view synthesis from very sparse views for PixelNeRF~\cite{yu2020pixelnerf}. }
   We show the view synthesis results with 3 input views on DTU MVS test dataset for PixelNeRF.
   }
   \label{fig:sup_pixelnerf}
   \end{figure*}

\begin{figure*}
   \centering
   \begin{tabular}{>{\footnotesize}c@{}@{}c@{\hspace{0.5mm}}c@{\hspace{0.5mm}}c@{\hspace{0.5 mm}}c@{\hspace{0.5mm}}c@{\hspace{0.5mm}}c@{}}
   \parbox[t]{4mm}{\multirow{3}{*}{\rotatebox[origin=c]{90}{\text{Input: 3 views of held-out scene}}}} & 
   \includegraphics[width=0.16\textwidth]{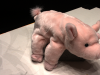}&
   \includegraphics[width=0.16\textwidth]{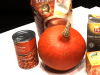}&
   \includegraphics[width=0.16\textwidth]{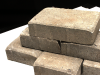}&
   \includegraphics[width=0.16\textwidth]{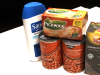}&
   \includegraphics[width=0.16\textwidth]{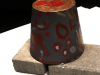}&
   \includegraphics[width=0.16\textwidth]{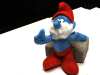}\\
&
  \includegraphics[width=0.16\textwidth]{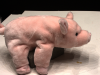}&
  \includegraphics[width=0.16\textwidth]{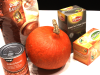}&
  \includegraphics[width=0.16\textwidth]{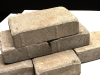}&
  \includegraphics[width=0.16\textwidth]{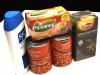}&
  \includegraphics[width=0.16\textwidth]{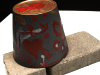}&
  \includegraphics[width=0.16\textwidth]{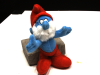}\\
   &
   \includegraphics[width=0.16\textwidth]{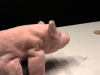}&
   \includegraphics[width=0.16\textwidth]{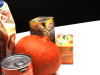}&
   \includegraphics[width=0.16\textwidth]{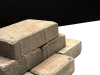}&
   \includegraphics[width=0.16\textwidth]{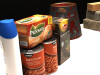}&
   \includegraphics[width=0.16\textwidth]{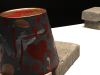}&
   \includegraphics[width=0.16\textwidth]{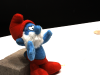}\\
   \bottomrule[2pt]
   \parbox[t]{4mm}{\multirow{5}{*}{\rotatebox[origin=c]{90}{\text{Novel views}}}} &
   \includegraphics[width=0.16\textwidth]{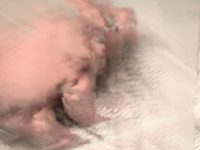}&
   \includegraphics[width=0.16\textwidth]{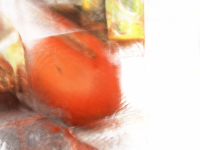}&
   \includegraphics[width=0.16\textwidth]{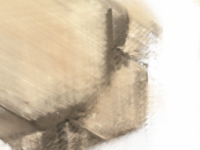}&
   \includegraphics[width=0.16\textwidth]{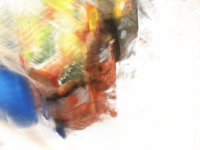}&
   \includegraphics[width=0.16\textwidth]{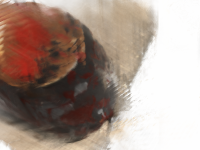}&
   \includegraphics[width=0.16\textwidth]{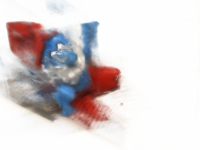}\\

   &
   \includegraphics[width=0.16\textwidth]{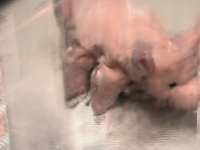}&
   \includegraphics[width=0.16\textwidth]{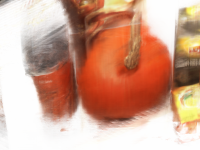}&
   \includegraphics[width=0.16\textwidth]{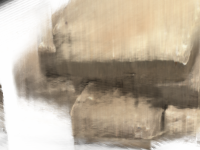}&
   \includegraphics[width=0.16\textwidth]{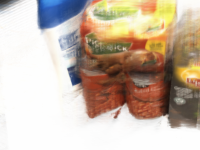}&
   \includegraphics[width=0.16\textwidth]{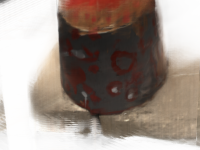}&
   \includegraphics[width=0.16\textwidth]{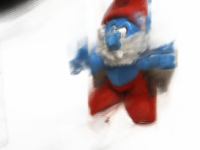}\\
   &
   \includegraphics[width=0.16\textwidth]{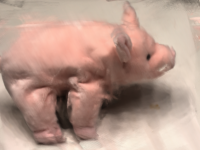}&
   \includegraphics[width=0.16\textwidth]{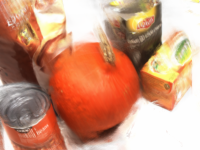}&
   \includegraphics[width=0.16\textwidth]{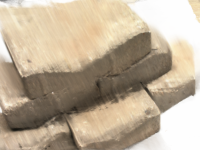}&
   \includegraphics[width=0.16\textwidth]{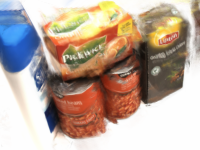}&
   \includegraphics[width=0.16\textwidth]{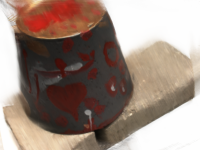}&
   \includegraphics[width=0.16\textwidth]{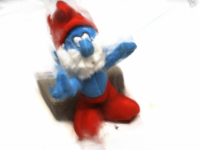}\\
   &
   \includegraphics[width=0.16\textwidth]{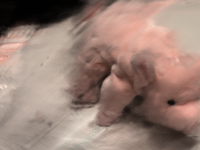}&
   \includegraphics[width=0.16\textwidth]{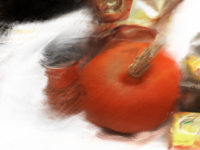}&
   \includegraphics[width=0.16\textwidth]{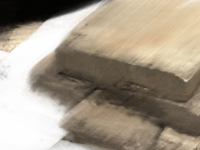}&
   \includegraphics[width=0.16\textwidth]{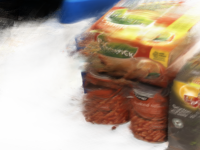}&
   \includegraphics[width=0.16\textwidth]{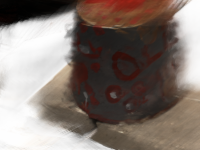}&
   \includegraphics[width=0.16\textwidth]{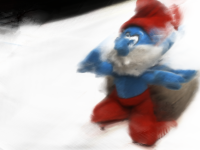}\\
   &
   \includegraphics[width=0.16\textwidth]{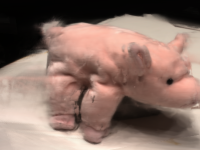}&
   \includegraphics[width=0.16\textwidth]{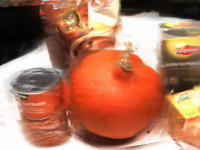}&
   \includegraphics[width=0.16\textwidth]{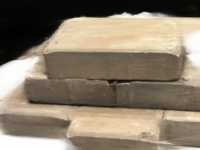}&
   \includegraphics[width=0.16\textwidth]{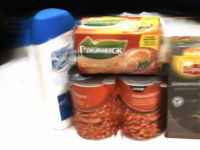}&
   \includegraphics[width=0.16\textwidth]{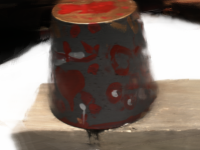}&
   \includegraphics[width=0.16\textwidth]{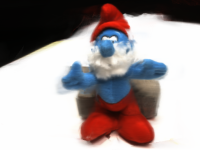}\\
   \end{tabular}
   \caption{\textbf{Efficient view synthesis from very sparse views for IBRNet~\cite{wang2021ibrnet}.}
   We show the view synthesis results with 3 input views on DTU MVS test dataset for IBRNet.
   }
   \label{fig:sup_dtu_IBRNet}
   \end{figure*}

\begin{figure*}
    \centering
    \begin{tabular}{>{\footnotesize}c@{}@{}c@{\hspace{0.5mm}}c@{\hspace{0.5mm}}c@{\hspace{0.5 mm}}c@{\hspace{0.5mm}}c@{\hspace{0.5mm}}c@{}}
    \parbox[t]{4mm}{\multirow{3}{*}{\rotatebox[origin=c]{90}{\text{Input: 3 views of held-out scene}}}} & 
    \includegraphics[width=0.16\textwidth]{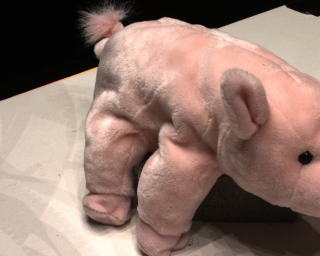}&
    \includegraphics[width=0.16\textwidth]{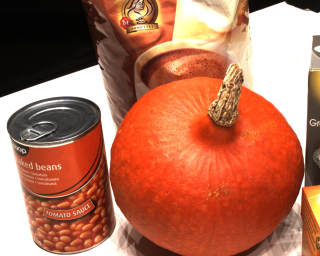}&
    \includegraphics[width=0.16\textwidth]{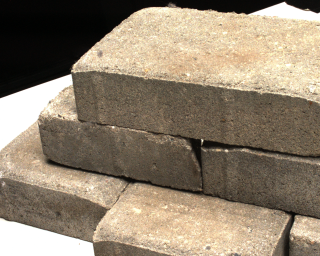}&
    \includegraphics[width=0.16\textwidth]{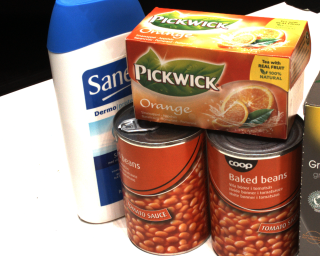}&
    \includegraphics[width=0.16\textwidth]{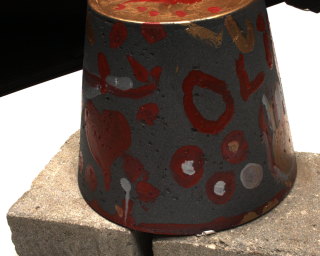}&
    \includegraphics[width=0.16\textwidth]{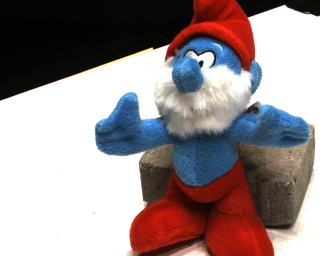}\\
 &
   \includegraphics[width=0.16\textwidth]{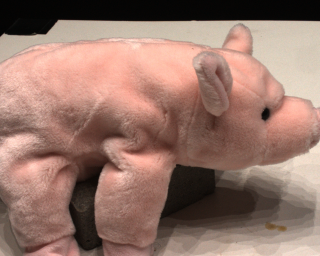}&
   \includegraphics[width=0.16\textwidth]{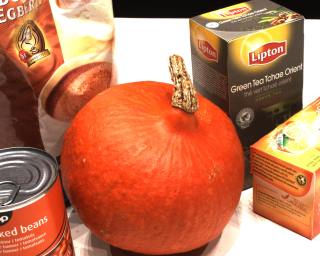}&
   \includegraphics[width=0.16\textwidth]{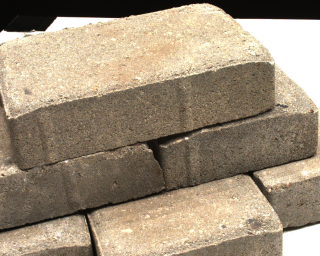}&
   \includegraphics[width=0.16\textwidth]{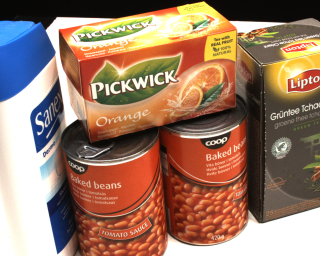}&
   \includegraphics[width=0.16\textwidth]{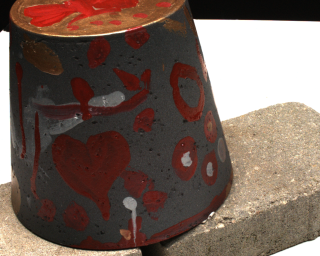}&
   \includegraphics[width=0.16\textwidth]{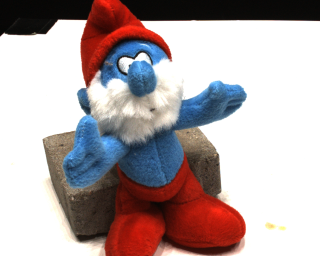}\\
    &
    \includegraphics[width=0.16\textwidth]{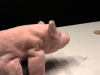}&
    \includegraphics[width=0.16\textwidth]{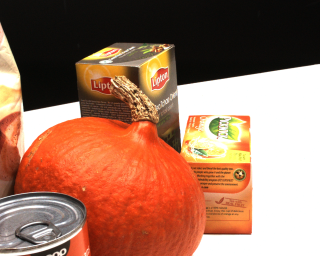}&
    \includegraphics[width=0.16\textwidth]{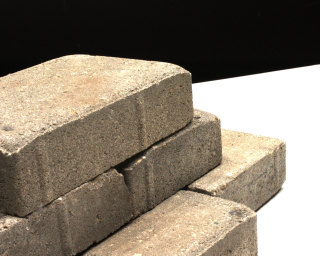}&
    \includegraphics[width=0.16\textwidth]{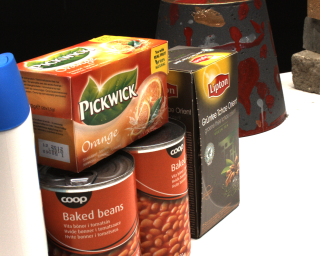}&
    \includegraphics[width=0.16\textwidth]{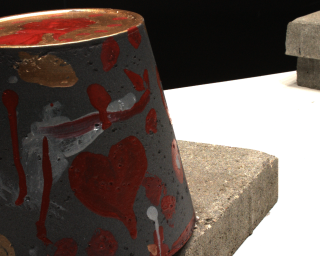}&
    \includegraphics[width=0.16\textwidth]{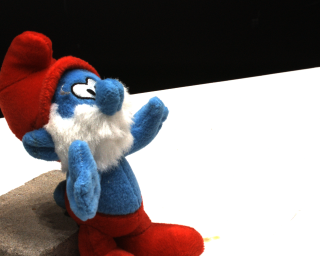}\\
    \bottomrule[2pt]
    \parbox[t]{4mm}{\multirow{5}{*}{\rotatebox[origin=c]{90}{\text{Novel views}}}} &
    \includegraphics[width=0.16\textwidth]{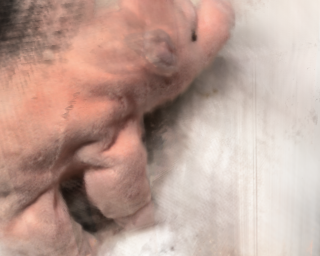}&
    \includegraphics[width=0.16\textwidth]{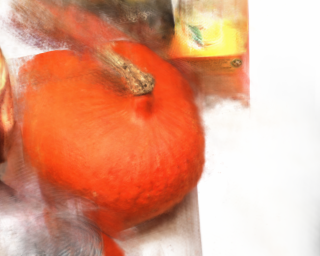}&
    \includegraphics[width=0.16\textwidth]{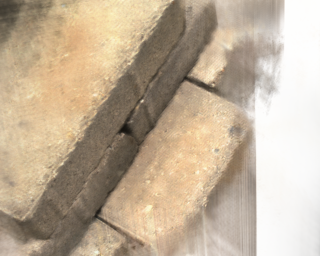}&
    \includegraphics[width=0.16\textwidth]{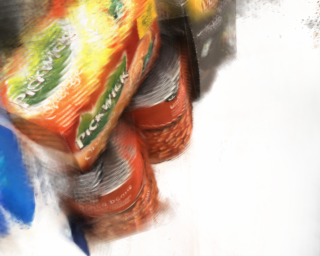}&
    \includegraphics[width=0.16\textwidth]{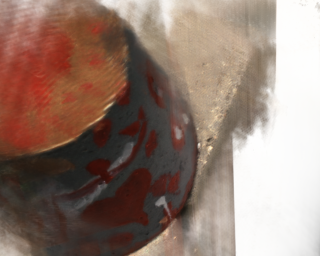}&
    \includegraphics[width=0.16\textwidth]{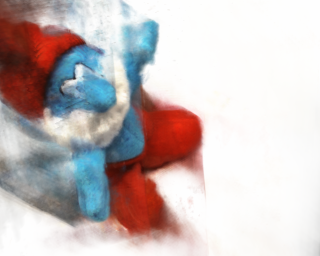}\\
 
    &
    \includegraphics[width=0.16\textwidth]{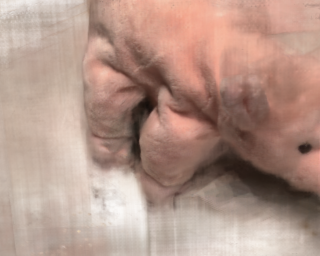}&
    \includegraphics[width=0.16\textwidth]{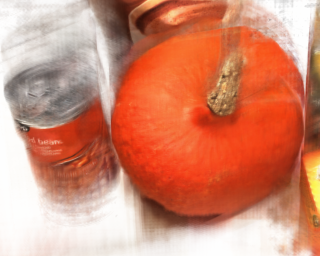}&
    \includegraphics[width=0.16\textwidth]{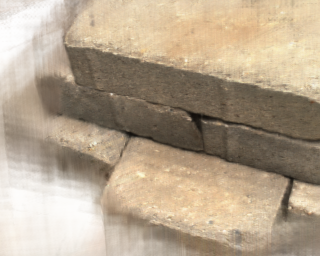}&
    \includegraphics[width=0.16\textwidth]{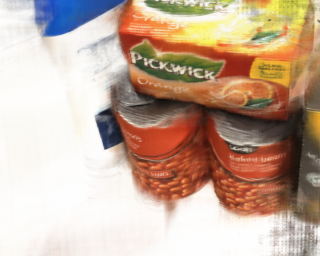}&
    \includegraphics[width=0.16\textwidth]{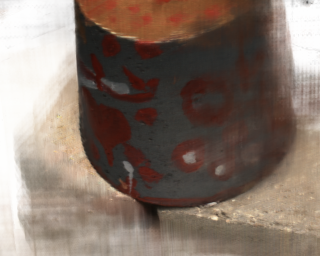}&
    \includegraphics[width=0.16\textwidth]{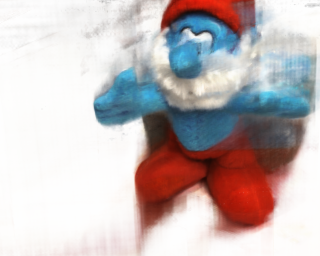}\\
    &
    \includegraphics[width=0.16\textwidth]{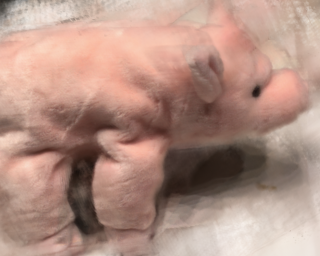}&
    \includegraphics[width=0.16\textwidth]{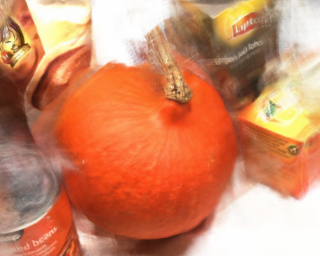}&
    \includegraphics[width=0.16\textwidth]{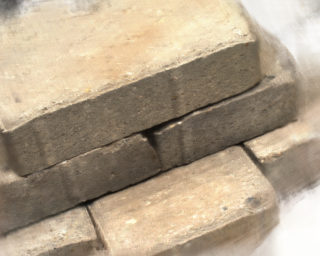}&
    \includegraphics[width=0.16\textwidth]{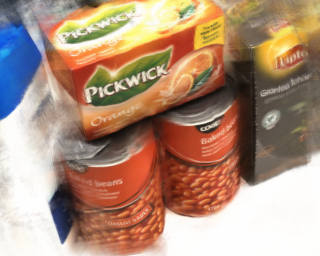}&
    \includegraphics[width=0.16\textwidth]{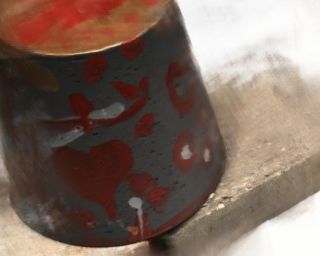}&
    \includegraphics[width=0.16\textwidth]{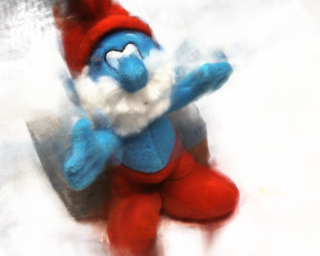}\\
    &
    \includegraphics[width=0.16\textwidth]{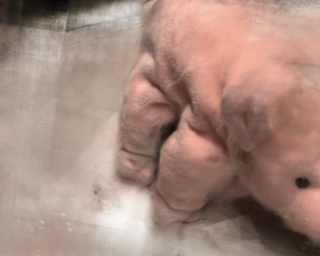}&
    \includegraphics[width=0.16\textwidth]{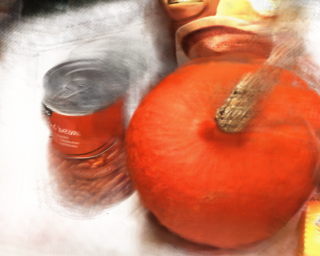}&
    \includegraphics[width=0.16\textwidth]{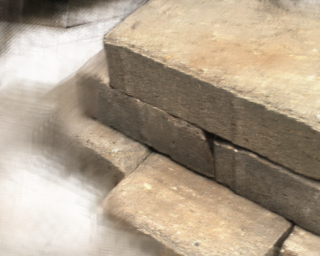}&
    \includegraphics[width=0.16\textwidth]{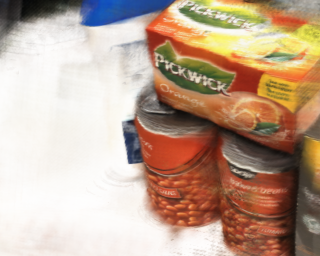}&
    \includegraphics[width=0.16\textwidth]{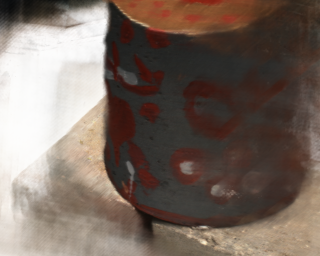}&
    \includegraphics[width=0.16\textwidth]{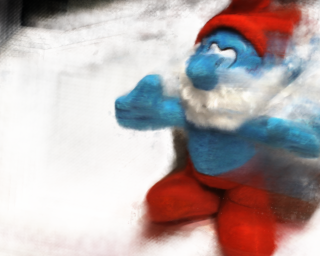}\\
    &
    \includegraphics[width=0.16\textwidth]{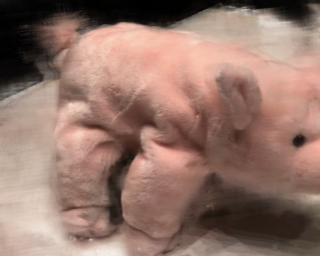}&
    \includegraphics[width=0.16\textwidth]{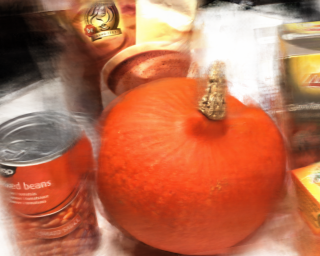}&
    \includegraphics[width=0.16\textwidth]{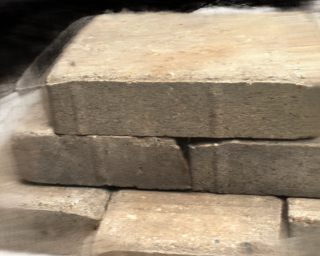}&
    \includegraphics[width=0.16\textwidth]{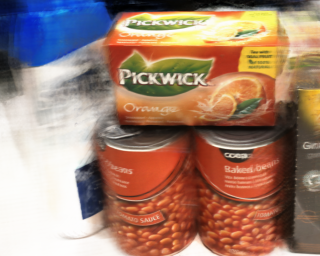}&
    \includegraphics[width=0.16\textwidth]{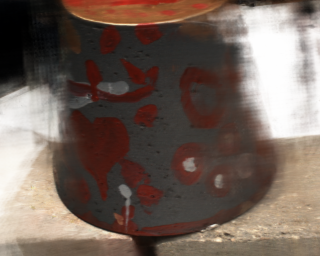}&
    \includegraphics[width=0.16\textwidth]{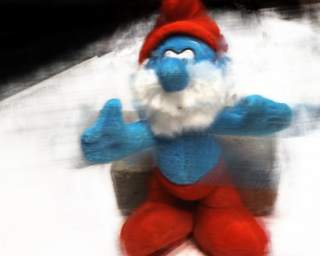}\\
    \end{tabular}
    \captionof{figure}{\textbf{Efficient view synthesis from very sparse views for MVSNeRF. }
    We show the view synthesis results with 3 input views on DTU MVS test dataset for MVSNeRF.
    }
    \label{fig:sup_dtu_mvsnerf}
    \end{figure*}

\begin{figure*}
    \centering
    \begin{tabular}{>{\footnotesize}c@{}@{}c@{\hspace{0.5mm}}c@{\hspace{0.5mm}}c@{\hspace{0.5 mm}}c@{\hspace{0.5mm}}c@{\hspace{0.5mm}}c@{}}
    \parbox[t]{4mm}{\multirow{3}{*}{\rotatebox[origin=c]{90}{\text{Input: 3 views of held-out scene}}}} & 
    \includegraphics[width=0.16\textwidth]{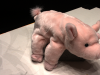}&
    \includegraphics[width=0.16\textwidth]{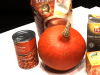}&
    \includegraphics[width=0.16\textwidth]{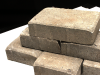}&
    \includegraphics[width=0.16\textwidth]{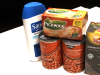}&
    \includegraphics[width=0.16\textwidth]{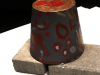}&
    \includegraphics[width=0.16\textwidth]{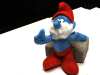}\\
 &
   \includegraphics[width=0.16\textwidth]{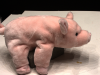}&
   \includegraphics[width=0.16\textwidth]{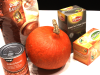}&
   \includegraphics[width=0.16\textwidth]{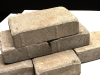}&
   \includegraphics[width=0.16\textwidth]{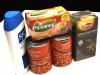}&
   \includegraphics[width=0.16\textwidth]{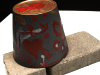}&
   \includegraphics[width=0.16\textwidth]{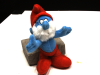}\\
    &
    \includegraphics[width=0.16\textwidth]{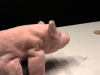}&
    \includegraphics[width=0.16\textwidth]{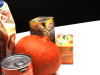}&
    \includegraphics[width=0.16\textwidth]{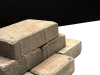}&
    \includegraphics[width=0.16\textwidth]{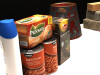}&
    \includegraphics[width=0.16\textwidth]{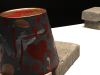}&
    \includegraphics[width=0.16\textwidth]{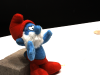}\\
    \bottomrule[2pt]
    \parbox[t]{4mm}{\multirow{5}{*}{\rotatebox[origin=c]{90}{\text{Novel views}}}} &
    \includegraphics[width=0.16\textwidth]{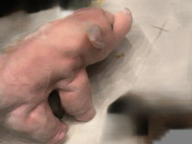}&
    \includegraphics[width=0.16\textwidth]{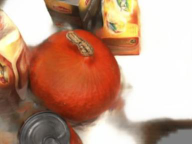}&
    \includegraphics[width=0.16\textwidth]{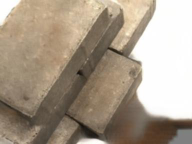}&
    \includegraphics[width=0.16\textwidth]{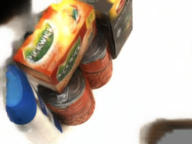}&
    \includegraphics[width=0.16\textwidth]{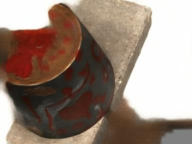}&
    \includegraphics[width=0.16\textwidth]{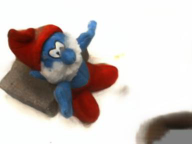}\\
 
    &
    \includegraphics[width=0.16\textwidth]{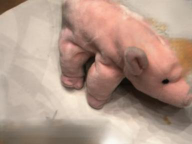}&
    \includegraphics[width=0.16\textwidth]{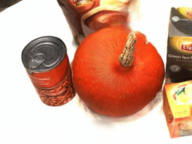}&
    \includegraphics[width=0.16\textwidth]{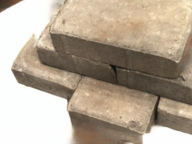}&
    \includegraphics[width=0.16\textwidth]{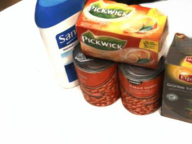}&
    \includegraphics[width=0.16\textwidth]{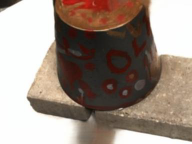}&
    \includegraphics[width=0.16\textwidth]{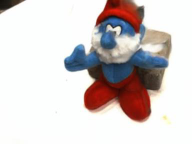}\\
    &
    \includegraphics[width=0.16\textwidth]{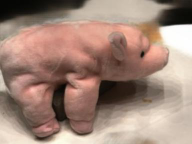}&
    \includegraphics[width=0.16\textwidth]{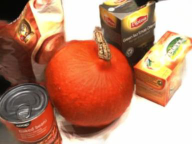}&
    \includegraphics[width=0.16\textwidth]{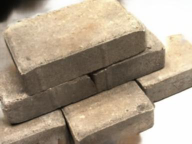}&
    \includegraphics[width=0.16\textwidth]{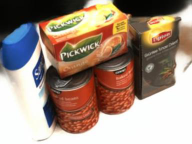}&
    \includegraphics[width=0.16\textwidth]{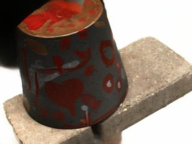}&
    \includegraphics[width=0.16\textwidth]{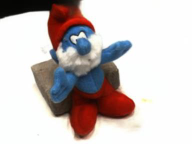}\\
    &
    \includegraphics[width=0.16\textwidth]{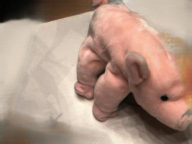}&
    \includegraphics[width=0.16\textwidth]{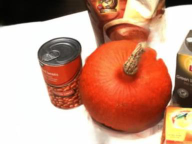}&
    \includegraphics[width=0.16\textwidth]{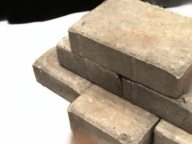}&
    \includegraphics[width=0.16\textwidth]{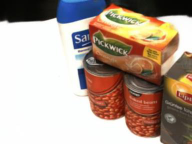}&
    \includegraphics[width=0.16\textwidth]{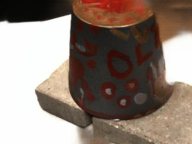}&
    \includegraphics[width=0.16\textwidth]{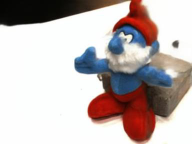}\\
    &
    \includegraphics[width=0.16\textwidth]{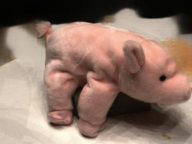}&
    \includegraphics[width=0.16\textwidth]{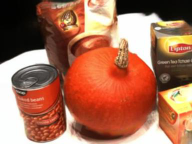}&
    \includegraphics[width=0.16\textwidth]{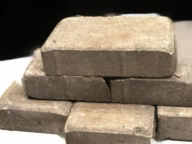}&
    \includegraphics[width=0.16\textwidth]{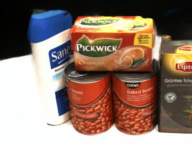}&
    \includegraphics[width=0.16\textwidth]{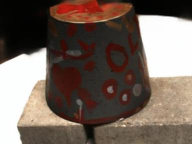}&
    \includegraphics[width=0.16\textwidth]{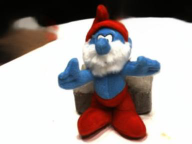}\\
    \end{tabular}
    \captionof{figure}{\textbf{Efficient view synthesis from very sparse views for FVS. }
    We show the view synthesis results with 3 input views on DTU MVS test dataset for FVS.
    }
    \label{fig:sup_dtu_FVS}
    \end{figure*}

\begin{figure*}
    \centering
    \begin{tabular}{>{\footnotesize}c@{}@{}c@{\hspace{0.5mm}}c@{\hspace{0.5mm}}c@{\hspace{0.5 mm}}c@{\hspace{0.5mm}}c@{\hspace{0.5mm}}c@{}}
    \parbox[t]{4mm}{\multirow{3}{*}{\rotatebox[origin=c]{90}{\text{Input: 3 views of held-out scene}}}} & 
    \includegraphics[width=0.16\textwidth]{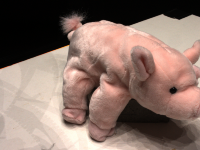}&
    \includegraphics[width=0.16\textwidth]{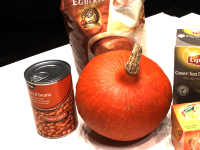}&
    \includegraphics[width=0.16\textwidth]{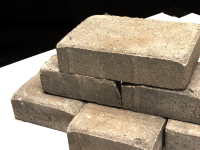}&
    \includegraphics[width=0.16\textwidth]{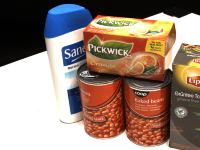}&
    \includegraphics[width=0.16\textwidth]{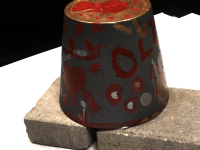}&
    \includegraphics[width=0.16\textwidth]{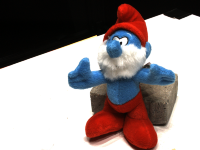}\\
 &
   \includegraphics[width=0.16\textwidth]{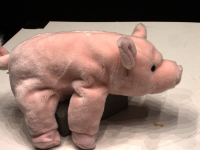}&
   \includegraphics[width=0.16\textwidth]{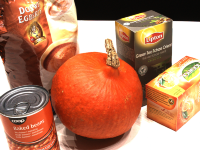}&
   \includegraphics[width=0.16\textwidth]{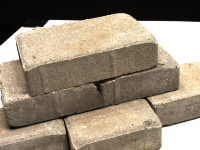}&
   \includegraphics[width=0.16\textwidth]{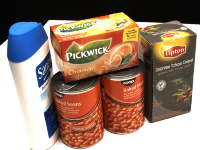}&
   \includegraphics[width=0.16\textwidth]{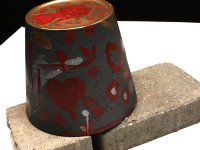}&
   \includegraphics[width=0.16\textwidth]{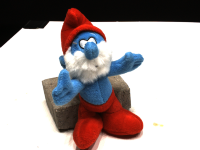}\\
    &
    \includegraphics[width=0.16\textwidth]{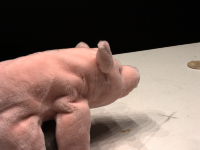}&
    \includegraphics[width=0.16\textwidth]{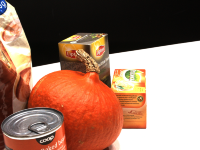}&
    \includegraphics[width=0.16\textwidth]{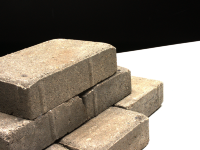}&
    \includegraphics[width=0.16\textwidth]{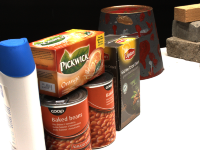}&
    \includegraphics[width=0.16\textwidth]{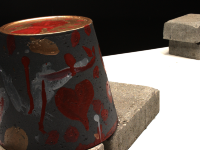}&
    \includegraphics[width=0.16\textwidth]{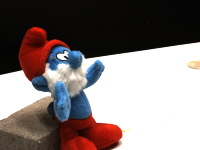}\\
    \bottomrule[2pt]
    \parbox[t]{4mm}{\multirow{5}{*}{\rotatebox[origin=c]{90}{\text{Novel views}}}} &
    \includegraphics[width=0.16\textwidth]{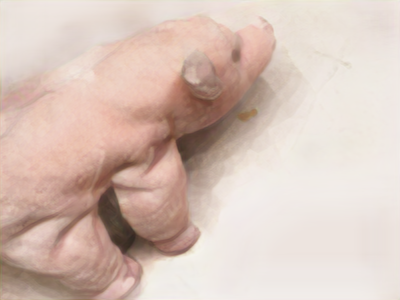}&
    \includegraphics[width=0.16\textwidth]{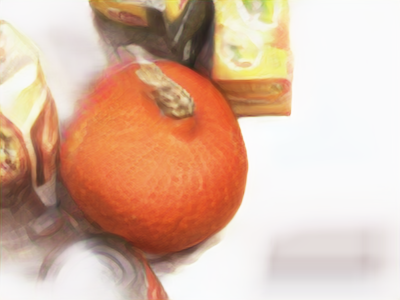}&
    \includegraphics[width=0.16\textwidth]{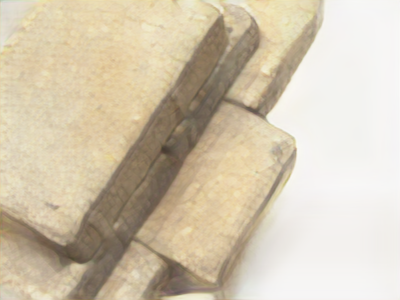}&
    \includegraphics[width=0.16\textwidth]{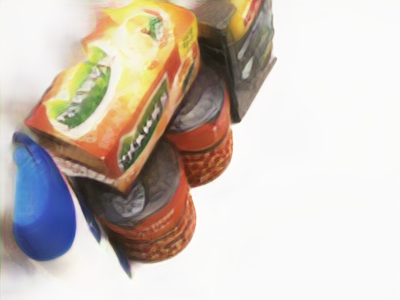}&
    \includegraphics[width=0.16\textwidth]{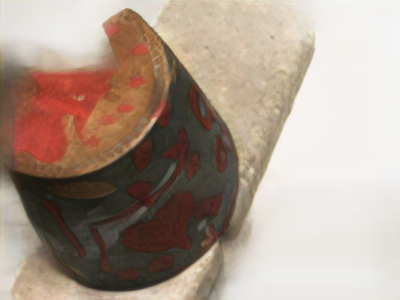}&
    \includegraphics[width=0.16\textwidth]{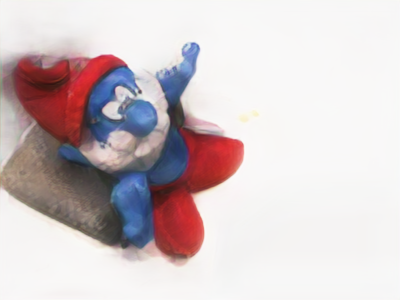}\\
 
    &
    \includegraphics[width=0.16\textwidth]{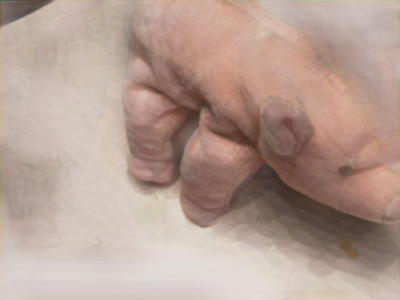}&
    \includegraphics[width=0.16\textwidth]{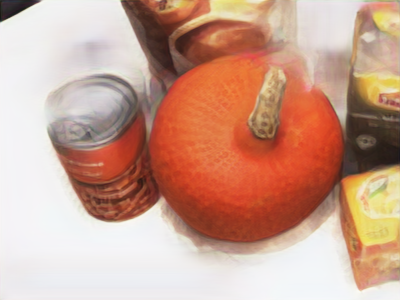}&
    \includegraphics[width=0.16\textwidth]{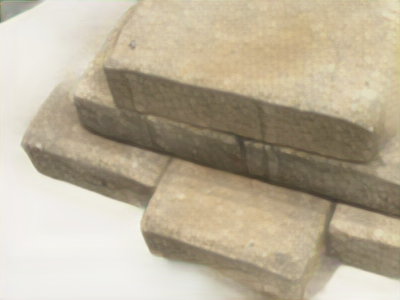}&
    \includegraphics[width=0.16\textwidth]{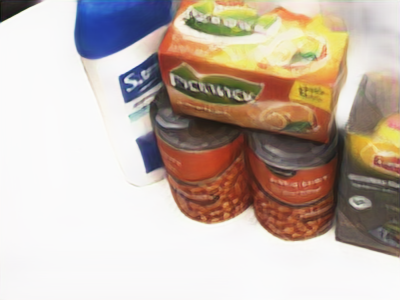}&
    \includegraphics[width=0.16\textwidth]{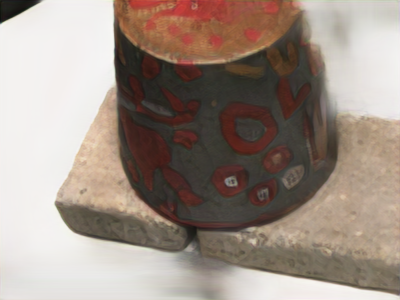}&
    \includegraphics[width=0.16\textwidth]{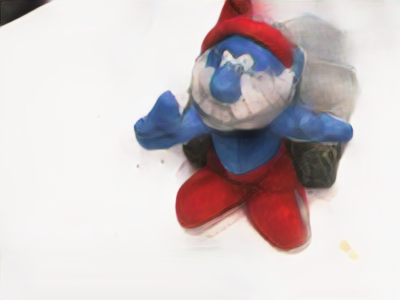}\\
    &
    \includegraphics[width=0.16\textwidth]{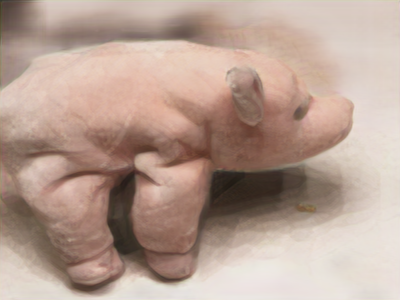}&
    \includegraphics[width=0.16\textwidth]{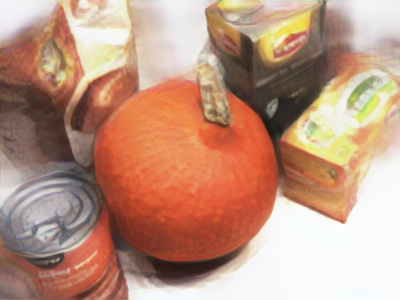}&
    \includegraphics[width=0.16\textwidth]{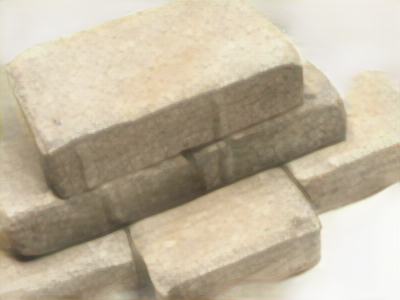}&
    \includegraphics[width=0.16\textwidth]{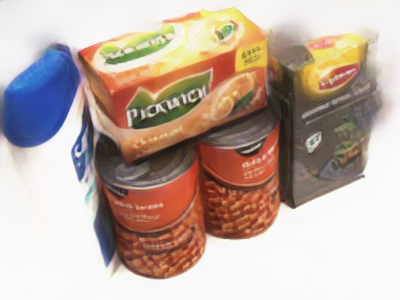}&
    \includegraphics[width=0.16\textwidth]{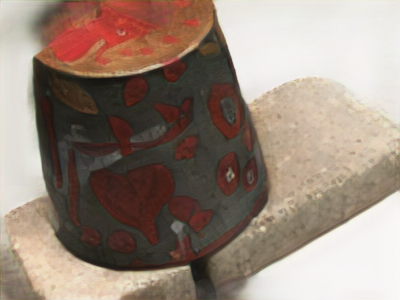}&
    \includegraphics[width=0.16\textwidth]{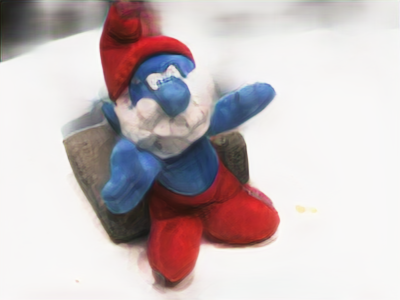}\\
    &
    \includegraphics[width=0.16\textwidth]{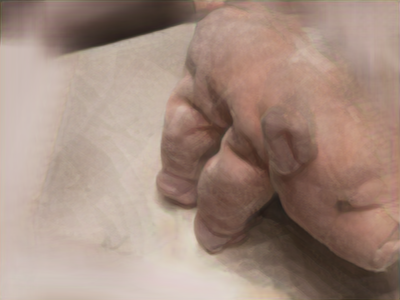}&
    \includegraphics[width=0.16\textwidth]{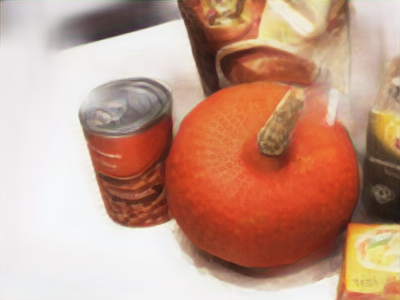}&
    \includegraphics[width=0.16\textwidth]{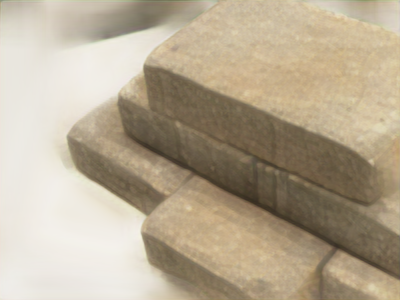}&
    \includegraphics[width=0.16\textwidth]{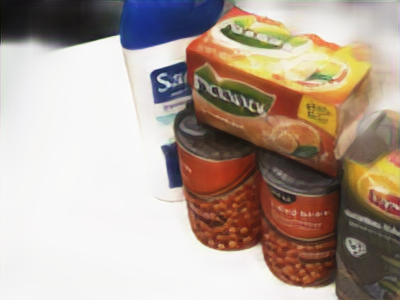}&
    \includegraphics[width=0.16\textwidth]{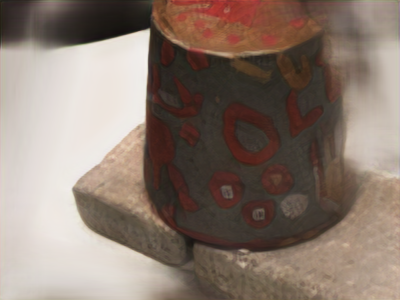}&
    \includegraphics[width=0.16\textwidth]{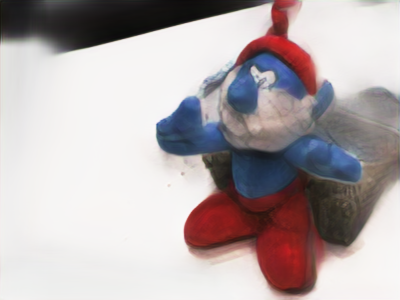}\\
    &
    \includegraphics[width=0.16\textwidth]{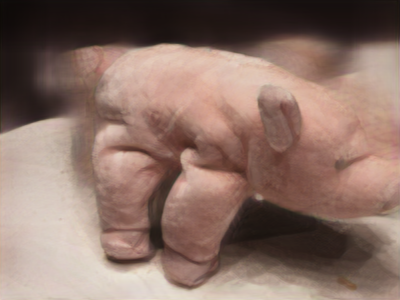}&
    \includegraphics[width=0.16\textwidth]{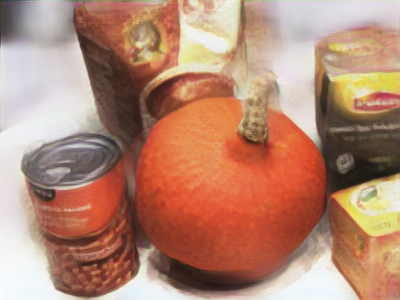}&
    \includegraphics[width=0.16\textwidth]{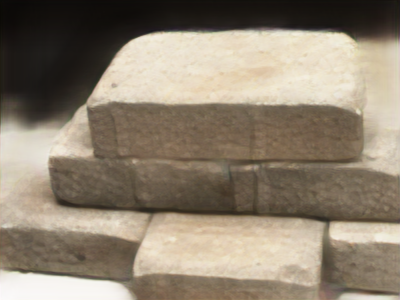}&
    \includegraphics[width=0.16\textwidth]{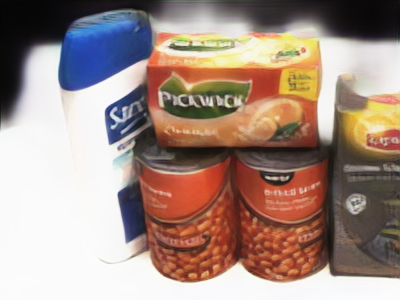}&
    \includegraphics[width=0.16\textwidth]{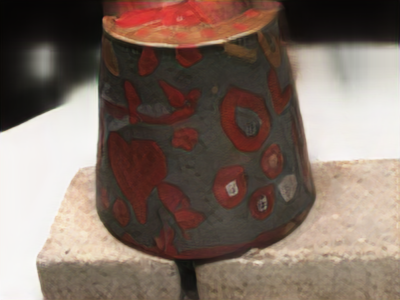}&
    \includegraphics[width=0.16\textwidth]{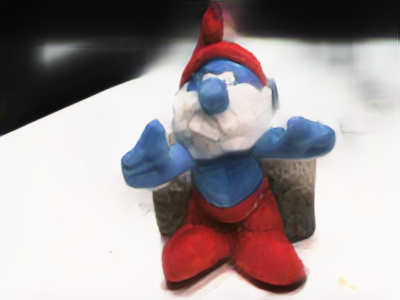}\\
    \end{tabular}
    \captionof{figure}{\textbf{Efficient view synthesis from very sparse views for Blending-R. }
    We show the view synthesis results with 3 input views on DTU MVS test dataset for Blending-R.
    }
    \label{fig:sup_dtu_blending}
    \end{figure*}

We show more synthesis results in the following.
For comparison, we show results of the same scene and views.
We first show the results of FWD-U in Figure~\ref{fig:fwd_u}, FWD in Figure~\ref{fig:fwd} and FWD-D in Figure~\ref{fig:fwd_d}.
We also show baseline results: PixelNeRF in Figure~\ref{fig:sup_pixelnerf}, IBRNet in Figure~\ref{fig:sup_dtu_IBRNet},
MVSNeRF in Figure~\ref{fig:sup_dtu_mvsnerf}, FVS in Figure~\ref{fig:sup_dtu_FVS} and Blending-R in Figure~\ref{fig:sup_dtu_blending}.
Again, please see attached videos for comparison between ours and other methods.

\end{document}